\pdfoutput=1

\documentclass[11pt]{article}

\usepackage[]{acl}

\usepackage{times}
\usepackage{latexsym}

\usepackage[T1]{fontenc}

\usepackage[utf8]{inputenc}

\usepackage{microtype}

\usepackage{amsmath}
\usepackage{amssymb}
\usepackage{booktabs}
\usepackage{graphicx}
\usepackage{subcaption}
\usepackage{adjustbox}
\usepackage{multirow}

\usepackage[utf8]{inputenc}
\usepackage{newunicodechar}
\newunicodechar{ḩ}{h}
\newunicodechar{ư}{u}

\title{Axis Tour: Word Tour Determines the Order of Axes in\\ICA-transformed Embeddings}

\author{
  Hiroaki Yamagiwa${}^{1}$ \qquad 
  Yusuke Takase${}^{1}$ \qquad
  Hidetoshi Shimodaira${}^{1,2}$ \\
  ${}^1\,$Kyoto University \qquad ${}^2\,$RIKEN AIP\\
  \texttt{\{hiroaki.yamagiwa,y.takase\}@sys.i.kyoto-u.ac.jp,}\\
  \texttt{shimo@i.kyoto-u.ac.jp}\\
}

\begin{document}
\maketitle
\begin{abstract}
Word embedding is one of the most important components in natural language processing, but interpreting high-dimensional embeddings remains a challenging problem. To address this problem, Independent Component Analysis (ICA) is identified as an effective solution. ICA-transformed word embeddings reveal interpretable semantic axes; however, the order of these axes are arbitrary. In this study, we focus on this property and propose a novel method, Axis Tour, which optimizes the order of the axes. Inspired by Word Tour, a one-dimensional word embedding method, we aim to improve the clarity of the word embedding space by maximizing the semantic continuity of the axes. Furthermore, we show through experiments on downstream tasks that Axis Tour yields better or comparable low-dimensional embeddings compared to both PCA and ICA.
\end{abstract}

\section{Introduction}\label{sec:intro}
\begin{figure}[t!]
    \centering
    \includegraphics[width=0.95\columnwidth]{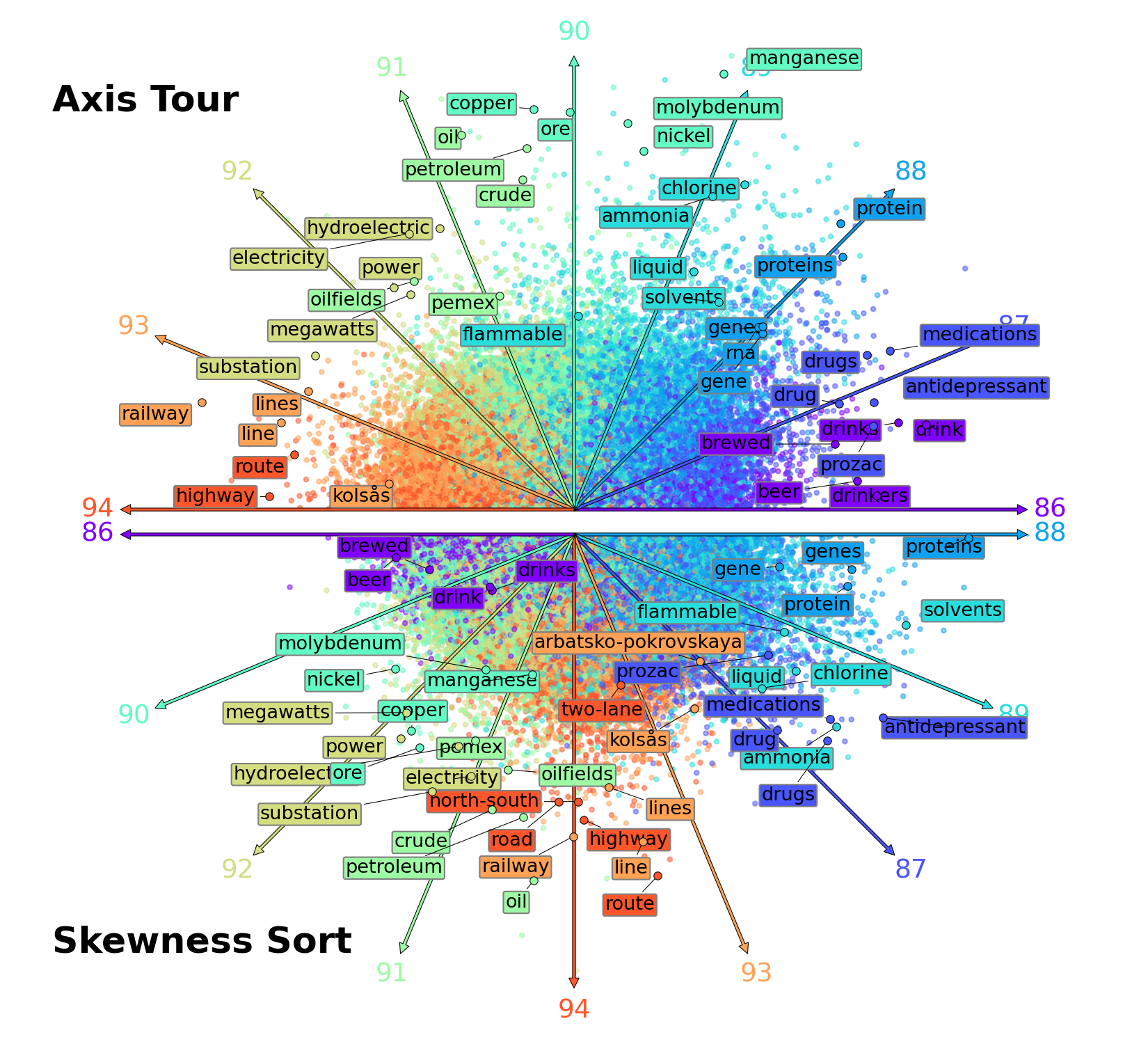}
    \caption{
Scatterplots of normalized ICA-transformed word embeddings whose axes are ordered by Axis Tour and Skewness Sort. In the upper part, Axis Tour is applied to 300-dimensional GloVe, with nine consecutive axes arranged counterclockwise. In the lower part, these nine axes are rearranged clockwise in descending order of skewness. The embeddings are projected onto two dimensions along these axes. The top five embeddings on each axis are labeled by their words. Each word is assigned the color of the axis with the highest value. In both cases, words that cross the horizontal axes are not displayed. Refer to Appendix~\ref{app:fig-explanation} for more details.
}
\label{fig:intro}
\end{figure}
Embedding is an important tool in natural language processing, but interpreting high-dimensional embeddings is challenging. 
To address this, Independent Component Analysis (ICA)~\cite{DBLP:journals/nn/HyvarinenO00} offers an effective solution~\cite{marecek-etal-2020,musil-marecek-2024-exploring,DBLP:conf/emnlp/YamagiwaOS23}. 
ICA-transformed embeddings reveal interpretable semantic axes; however, the order of these axes is arbitrary~\cite{DBLP:books/wi/HyvarinenKO01}.
In this study, inspired by a one-dimensional word embedding method, Word Tour~\cite{DBLP:conf/naacl/Sato22}, which leverages the Traveling Salesman Problem (TSP), we aim to improve the clarity of the word embedding space by maximizing the semantic continuity of the axes. 

Figure~\ref{fig:intro} shows two sets of two-dimensional projections of word embeddings: one is ordered by Axis Tour (our proposal), and the other is sorted in descending order of skewness (Skewness Sort). 
In Axis Tour, the top words of the axes are positioned farther from the center, with the meanings of the axes changing continuously. 
Conversely, in Skewness Sort, the top words are closer to the center, and the axes with different meanings are placed adjacently. 
In fact, the average distance from the origin to the top words in Fig.~\ref{fig:intro} is $0.76$ in Axis Tour, compared to $0.61$ in Skewness Sort. 

We also assume that the consecutive axes in the Axis Tour embeddings can be considered a subspace with axes that have similar meanings.
Based on this idea, we project each subspace onto a single dimension for dimensionality reduction.
We show through experiments that Axis Tour yields better or comparable low-dimensional embeddings compared to both PCA and ICA.

\section{Related work}
Some studies transform embeddings by rotation~\cite{DBLP:conf/emnlp/ParkBO17} from Factor Analysis~\cite{crawford1970general, browne2001overview} or Principal Component Analysis (PCA)~\cite{DBLP:conf/tsd/Musil19}. 
Independent Component Analysis (ICA)~\cite{DBLP:journals/nn/HyvarinenO00} has gained attention for its ability to reveal interpretable semantic axes in the transformed embeddings~\cite{marecek-etal-2020,musil-marecek-2024-exploring,DBLP:conf/emnlp/YamagiwaOS23}. 

Research on interpreting embeddings by focusing on axes representing opposing concepts (e.g., \emph{cold} vs. \emph{hot}, \emph{soft} vs. \emph{hard}) is also actively pursued.
Approaches such as SemAxis~\cite{DBLP:conf/acl/AnKA18}, POLAR~\cite{DBLP:conf/www/MathewSLS20}, and FrameAxis~\cite{DBLP:journals/peerj-cs/KwakAJA21} deal with static embeddings, while BiImp~\cite{DBLP:journals/ipm/SenelSYSCK22} and SensePOLAR~\cite{DBLP:conf/emnlp/EnglerSLS22} deal with dynamic embeddings.
In particular, Section~\ref{sec:downstream} provides a comparison of the Axis Tour embeddings and those from POLAR.

Relevant to our study is Topographic ICA (TICA)~\cite{DBLP:books/sp/Kohonen01,DBLP:journals/neco/HyvarinenHI01}. 
TICA relaxes the assumption of statistical independence and assumes higher-order correlations between adjacent axes. then estimates the order of the axes. 
Unlike TICA, Axis Tour is applied to ordinary ICA-transformed embeddings and uses the embeddings themselves to measure axis similarity. 
For more details on TICA, refer to Appendix~\ref{app:tica}.

\section{Background}
The pre-trained word embedding matrix is given by $\mathbf{X}=[\mathbf{x}_1,\ldots,\mathbf{x}_n]^\top\in\mathbb{R}^{n\times d}$, where $\mathbf{X}$ is \emph{centered} (i.e., the mean of each column is zero).
Here, $\mathbf{x}_{i} \in \mathbb{R}^d$ represents the word embedding of the $i$-th word.

\subsection{ICA-transformed word embeddings}
ICA~\cite{DBLP:journals/nn/HyvarinenO00} finds the transformation matrix $\mathbf{B}\in\mathbb{R}^{d\times d}$ such that the columns of the matrix $\mathbf{S}\in\mathbb{R}^{n\times d}$, represented by the following equation, are as independent as possible:
\begin{align}
    \mathbf{S} = \mathbf{X}\mathbf{B},\label{eq:ica}
\end{align}
where $\mathbf{S}$ is \emph{whitened} (i.e., the variances of the columns are $1$ and their correlations are all $0$). The columns of $\mathbf{S}$ are called independent components\footnote{Unless otherwise noted, flip the sign of each axis as needed so that the skewness is positive.}. 
While $\mathbf{S}$ has interpretable semantic axes~\cite{marecek-etal-2020,musil-marecek-2024-exploring,DBLP:conf/emnlp/YamagiwaOS23}, the order of these axes are arbitrary~\cite{DBLP:books/wi/HyvarinenKO01}. 

\subsection{Word Tour}\label{sec:wordtour}
Let $\mathcal{P}([n])$ be the set of all permutations of $[n]$, where $[n] = \{1,\ldots,n\}$.
Word Tour~\cite{DBLP:conf/naacl/Sato22} is a one-dimensional word embedding method that solves the following Traveling Salesman Problem (TSP):
\begin{align}
    \underset{\tau\in\mathcal{P}([n])} {\operatorname{min}}\,\|\mathbf{x}_{\tau_1}-\mathbf{x}_{\tau_n}\|+\sum_{i=1}^{n-1}\|\mathbf{x}_{\tau_i}-\mathbf{x}_{\tau_{i+1}}\|.\label{eq:wordtour}
\end{align}
The resulting one-dimensional embeddings have similar meanings when they are close in order.

\section{Axis Tour}\label{sec:axistour}
This section explains Axis Tour and the dimensionality reduction method using Axis Tour. 
As mentioned in Section~\ref{sec:wordtour}, $[d] = \{1,\ldots,d\}$ and $\mathcal{P}([d])$ is the set of all permutations of $[d]$. 

\subsection{Definition of axis embedding}\label{sec:def-v}
We define \emph{axis embedding} for use in Word Tour. The embedding represents the meaning of the axis of the ICA-transformed embeddings $\mathbf{S}$.

In preparation, we define the \emph{normalized ICA-transformed embeddings} $\hat{\mathbf{S}}\in\mathbb{R}^{n\times d}$ as the normalization of the embeddings $\mathbf{S}$, where the row vectors are given by $\hat{\mathbf{s}}_i = \mathbf{s}_i / \|\mathbf{s}_i\|$. 
Here, the $i$-th word embeddings of $\mathbf{S}$ and $\hat{\mathbf{S}}$ are denoted by $\mathbf{s}_i$ and $\hat{\mathbf{s}}_i \in \mathbb{R}^d$, respectively.
We compare the elements of the $\ell$-th axis of $\hat{\mathbf{S}}$ and denote the index set of words corresponding to the top $k$ elements as $\text{Top}_k^\ell$.
We then define the $\ell$-th axis embedding $\mathbf{v}_\ell$ for $\mathbf{S}$ as follows:
\begin{align}
    \mathbf{v}_\ell := \frac{1}{k}\sum_{i\in \text{Top}_k^\ell}\hat{\mathbf{s}}_{i}\in\mathbb{R}^d.\label{eq:axis-embedding}
\end{align}
As we saw in Fig.~\ref{fig:intro}, since the meaning of an axis can be interpreted from the top words, $\mathbf{v}_\ell$ can be considered the embedding that represents the meaning of the $\ell$-th axis of $\mathbf{S}$.

\begin{table*}[t]
\centering
\scriptsize
\begin{tabular}{ccccccccc}
\toprule
        23 &         24 &      25 &       26 &      27 &            28 &          29 &            30 &         31 \\
\midrule
      serb &    russian &   czech &  germany &  france &        canada &   australia &     wiltshire &       liga \\
   bosnian &     russia &  prague &   german &  french &      canadian &  australian &    shrewsbury &  relegated \\
   croatia &     moscow &  poland &   berlin &      le &       ontario &  queensland &  lincolnshire &         fc \\
  croatian &     sergei &  polish &      von &   paris &        quebec &    brisbane &  peterborough &       f.c. \\
   serbian &  aleksandr &  warsaw &  cologne &      du &  saskatchewan &       perth &       croydon &      serie \\
\midrule
\midrule
101 &         102 &          103 &        104 &         105 &          106 &         107 &          108 &         109 \\
\midrule
pay &        land &         laws &      court &    lawsuits &      charges &        camp &      corpses &      remain \\
fees &    property &  regulations &      judge &     lawsuit &      alleged &      prison &       corpse &    remained \\
payments &       lands &      enacted &  appellate &  litigation &  prosecutors &  buchenwald &      exhumed &      stayed \\
payment &      estate &          law &    appeals &       suits &     indicted &       camps &  dismembered &  stubbornly \\
paid &  bergisches &   provisions &    supreme &        suit &    convicted &     inmates &       bodies &       stays \\
\midrule
\midrule
237 &       238 &          239 &          240 &         241 &         242 &          243 &         244 &          245 \\
\midrule
 award &      film &    superhero &        album &       piano &   paintings &   manuscript &    language &         name \\
awards &     films &       marvel &       albums &      violin &    painting &  biographies &   languages &        names \\
awarded &     movie &     spin-off &         band &       cello &         art &        pages &      pashto &      surname \\
prize &  starring &  superheroes &  self-titled &  percussion &   sculpture &         book &  colloquial &       phrase \\
emmy &  directed &   characters &           ep &  orchestral &  watercolor &  handwritten &     dialect &  misspelling \\
\bottomrule
\end{tabular}
\caption{
Semantic continuity of axes by Axis Tour for normalized ICA-transformed embeddings. We apply Axis Tour to 300-dimensional GloVe and show the top five words for each axis. See Appendix~\ref{app:all-examples} for all axes results.
}
\label{tab:examples}
\end{table*}

\subsection{Determining the order of axes}\label{sec:axistour-object}
Axis Tour is a method that uses $\mathbf{v}_\ell$ in (\ref{eq:axis-embedding}) to perform Word Tour and determines the order of axes in ICA-transformed word embeddings. In Axis Tour, the cost between the axis embeddings $\mathbf{v}_\ell$ and $\mathbf{v}_m$ for the TSP is defined by $1-\cos{(\mathbf{v}_\ell,\mathbf{v}_m)}$ instead of $\|\mathbf{v}_\ell-\mathbf{v}_m\|$. This approach then maximizes the sum of cosine similarities between adjacent axis embeddings. Therefore, the problem is formulated as follows\footnote{Note that due to the cyclic nature of $\tau$, we set $\tau_1$ such that $\cos(\mathbf{v}_{\tau_1},\mathbf{v}_{\tau_d})$ is the smallest of the cosine similarities.}:
\begin{align}
    \underset{\tau\in\mathcal{P}([d])} {\operatorname{max}}\,\cos(\mathbf{v}_{\tau_1},\mathbf{v}_{\tau_d})+\sum_{\ell=1}^{d-1}\cos(\mathbf{v}_{\tau_\ell},\mathbf{v}_{\tau_{\ell+1}}).\label{eq:axistour}
\end{align}
The sum of cosine similarities between adjacent axis embeddings can be considered as a metric of the semantic continuity of the axes. Thus, Axis Tour determines the order of the axes by maximizing this metric.

\subsection{Dimensionality reduction}\label{sec:dim-reduction}
Let $\mathbf{T}=[\mathbf{t}_1,\ldots,\mathbf{t}_n]^\top\in\mathbb{R}^{n\times d}$ be the matrix $\mathbf{S}$ with Axis Tour applied, where the optimal $\tau$ is applied to the columns of $\mathbf{S}$ to produce $\mathbf{T}$.
We consider reducing the dimensions from $d$ to $p\,(\leq d)$ by merging the consecutive axes of $\mathbf{T}$. In preparation, we divide $[d]$ into $p$ equal-length intervals\footnote{
The first $d\% p$ intervals are $\lfloor d/p \rfloor+1$ in length, and the rest are $\lfloor d/p \rfloor$ in length, where $\lfloor\cdot\rfloor$ is the floor function.
} and define the index set for the $r$-th interval as $I_{r}:=\{a_r,\ldots,b_r\}\,(a_r,b_r\in [d], a_r\leq b_r)$. Let $\gamma_\ell \in \mathbb{R}_{\geq 0}$ be the skewness\footnote{Since the skewness of the axis of $\mathbf{S}$ is positive, $\gamma_\ell \geq 0$.} of the $\ell$-th axis of $\mathbf{T}$.

First, we consider reducing the dimensionality of $\mathbf{T}$ along $I_r$, $r=1,\ldots,p$. To do this, we define a unit vector $\mathbf{f}_r:=(f_r^{(\ell)})_{\ell=1}^d\in\mathbb{R}_{\geq 0}^d$ for each $I_r$ as follows:
\begin{align}
   f_r^{(\ell)} = 
   \begin{cases} 
\gamma_\ell^\alpha/\sqrt{\sum_{m=a_r}^{b_r}\gamma_m^{2\alpha}} & \text{for }\ell\in I_r \\
   0 & \text{otherwise},
   \end{cases}
   \label{eq:def-f}
\end{align}
where $\alpha \in \mathbb{R}_{\geq 0}$. Then $\mathbf{T}\mathbf{f}_r \in \mathbb{R}^{n}$ can be considered as a projection of the subspace spanned by the axes of $\mathbf{T}$ corresponding to $I_r$ onto a one-dimensional space. Fig.~\ref{fig:3dproj} in Appendix~\ref{app:dim-reduction} shows the projection for three consecutive axes.

Next, we define the matrix $\mathbf{F} := [\mathbf{f}_1, \ldots, \mathbf{f}_p] \in \mathbb{R}^{d \times p}$. Then $\mathbf{T}\mathbf{F}\in\mathbb{R}^{n\times p}$ represents the concatenated projections, that is, a dimensionality reduction of the $d$-dimensional embeddings $\mathbf{T}$ to $p$ dimensions. For more details, refer to Appendix~\ref{app:dim-reduction}.

\section{Experiments}\label{sec:experiments}
Similar to the Word Tour experiments, we used 300-dimensional GloVe
~\cite{DBLP:conf/emnlp/PenningtonSM14} with $n=400{,}000$, and the LKH solver\footnote{The LKH solver is an implementation of Lin-Kernighan algorithm~\cite{lin1973effective, helsgaun2000effective}.}~\cite{lkh_webpage} for the optimization of the TSP.

For ICA, we used FastICA~\cite{DBLP:journals/tnn/Hyvarinen99} from scikit-learn~\cite{DBLP:journals/jmlr/PedregosaVGMTGBPWDVPCBPD11}, setting the iterations to $10{,}000$ and the tolerance to $10^{-10}$, consistent with \citet{DBLP:conf/emnlp/YamagiwaOS23}. We computed the axis embeddings\footnote{See Appendix~\ref{app:comp-k} for a discussion of the choice of $k$.} $\mathbf{v}_\ell$ in (\ref{eq:axis-embedding}) with $k=100$. For baselines, we used whitened {\bf PCA}-transformed embeddings\footnote{Whitened ICA-transformed embeddings are obtained by applying an orthogonal matrix to these embeddings.}, along with two types of whitened ICA-transformed embeddings\footnote{See Appendix~\ref{app:tica} for comparisons of Axis Tour and TICA.}: {\bf Random Order}, which randomly flips the sign of the axes in $\mathbf{S}$ and randomizes the order of the axes, and {\bf Skewness Sort}, which sorts the axes of $\mathbf{S}$ in descending order of skewness. See Appendix~\ref{app:additional-experiments} for additional experiments, including those of other embeddings.

\subsection{Qualitative observation of semantic continuity in axis order}
Table~\ref{tab:examples} presents an illustrative example of consecutive axes of the Axis Tour embeddings, where the three rows correspond to the meanings of \emph{countries}, \emph{law}, and \emph{art}, respectively.
We observe that the meanings of the axes change continuously. 
For instance, in the top row, the axis meaning shifts from \emph{Eastern Europe} to \emph{Germany} and \emph{France}, followed by \emph{Canada} (which shares a connection with France), then to \emph{Australia} (English-speaking regions),  \emph{the regions in England}, and finally to \emph{soccer} (a popular sport in England), demonstrating geographic and cultural continuity.

\subsection{Quantitative evaluation of semantic continuity in axis order}
This section quantitatively evaluates the semantic continuity of the axes in Axis Tour embeddings.

\subsubsection{Evaluation by cosine similarity}
First, we evaluate the two orderings of axes shown in Fig.~\ref{fig:intro}.
The semantic continuity of these axes is assessed by calculating the average cosine similarity between adjacent axis embeddings.
For Axis Tour, the average cosine similarity is 0.269, but it decreases to 0.185 when these axes are rearranged by skewness, confirming the higher semantic continuity of the axes in Axis Tour.

\begin{figure}
    \centering
    \includegraphics[width=0.9\columnwidth]{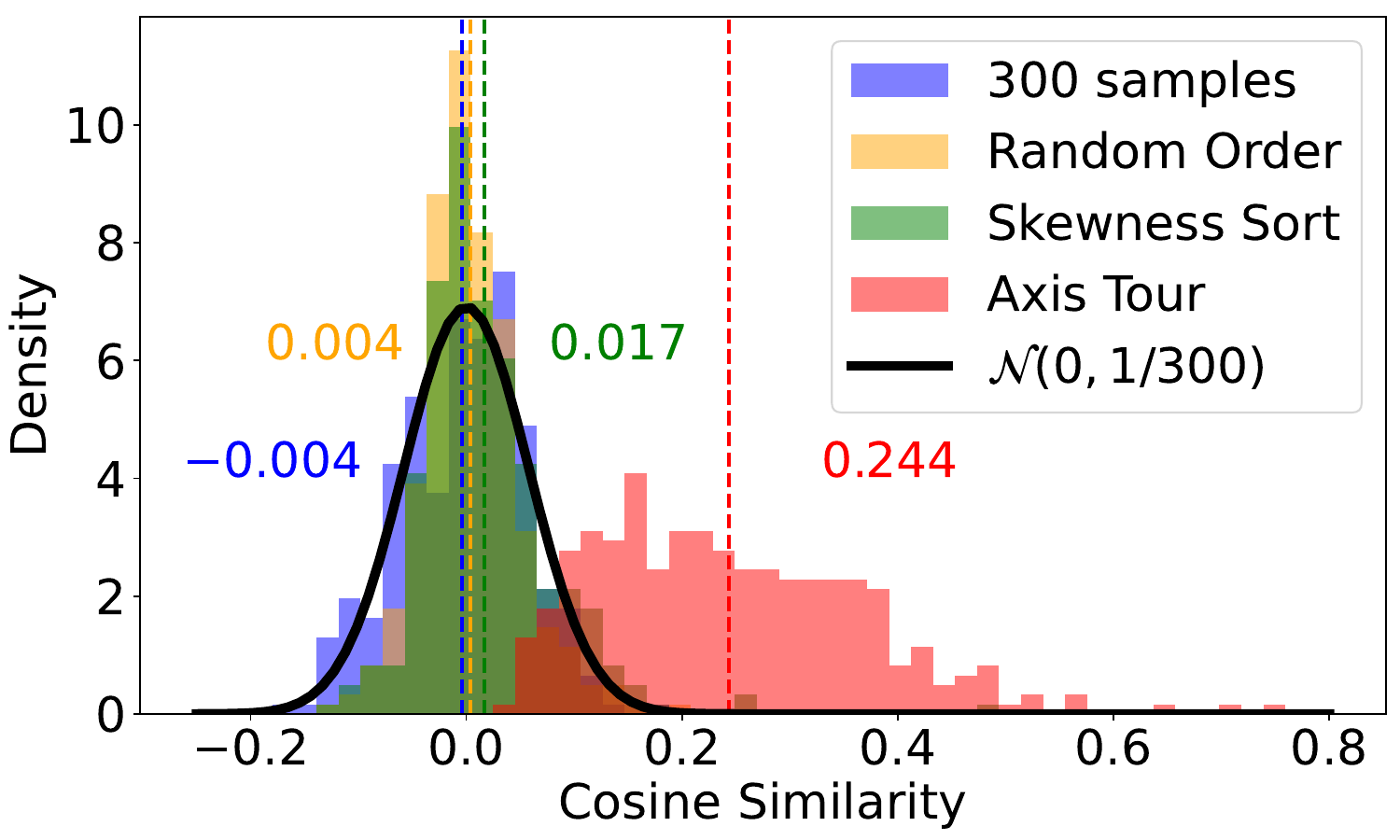}
    \caption{
Histogram of $\cos(\mathbf{v}_\ell, \mathbf{v}_{\ell+1})$.  As an additional baseline, we sampled 300 random words from the Random Order embeddings and arranged them in random order. The dashed lines represent the average similarity for each method. The distribution for Axis Tour shifts towards a more positive mean, while the others roughly follow a normal distribution with means close to 0. For more details, refer to Appendix~\ref{app:dist-cos-sim}. 
}
    \label{fig:cos-hist}
\end{figure}

Next, we consider the whole $d\,(=300)$ consecutive axes.
Figure~\ref{fig:cos-hist} shows the histograms of $\cos(\mathbf{v}_\ell,\mathbf{v}_{\ell+1})$ for Axis Tour and the baselines. In Axis Tour, the values of $\cos(\mathbf{v}_\ell,\mathbf{v}_{\ell+1})$ are consistently higher, while this trend is not observed in the other baselines. 
The average cosine similarity is 0.244 for Axis Tour, while it is only 0.017 for Skewness Sort.
This result is consistent with the formulation in (\ref{eq:axistour}).

\subsubsection{Evaluation by GPT models}\label{sec:GPT}
We also evaluate the semantic continuity of axes for Axis Tour and Skewness Sort using the OpenAI API. 
By focusing on the common axes in each embedding, we ask the model to determine, based on the top 10 words, whether the next axis of Axis Tour or that of Skewness Sort is more semantically related.
The number of queries corresponds to $d\,(=300)$, and we use four GPT models: GPT-3.5 Turbo, GPT-4 Turbo, GPT-4o, and GPT-4o mini (see Appendix~\ref{app:gpt} for model versions and prompts).

\begin{figure}
    \centering
    \includegraphics[width=0.85\columnwidth]{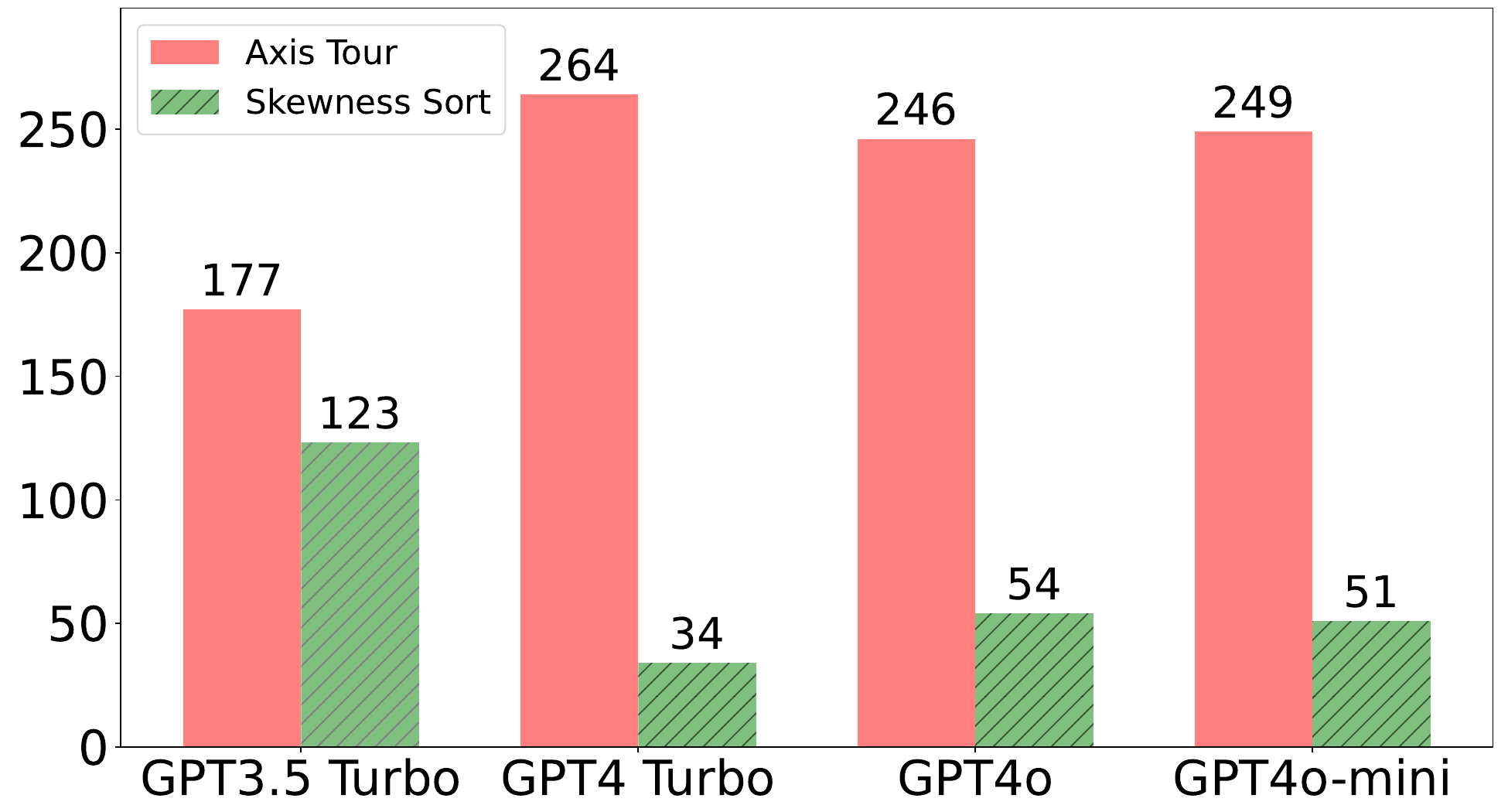}
    \caption{
Comparison of the number of related axes in the GPT models. 
In each model, Axis Tour exhibits a greater number of related axes than Skewness Sort.
}
    \label{fig:GPT}
\end{figure}

As shown in Fig.~\ref{fig:GPT}, Axis Tour has a greater number of related axes compared to Skewness Sort for each model, implying more continuous changes in axis meanings.
The smallest difference was observed with GPT-3.5 Turbo, the least performant model.
For the other models (i.e., the GPT-4 models), the difference was at least four times larger.

\begin{figure*}[t]
    \centering
    \includegraphics[width=0.80\linewidth]{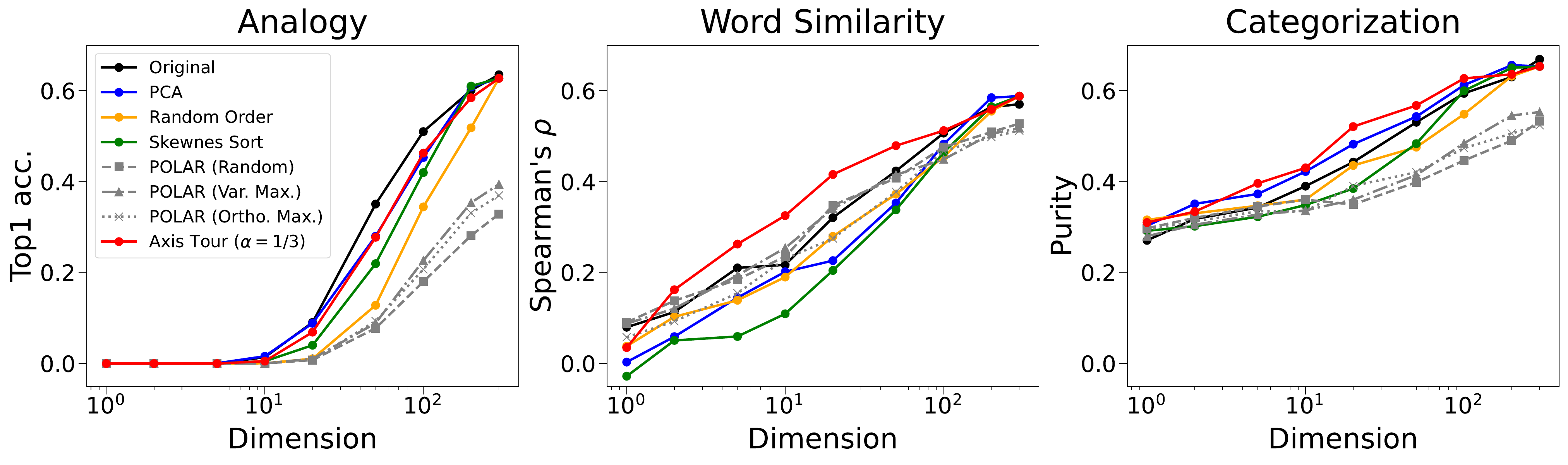}
    \caption{
The performance of dimensionality reduction for embeddings. Each value represents the average of 30 analogy tasks, 8 word similarity tasks, or 6 categorization tasks. See Appendix~\ref{app:experiments} for detailed experimental results.
}
    \label{fig:downstream}
\end{figure*}

\begin{figure}[!t]
\centering
\begin{subfigure}{\columnwidth}
    \includegraphics[width=\textwidth]{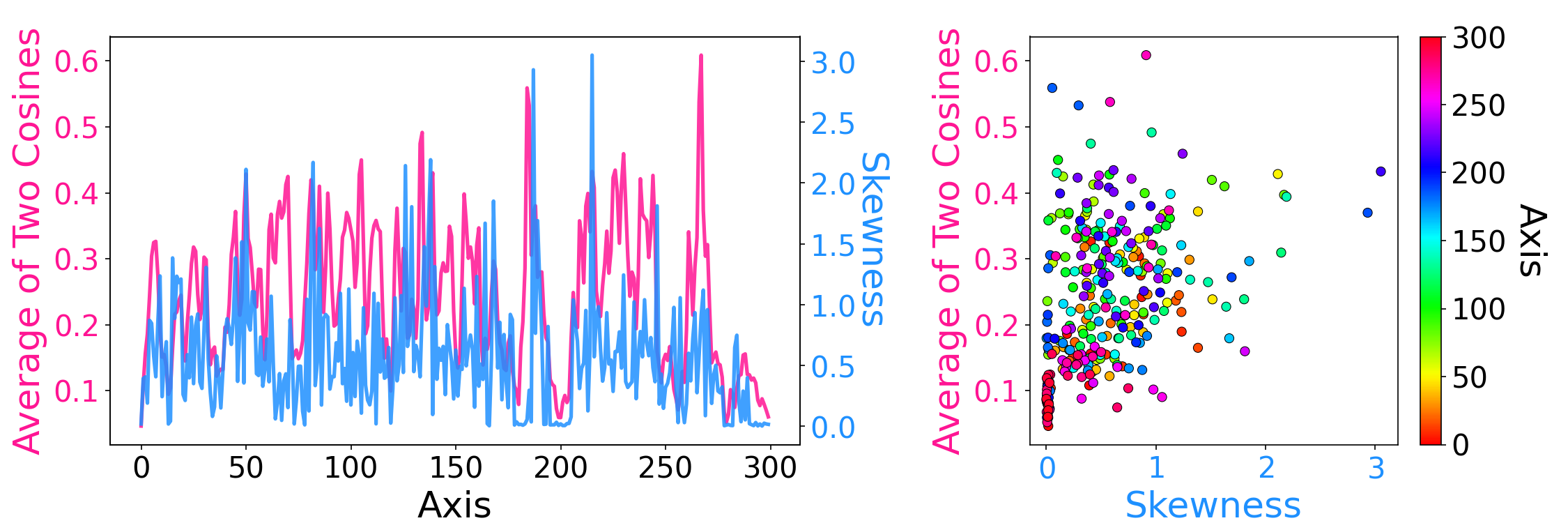}
    \subcaption{Axis Tour}
    \label{fig:cos-skew-plot-axistour}
\end{subfigure}
\begin{subfigure}{\columnwidth}
    \includegraphics[width=\textwidth]{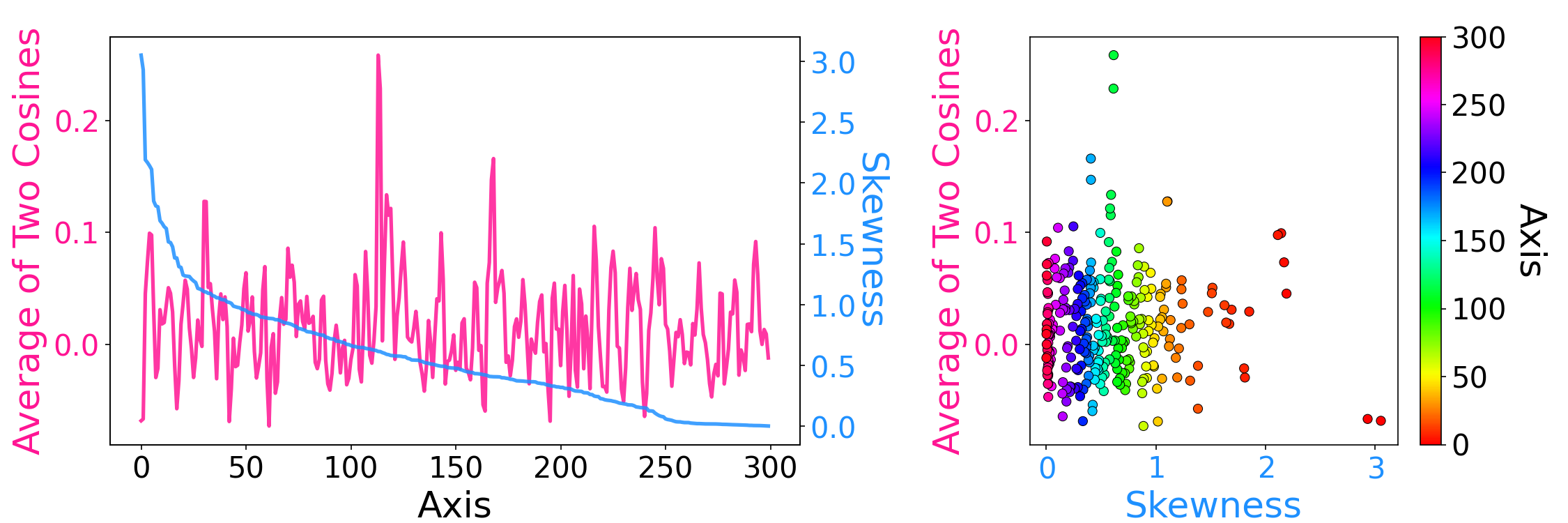}
    \subcaption{Skewness Sort}
    \label{fig:cos-skew-plot-skewnessort}
\end{subfigure}
    \caption{
Relationship between the skewness $\gamma_\ell$ and the average of two consecutive cosines $(\cos(\mathbf{v}_{\ell-1}, \mathbf{v}_\ell) + \cos(\mathbf{v}_\ell, \mathbf{v}_{\ell+1}))/2$ for all the axes $\ell=1,\ldots, d$ in (a) Axis Tour and (b) Skewness Sort. The left plot shows the skewness and the average of two cosines on both $y$-axes, while the right plot shows the scatter plot of these values. Spearman's rank correlation is 0.43 for Axis Tour, while it is 0.04 for Skewness Sort.
}
\label{fig:cos-skew-plot}
\end{figure}

\subsection{Dimensionality reduction: analogy, word similarity, and categorization tasks}\label{sec:downstream}
Using Word Embedding Benchmark~\cite{DBLP:journals/corr/JastrzebskiLC17}\footnote{\url{https://github.com/kudkudak/word-embeddings-benchmarks}.}, we evaluate the performance of dimensionality reduction in analogy, word similarity, and categorization tasks.
PCA selects the axes in descending order of eigenvalue. Random Order and Skewness Sort select the axes sequentially from first to last. 
Axis Tour adopts the dimensionality reduction\footnote{Fig.~\ref{fig:comp-alpha} in Appendix~\ref{app:comp-alpha} shows the results for different $\alpha$.} in Section~\ref{sec:dim-reduction} with $\alpha=1/3$.

We consider the original GloVe embeddings $\mathbf{X}$ as well as the embeddings obtained by applying POLAR to $\mathbf{X}$. 
POLAR is a method that uses pairs of words with opposite meanings and finds the axes where these words are positioned at the opposite ends. 
There are three methods for selecting the axes: Random Selection, Variance Maximization, and Orthogonality Maximization. 
We used the publicly available code for our experiments. 
For both the original and POLAR-applied embeddings, axes were selected sequentially from first to last.

Figure~\ref{fig:downstream} shows that the dimensionality reduction with the ordering in Axis Tour is better than or comparable to the baselines for most dimensionalities in each task. This result suggests that Axis Tour efficiently merges axes with similar meanings. 
For more details, refer to Appendix~\ref{app:experiments}.

\section{Discussion}
We confirmed through both qualitative and quantitative experiments that the axis order determined by Axis Tour exhibits high semantic continuity, and its effectiveness was also validated in the downstream task of dimensionality reduction.

There are two advantages to ordering the axes of ICA-transformed embeddings using Axis Tour. 
First, as shown in Fig.~\ref{fig:intro}, when projecting the embedding space, the scatterplot becomes easier to interpret and more visually accessible because higher-ranking words on each axis are farther from the origin. 
Second, although the Axis Tour embeddings $\mathbf{T}$ are the same size as the ICA-transformed embeddings $\mathbf{S}$, the axis order in Axis Tour preserves information about the similarities between the axes. 

An interesting relationship between semantic continuity and skewness is shown in Fig.~\ref{fig:cos-skew-plot}, where the semantic continuity of axis $\ell$ is measured by the average of two cosines.
In Axis Tour, there is a high correlation between skewness and semantic continuity, while in Skewness Sort, they are almost uncorrelated.
A large skewness indicates that an axis has a distinctive meaning, and in Axis Tour, this seems to contribute to semantic continuity.

\section{Conclusion}
In this study, we proposed a novel method, Axis Tour, which optimizes the order of axes in ICA-transformed word embeddings. We focused on the fact that the word embeddings reveal interpretable semantic axes while the order of these axes is arbitrary. Axis Tour aims to improve the clarity of the word embedding space by maximizing the semantic continuity of the axes. Additionally, we demonstrated through experiments on downstream tasks that Axis Tour yields better or comparable low-dimensional embeddings compared to both PCA and ICA.

\clearpage
\section*{Limitations}
\begin{itemize}
    \item While the dimension reduction experiments showed the improvement of the downstream task performance for the Axis Tour embeddings, there are three aspects that could be further improved:
    \begin{enumerate}
        \item Dimension reduction is performed using the vector $\mathbf{f}_r$, but its definition (\ref{eq:def-f}) is empirical, and better vectors may be designed. In addition, nonlinear transformations beyond linear ones could be considered for dimension reduction. Details on the definition of $\mathbf{f}_r$ can be found in Appendix~\ref{app:dim-reduction}.
        \item The method in Section~\ref{sec:dim-reduction} simply divides $[d]$ into $p$ equal intervals to merge the axes. However, adaptively determining the division points could allow selecting more semantically coherent groups of axes.
        \item To construct optimal low-dimensional vectors using ICA-transformed embeddings, applying clustering methods such as $K$-means to axis embeddings may improve performance. In this case, the overall optimized axis order may not be determined as in Axis Tour, but performing Axis Tour within each cluster and then concatenating these could determine an axis order depending on the number of clusters.
    \end{enumerate}
    However, this study focuses on a method for maximizing the semantic continuity of axes in ICA-transformed embeddings, leaving detailed investigation of the effective low-dimensional vector as future work.
    \item In Axis Tour, while adjacent axes may have similar meanings, axes with similar meanings may not be in close order.
This is due to the fact that in Word Tour, high-dimensional embeddings result in one-dimensional embeddings, and the meanings of words are similar when the word order is close, but semantically similar words are not always embedded close to each other. 
    \item As seen in Fig.~\ref{fig:intro}, projecting multiple axes of ICA-transformed embeddings into two dimensions can effectively represent the shape of the embeddings. However, as the number of axes increases, the angles between the axes become small, resulting in crowded axes. This can cause problems such as the top words of the axes being closer to the origin, which can be difficult to interpret.
    \item In Axis Tour, the dimension of the ICA-transformed embeddings corresponds to the number of cities in TSP. Therefore, as the dimension of the embeddings increases, the computation time for Axis Tour becomes longer. Note that for the 300-dimensional GloVe used in this study, the computation time for Axis Tour is about one second. For reference, Word Tour with $n = 40{,}000$ is known to take several hours\footnote{\url{https://github.com/joisino/wordtour}.}.
\end{itemize}

\section*{Ethics Statement}
A potential risk of this method is that we interpret the meanings of the axes of the ICA-transformed embeddings by the top words of each axis. If the embeddings contain personal information, such as email addresses or phone numbers, and these are contained in the top words, this can be problematic. Therefore, in this study, URLs, email addresses, and phone numbers were anonymized to avoid revealing such information.

\section*{Acknowledgements}
We would like to thank Momose Oyama for the discussion and anonymous reviewers for their helpful advice. This study was partially supported by JSPS KAKENHI 22H05106, 23H03355, JST CREST JPMJCR21N3, JST SPRING JPMJSP2110.

\section*{Code availability}
Our code is available at \url{https://github.com/ymgw55/Axis-Tour}.

\bibliography{custom}

\begin{thebibliography}{43}
\expandafter\ifx\csname natexlab\endcsname\relax\def\natexlab#1{#1}\fi

\bibitem[{Almuhareb and Poesio(2005)}]{cat-AP}
Abdulrahman Almuhareb and Massimo Poesio. 2005.
\newblock \href {https://escholarship.org/uc/item/8gh9h462} {Concept learning and categorization from the web}.
\newblock \emph{Proceedings of the Annual Meeting of the Cognitive Science Society}, 27.

\bibitem[{An et~al.(2018)An, Kwak, and Ahn}]{DBLP:conf/acl/AnKA18}
Jisun An, Haewoon Kwak, and Yong{-}Yeol Ahn. 2018.
\newblock \href {https://doi.org/10.18653/V1/P18-1228} {Semaxis: {A} lightweight framework to characterize domain-specific word semantics beyond sentiment}.
\newblock In \emph{Proceedings of the 56th Annual Meeting of the Association for Computational Linguistics, {ACL} 2018, Melbourne, Australia, July 15-20, 2018, Volume 1: Long Papers}, pages 2450--2461. Association for Computational Linguistics.

\bibitem[{Baroni et~al.(2008)Baroni, Evert, and Lenci}]{cat-ESSLLI}
Marco Baroni, Stefan Evert, and Alessandro Lenci, editors. 2008.
\newblock \href {http://wordspace.collocations.de/doku.php/workshop:esslli:proceedings} {\emph{Proceedings of the ESSLLI Workshop on Distributional Lexical Semantics: Bridging the Gap between Semantic Theory and Computational Simulations}}. European Summer School in Logic, Language and Information (ESSLLI), Hamburg, Germany.

\bibitem[{Baroni and Lenci(2011)}]{cat-BLESS}
Marco Baroni and Alessandro Lenci. 2011.
\newblock \href {https://aclanthology.org/W11-2501/} {How we blessed distributional semantic evaluation}.
\newblock In \emph{Proceedings of the {GEMS} 2011 Workshop on GEometrical Models of Natural Language Semantics, Edinburgh, UK, July 31, 2011}, pages 1--10. Association for Computational Linguistics.

\bibitem[{Battig and Montague(1969)}]{cat-Battig}
William~F. Battig and William~E. Montague. 1969.
\newblock \href {https://psycnet.apa.org/doiLanding?doi=10.1037%2Fh0027577} {Category norms of verbal items in 56 categories: A replication and extension of the connecticut category norms}.
\newblock \emph{Journal of Experimental Psychology}, 80(3, Pt.2):1--46.

\bibitem[{Billingsley(1995)}]{billingsley1995probability}
Patrick Billingsley. 1995.
\newblock \emph{Probability and Measure, Third Edition}.
\newblock Wiley Series in Probability and Statistics. Wiley.

\bibitem[{Browne(2001)}]{browne2001overview}
Michael~W Browne. 2001.
\newblock An overview of analytic rotation in exploratory factor analysis.
\newblock \emph{Multivariate behavioral research}, 36(1):111--150.

\bibitem[{Bruni et~al.(2014)Bruni, Tran, and Baroni}]{ws-MEN}
Elia Bruni, Nam-Khanh Tran, and Marco Baroni. 2014.
\newblock Multimodal distributional semantics.
\newblock \emph{Journal of Artificial Intelligence Research}, 49:1--47.

\bibitem[{Chelba et~al.(2014)Chelba, Mikolov, Schuster, Ge, Brants, Koehn, and Robinson}]{DBLP:conf/interspeech/ChelbaMSGBKR14}
Ciprian Chelba, Tom{\'{a}}s Mikolov, Mike Schuster, Qi~Ge, Thorsten Brants, Phillipp Koehn, and Tony Robinson. 2014.
\newblock \href {http://www.isca-speech.org/archive/interspeech\_2014/i14\_2635.html} {One billion word benchmark for measuring progress in statistical language modeling}.
\newblock In \emph{{INTERSPEECH} 2014, 15th Annual Conference of the International Speech Communication Association, Singapore, September 14-18, 2014}, pages 2635--2639. {ISCA}.

\bibitem[{Crawford and Ferguson(1970)}]{crawford1970general}
Charles~B Crawford and George~A Ferguson. 1970.
\newblock A general rotation criterion and its use in orthogonal rotation.
\newblock \emph{Psychometrika}, 35(3):321--332.

\bibitem[{Cunningham et~al.(2023)Cunningham, Ewart, Riggs, Huben, and Sharkey}]{DBLP:journals/corr/abs-2309-08600}
Hoagy Cunningham, Aidan Ewart, Logan Riggs, Robert Huben, and Lee Sharkey. 2023.
\newblock \href {https://doi.org/10.48550/ARXIV.2309.08600} {Sparse autoencoders find highly interpretable features in language models}.
\newblock \emph{CoRR}, abs/2309.08600.

\bibitem[{Devlin et~al.(2019)Devlin, Chang, Lee, and Toutanova}]{DBLP:conf/naacl/DevlinCLT19}
Jacob Devlin, Ming{-}Wei Chang, Kenton Lee, and Kristina Toutanova. 2019.
\newblock \href {https://doi.org/10.18653/V1/N19-1423} {{BERT:} pre-training of deep bidirectional transformers for language understanding}.
\newblock In \emph{Proceedings of the 2019 Conference of the North American Chapter of the Association for Computational Linguistics: Human Language Technologies, {NAACL-HLT} 2019, Minneapolis, MN, USA, June 2-7, 2019, Volume 1 (Long and Short Papers)}, pages 4171--4186. Association for Computational Linguistics.

\bibitem[{Engler et~al.(2022)Engler, Sikdar, Lutz, and Strohmaier}]{DBLP:conf/emnlp/EnglerSLS22}
Jan Engler, Sandipan Sikdar, Marlene Lutz, and Markus Strohmaier. 2022.
\newblock \href {https://doi.org/10.18653/V1/2022.FINDINGS-EMNLP.338} {Sensepolar: Word sense aware interpretability for pre-trained contextual word embeddings}.
\newblock In \emph{Findings of the Association for Computational Linguistics: {EMNLP} 2022, Abu Dhabi, United Arab Emirates, December 7-11, 2022}, pages 4607--4619. Association for Computational Linguistics.

\bibitem[{Finkelstein et~al.(2002)Finkelstein, Gabrilovich, Matias, Rivlin, Solan, Wolfman, and Ruppin}]{ws-WS353}
Lev Finkelstein, Evgeniy Gabrilovich, Yossi Matias, Ehud Rivlin, Zach Solan, Gadi Wolfman, and Eytan Ruppin. 2002.
\newblock Placing search in context: The concept revisited.
\newblock \emph{ACM Transactions on information systems}, 20(1):116--131.

\bibitem[{Helsgaun(2000)}]{helsgaun2000effective}
Keld Helsgaun. 2000.
\newblock \href {https://www.sciencedirect.com/science/article/abs/pii/S0377221799002842} {An effective implementation of the lin-kernighan traveling salesman heuristic}.
\newblock \emph{Eur. J. Oper. Res.}, 126(1):106--130.

\bibitem[{Helsgaun(2018)}]{lkh_webpage}
Keld Helsgaun. 2018.
\newblock \href {http://webhotel4.ruc.dk/~keld/research/LKH/} {{LKH} ({Keld Helsgaun})}.

\bibitem[{Hill et~al.(2015)Hill, Reichart, and Korhonen}]{ws-SimLex999}
Felix Hill, Roi Reichart, and Anna Korhonen. 2015.
\newblock Simlex-999: Evaluating semantic models with (genuine) similarity estimation.
\newblock \emph{Computational Linguistics}, 41(4):665--695.

\bibitem[{Hyv{\"{a}}rinen(1999)}]{DBLP:journals/tnn/Hyvarinen99}
Aapo Hyv{\"{a}}rinen. 1999.
\newblock \href {https://doi.org/10.1109/72.761722} {Fast and robust fixed-point algorithms for independent component analysis}.
\newblock \emph{{IEEE} Trans. Neural Networks}, 10(3):626--634.

\bibitem[{Hyv{\"{a}}rinen et~al.(2001{\natexlab{a}})Hyv{\"{a}}rinen, Hoyer, and Inki}]{DBLP:journals/neco/HyvarinenHI01}
Aapo Hyv{\"{a}}rinen, Patrik~O. Hoyer, and Mika Inki. 2001{\natexlab{a}}.
\newblock \href {https://doi.org/10.1162/089976601750264992} {Topographic independent component analysis}.
\newblock \emph{Neural Comput.}, 13(7):1527--1558.

\bibitem[{Hyv{\"{a}}rinen et~al.(2001{\natexlab{b}})Hyv{\"{a}}rinen, Karhunen, and Oja}]{DBLP:books/wi/HyvarinenKO01}
Aapo Hyv{\"{a}}rinen, Juha Karhunen, and Erkki Oja. 2001{\natexlab{b}}.
\newblock \href {https://doi.org/10.1002/0471221317} {\emph{Independent Component Analysis}}.
\newblock Wiley.

\bibitem[{Hyv{\"{a}}rinen and Oja(2000)}]{DBLP:journals/nn/HyvarinenO00}
Aapo Hyv{\"{a}}rinen and Erkki Oja. 2000.
\newblock \href {https://doi.org/10.1016/S0893-6080(00)00026-5} {Independent component analysis: algorithms and applications}.
\newblock \emph{Neural Networks}, 13(4-5):411--430.

\bibitem[{Jastrzebski et~al.(2017)Jastrzebski, Lesniak, and Czarnecki}]{DBLP:journals/corr/JastrzebskiLC17}
Stanislaw Jastrzebski, Damian Lesniak, and Wojciech~Marian Czarnecki. 2017.
\newblock \href {http://arxiv.org/abs/1702.02170} {How to evaluate word embeddings? on importance of data efficiency and simple supervised tasks}.
\newblock \emph{CoRR}, abs/1702.02170.

\bibitem[{Kohonen(2001)}]{DBLP:books/sp/Kohonen01}
Teuvo Kohonen. 2001.
\newblock \href {https://doi.org/10.1007/978-3-642-56927-2} {\emph{Self-Organizing Maps, Third Edition}}.
\newblock Springer Series in Information Sciences. Springer.

\bibitem[{Kwak et~al.(2021)Kwak, An, Jing, and Ahn}]{DBLP:journals/peerj-cs/KwakAJA21}
Haewoon Kwak, Jisun An, Elise Jing, and Yong{-}Yeol Ahn. 2021.
\newblock \href {https://doi.org/10.7717/PEERJ-CS.644} {Frameaxis: characterizing microframe bias and intensity with word embedding}.
\newblock \emph{PeerJ Comput. Sci.}, 7:e644.

\bibitem[{Lin and Kernighan(1973)}]{lin1973effective}
Shen Lin and Brian~W. Kernighan. 1973.
\newblock \href {https://pubsonline.informs.org/doi/10.1287/opre.21.2.498} {An effective heuristic algorithm for the traveling-salesman problem}.
\newblock \emph{Oper. Res.}, 21(2):498--516.

\bibitem[{Luong et~al.(2013)Luong, Socher, and Manning}]{ws-RW}
Thang Luong, Richard Socher, and Christopher Manning. 2013.
\newblock Better word representations with recursive neural networks for morphology.
\newblock In \emph{Proceedings of the Seventeenth Conference on Computational Natural Language Learning}, pages 104--113.

\bibitem[{Mare{\v{c}}ek et~al.(2020)Mare{\v{c}}ek, Libovick{\'{y}}, Musil, Rosa, and Limisiewicz}]{marecek-etal-2020}
David Mare{\v{c}}ek, Jind{\v{r}}ich Libovick{\'{y}}, Tom{\'{a}}{\v{s}} Musil, Rudolf Rosa, and Tomasz Limisiewicz. 2020.
\newblock \href {https://ufal.mff.cuni.cz/books/2020-marecek} {\emph{Hidden in the Layers: Interpretation of Neural Networks for Natural Language Processing}}, volume~20 of \emph{Studies in Computational and Theoretical Linguistics}.
\newblock Institute of Formal and Applied Linguistics, Prague, Czechia.

\bibitem[{Mathew et~al.(2020)Mathew, Sikdar, Lemmerich, and Strohmaier}]{DBLP:conf/www/MathewSLS20}
Binny Mathew, Sandipan Sikdar, Florian Lemmerich, and Markus Strohmaier. 2020.
\newblock \href {https://doi.org/10.1145/3366423.3380227} {The {POLAR} framework: Polar opposites enable interpretability of pre-trained word embeddings}.
\newblock In \emph{{WWW} '20: The Web Conference 2020, Taipei, Taiwan, April 20-24, 2020}, pages 1548--1558. {ACM} / {IW3C2}.

\bibitem[{Mikolov et~al.(2013{\natexlab{a}})Mikolov, Chen, Corrado, and Dean}]{analogy-google}
Tom{\'{a}}s Mikolov, Kai Chen, Greg Corrado, and Jeffrey Dean. 2013{\natexlab{a}}.
\newblock \href {http://arxiv.org/abs/1301.3781} {Efficient estimation of word representations in vector space}.
\newblock In \emph{1st International Conference on Learning Representations, {ICLR} 2013, Scottsdale, Arizona, USA, May 2-4, 2013, Workshop Track Proceedings}.

\bibitem[{Mikolov et~al.(2013{\natexlab{b}})Mikolov, Chen, Corrado, and Dean}]{DBLP:journals/corr/abs-1301-3781}
Tom{\'{a}}s Mikolov, Kai Chen, Greg Corrado, and Jeffrey Dean. 2013{\natexlab{b}}.
\newblock \href {http://arxiv.org/abs/1301.3781} {Efficient estimation of word representations in vector space}.
\newblock In \emph{1st International Conference on Learning Representations, {ICLR} 2013, Scottsdale, Arizona, USA, May 2-4, 2013, Workshop Track Proceedings}.

\bibitem[{Mikolov et~al.(2013{\natexlab{c}})Mikolov, Yih, and Zweig}]{analogy-msr}
Tom{\'{a}}s Mikolov, Wen{-}tau Yih, and Geoffrey Zweig. 2013{\natexlab{c}}.
\newblock \href {https://aclanthology.org/N13-1090/} {Linguistic regularities in continuous space word representations}.
\newblock In \emph{Human Language Technologies: Conference of the North American Chapter of the Association of Computational Linguistics, Proceedings, June 9-14, 2013, Westin Peachtree Plaza Hotel, Atlanta, Georgia, {USA}}, pages 746--751. The Association for Computational Linguistics.

\bibitem[{Musil(2019)}]{DBLP:conf/tsd/Musil19}
Tom{\'{a}}s Musil. 2019.
\newblock \href {https://doi.org/10.1007/978-3-030-27947-9\_18} {Examining structure of word embeddings with {PCA}}.
\newblock In \emph{Text, Speech, and Dialogue - 22nd International Conference, {TSD} 2019, Ljubljana, Slovenia, September 11-13, 2019, Proceedings}, volume 11697 of \emph{Lecture Notes in Computer Science}, pages 211--223. Springer.

\bibitem[{Musil and Mare{\v{c}}ek(2024)}]{musil-marecek-2024-exploring}
Tom{\'a}{\v{s}} Musil and David Mare{\v{c}}ek. 2024.
\newblock \href {https://aclanthology.org/2024.lrec-main.605} {Exploring interpretability of independent components of word embeddings with automated word intruder test}.
\newblock In \emph{Proceedings of the 2024 Joint International Conference on Computational Linguistics, Language Resources and Evaluation (LREC-COLING 2024)}, pages 6922--6928, Torino, Italia. ELRA and ICCL.

\bibitem[{Park et~al.(2017)Park, Bak, and Oh}]{DBLP:conf/emnlp/ParkBO17}
Sungjoon Park, JinYeong Bak, and Alice Oh. 2017.
\newblock \href {https://doi.org/10.18653/V1/D17-1041} {Rotated word vector representations and their interpretability}.
\newblock In \emph{Proceedings of the 2017 Conference on Empirical Methods in Natural Language Processing, {EMNLP} 2017, Copenhagen, Denmark, September 9-11, 2017}, pages 401--411. Association for Computational Linguistics.

\bibitem[{Pedregosa et~al.(2011)Pedregosa, Varoquaux, Gramfort, Michel, Thirion, Grisel, Blondel, Prettenhofer, Weiss, Dubourg, VanderPlas, Passos, Cournapeau, Brucher, Perrot, and Duchesnay}]{DBLP:journals/jmlr/PedregosaVGMTGBPWDVPCBPD11}
Fabian Pedregosa, Ga{\"{e}}l Varoquaux, Alexandre Gramfort, Vincent Michel, Bertrand Thirion, Olivier Grisel, Mathieu Blondel, Peter Prettenhofer, Ron Weiss, Vincent Dubourg, Jake VanderPlas, Alexandre Passos, David Cournapeau, Matthieu Brucher, Matthieu Perrot, and Edouard Duchesnay. 2011.
\newblock \href {https://doi.org/10.5555/1953048.2078195} {Scikit-learn: Machine learning in python}.
\newblock \emph{J. Mach. Learn. Res.}, 12:2825--2830.

\bibitem[{Pennington et~al.(2014)Pennington, Socher, and Manning}]{DBLP:conf/emnlp/PenningtonSM14}
Jeffrey Pennington, Richard Socher, and Christopher~D. Manning. 2014.
\newblock \href {https://doi.org/10.3115/V1/D14-1162} {Glove: Global vectors for word representation}.
\newblock In \emph{Proceedings of the 2014 Conference on Empirical Methods in Natural Language Processing, {EMNLP} 2014, October 25-29, 2014, Doha, Qatar, {A} meeting of SIGDAT, a Special Interest Group of the {ACL}}, pages 1532--1543. {ACL}.

\bibitem[{Radinsky et~al.(2011)Radinsky, Agichtein, Gabrilovich, and Markovitch}]{ws-MTurk}
Kira Radinsky, Eugene Agichtein, Evgeniy Gabrilovich, and Shaul Markovitch. 2011.
\newblock A word at a time: Computing word relatedness using temporal semantic analysis.
\newblock In \emph{Proceedings of the 20th International Conference on World Wide Web}, page 337–346.

\bibitem[{Rubenstein and Goodenough(1965)}]{ws-RG65}
Herbert Rubenstein and John~B. Goodenough. 1965.
\newblock \href {https://doi.org/10.1145/365628.365657} {Contextual correlates of synonymy}.
\newblock \emph{Commun. {ACM}}, 8(10):627--633.

\bibitem[{Sato(2022)}]{DBLP:conf/naacl/Sato22}
Ryoma Sato. 2022.
\newblock \href {https://doi.org/10.18653/V1/2022.NAACL-MAIN.157} {Word tour: One-dimensional word embeddings via the traveling salesman problem}.
\newblock In \emph{Proceedings of the 2022 Conference of the North American Chapter of the Association for Computational Linguistics: Human Language Technologies, {NAACL} 2022, Seattle, WA, United States, July 10-15, 2022}, pages 2166--2172. Association for Computational Linguistics.

\bibitem[{Senel et~al.(2022)Senel, Sahinu{\c{c}}, Y{\"{u}}cesoy, Sch{\"{u}}tze, {\c{C}}ukur, and Ko{\c{c}}}]{DBLP:journals/ipm/SenelSYSCK22}
L{\"{u}}tfi~Kerem Senel, Furkan Sahinu{\c{c}}, Veysel Y{\"{u}}cesoy, Hinrich Sch{\"{u}}tze, Tolga {\c{C}}ukur, and Aykut Ko{\c{c}}. 2022.
\newblock \href {https://doi.org/10.1016/J.IPM.2022.102925} {Learning interpretable word embeddings via bidirectional alignment of dimensions with semantic concepts}.
\newblock \emph{Inf. Process. Manag.}, 59(3):102925.

\bibitem[{Speer(2022)}]{robyn_speer_2022_7199437}
Robyn Speer. 2022.
\newblock \href {https://doi.org/10.5281/zenodo.7199437} {rspeer/wordfreq: v3.0}.

\bibitem[{Wolf et~al.(2020)Wolf, Debut, Sanh, Chaumond, Delangue, Moi, Cistac, Rault, Louf, Funtowicz, Davison, Shleifer, von Platen, Ma, Jernite, Plu, Xu, Scao, Gugger, Drame, Lhoest, and Rush}]{DBLP:conf/emnlp/WolfDSCDMCRLFDS20}
Thomas Wolf, Lysandre Debut, Victor Sanh, Julien Chaumond, Clement Delangue, Anthony Moi, Pierric Cistac, Tim Rault, R{\'{e}}mi Louf, Morgan Funtowicz, Joe Davison, Sam Shleifer, Patrick von Platen, Clara Ma, Yacine Jernite, Julien Plu, Canwen Xu, Teven~Le Scao, Sylvain Gugger, Mariama Drame, Quentin Lhoest, and Alexander~M. Rush. 2020.
\newblock \href {https://doi.org/10.18653/V1/2020.EMNLP-DEMOS.6} {Transformers: State-of-the-art natural language processing}.
\newblock In \emph{Proceedings of the 2020 Conference on Empirical Methods in Natural Language Processing: System Demonstrations, {EMNLP} 2020 - Demos, Online, November 16-20, 2020}, pages 38--45. Association for Computational Linguistics.

\bibitem[{Yamagiwa et~al.(2023)Yamagiwa, Oyama, and Shimodaira}]{DBLP:conf/emnlp/YamagiwaOS23}
Hiroaki Yamagiwa, Momose Oyama, and Hidetoshi Shimodaira. 2023.
\newblock \href {https://aclanthology.org/2023.emnlp-main.283} {Discovering universal geometry in embeddings with {ICA}}.
\newblock In \emph{Proceedings of the 2023 Conference on Empirical Methods in Natural Language Processing, {EMNLP} 2023, Singapore, December 6-10, 2023}, pages 4647--4675. Association for Computational Linguistics.

\end{thebibliography}
\bibliographystyle{acl_natbib}

\appendix
\section{Details of scatterplots from two-dimensional projection}\label{app:fig-explanation}
This section explains the two-dimensional projection method used for the scatterplots in Fig.~\ref{fig:intro}. We then present similar scatterplots for the three examples in Table~\ref{tab:examples}. Finally, we define the metrics used in Section~\ref{sec:intro} to evaluate the quality of the scatterplots.

\subsection{Scatterplot drawing method}\label{app:fig-process}
We will explain the scatterplot drawing method using the Axis Tour embeddings.

First, we define a set of axis indices for projection. Let $I :=\{a,\ldots,b\}\,(a,b\in [d], a < b)$ be a consecutive interval of indices. The number of indices in $I$ is $|I|=b-a+1$. For example, in Fig.~\ref{fig:intro}, $a=86$ and $b=94$.

Next, we define a matrix by extracting the axes of $\hat{\mathbf{T}}$ corresponding to $I$, where $\hat{\mathbf{T}}$ is the matrix obtained by normalizing the matrix $\mathbf{T}$. We denote this extracted matrix as $\hat{\mathbf{T}}_I\in\mathbb{R}^{n\times|I|}$.

We consider the two-dimensional projection of $\hat{\mathbf{T}}_I$. 
For this projection, we define the matrix $\mathbf{P}_I\in\mathbb{R}^{|I|\times2}$ as follows\footnote{$I$ is a subinterval of $d$ indices, and to prevent angles between $\varphi_a$ and $\varphi_b$ from becoming smaller, $\theta_\ell$ is defined so that $\theta_a=0$ and $\theta_b=\pi$. If $\theta_\ell = \frac{2(\ell-a)\pi}{b-a+1}$, as we see in Fig. 15 in Appendix C of~\citet{DBLP:conf/emnlp/YamagiwaOS23}, the angle between $\varphi_a$ and $\varphi_b$ will be the same as between $\varphi_\ell$ and $\varphi_{\ell+1}$.}:
\begin{align}
\mathbf{P}_I := \begin{bmatrix}
\varphi_a^\top\\
\vdots\\
\varphi_b^\top
\end{bmatrix}\in\mathbb{R}^{|I|\times2},
\end{align}
where
\begin{align}
    \varphi_\ell := (\cos{\theta_\ell},\sin{\theta_\ell})^\top\in\mathbb{R}^2,
\end{align}
where
\begin{align}
    \theta_\ell := \frac{(\ell-a)\pi}{b-a}.
\end{align}
Then we get the two-dimensional projection as $\mathbf{Q}_I:=\hat{\mathbf{T}}_I\mathbf{P}_I\in\mathbb{R}^{n\times2}$. 
In $\mathbf{Q}_I$, the $\ell$-th axis of $\hat{\mathbf{T}}$ is projected along the direction of $\varphi_\ell$.

We denote the $i$-th word embedding of $\mathbf{Q}_I$ by $\mathbf{q}_i = (q_i^x, q_i^y)^\top\in\mathbb{R}^2$. 
When we plot the scatterplot of $\mathbf{Q}_I$, we do not plot the $i$-th word embedding if $q_i^y < 0$ for visual clarity, and show the top five words for each axis. 
The indices of these top words equal to the following index set $\text{Show}_I$ defined with $\text{Top}_k^\ell$ from section~\ref{sec:def-v}:
\begin{align}
    \text{Show}_I := \{i\in [n]\mid q_i^y \geq 0\} \cap \bigcup_{\ell\in I}\text{Top}_5^\ell.\label{eq:show_set}
\end{align}

Similarly, we can apply the same procedure to the Skewness Sort embeddings and obtain the two-dimensional scatterplot.

\subsection{Scatterplots of Table~\ref{tab:examples}}
\begin{figure*}[!t]
    \centering
    \begin{minipage}{0.33\linewidth}
        \centering
        \includegraphics[width=\linewidth]{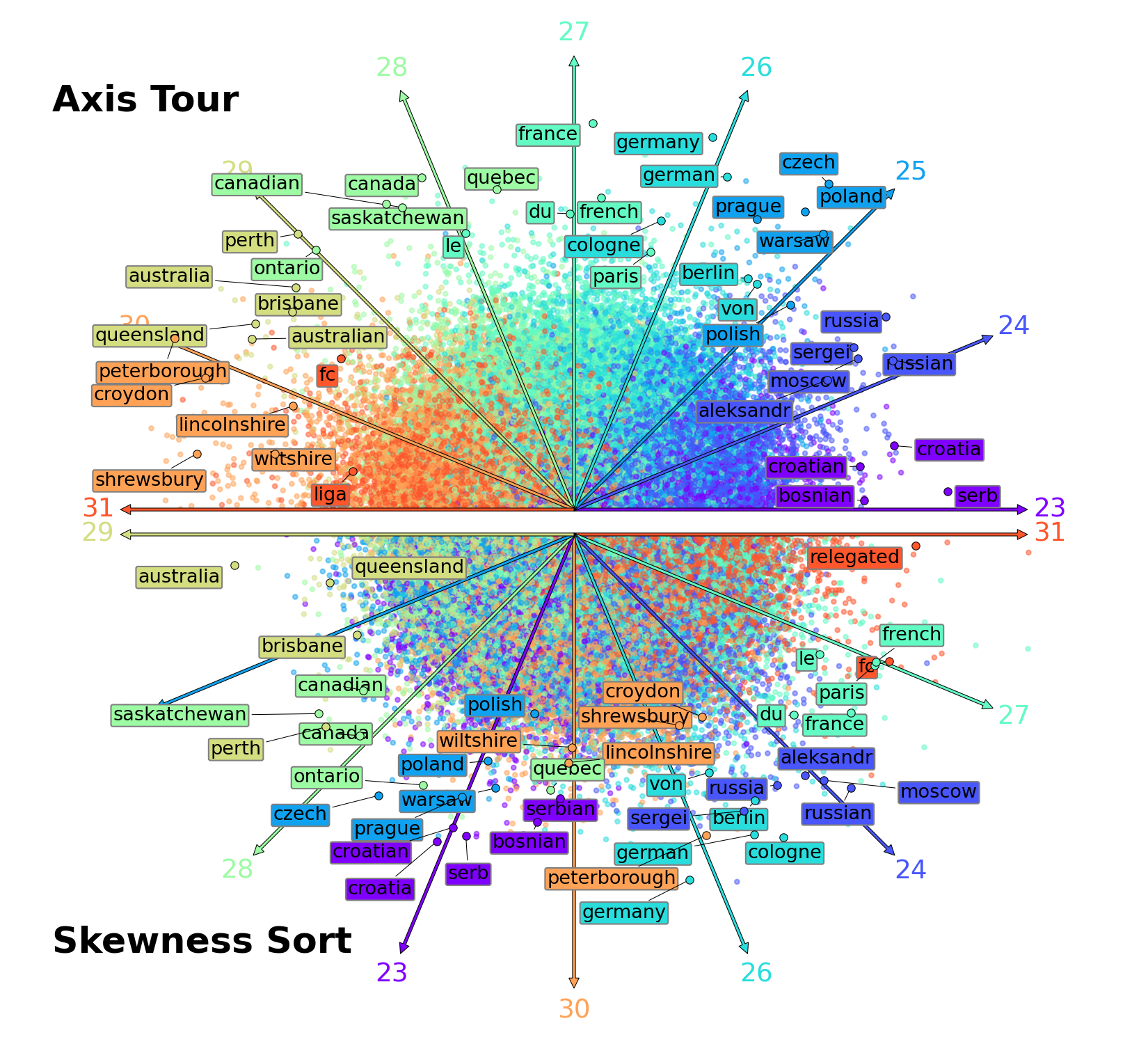}
        \subcaption{The 23rd axis to the 31st axis}
        \label{fig:axis27}
    \end{minipage}\hfill
    \begin{minipage}{0.33\linewidth}
        \centering
        \includegraphics[width=\linewidth]{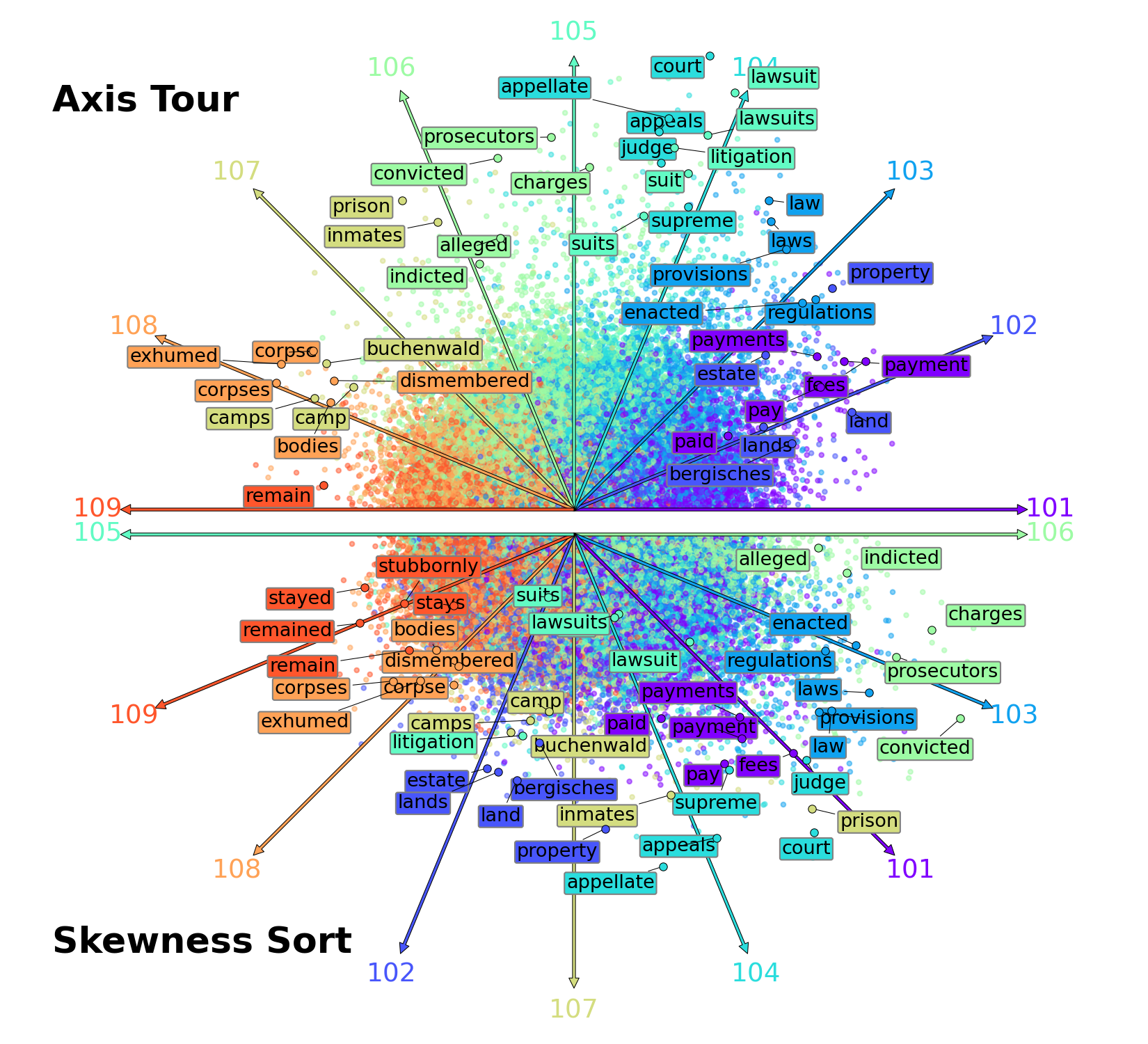}
        \subcaption{The 101st axis to the 109th axis}
        \label{fig:axis105}
    \end{minipage}
    \begin{minipage}{0.33\linewidth}
        \centering
        \includegraphics[width=\linewidth]{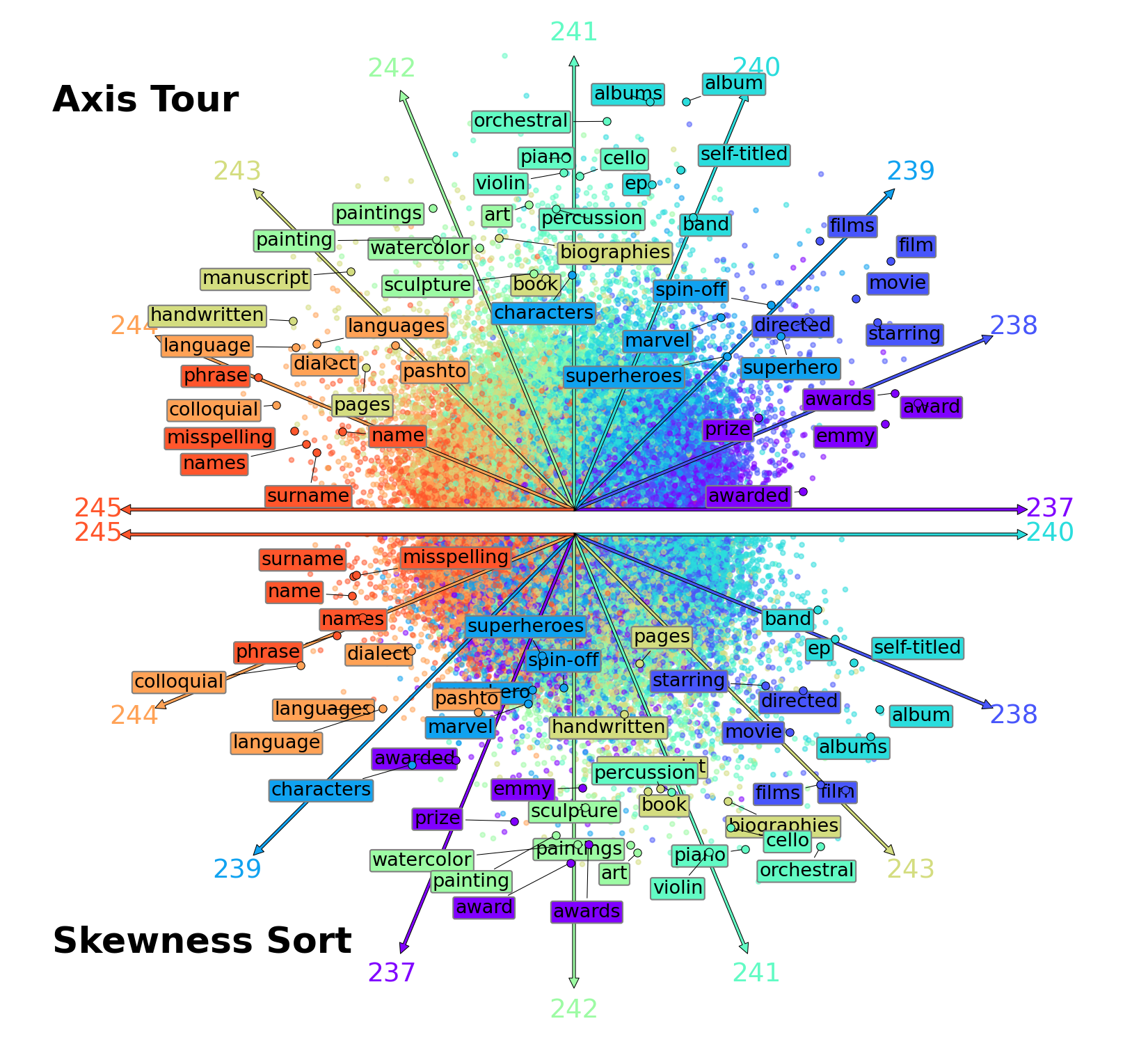}
        \subcaption{The 237th axis to the 245th axis}
        \label{fig:axis241}
    \end{minipage}
    \caption{
The scatterplots of the two-dimensional projections for the axes of the Axis Tour embeddings in Table~\ref{tab:examples}. 
Similar to Fig.~\ref{fig:intro}, we used the procedure described in Appendix~\ref{app:fig-process}.
}
\label{fig:examples}
\end{figure*}
Figure~\ref{fig:examples} shows the scatterplots of the two-dimensional projections for the axes of the Axis Tour embeddings in Table~\ref{tab:examples}, using the procedure described in Appendix~\ref{app:fig-process}. 
Similar to Fig.~\ref{fig:intro}, it is evident that the top words of the axes of the Axis Tour embeddings are farther from the origin than those of the Skewness Sort, and the meanings of the adjacent axes change continuously.

\subsection{Evaluation metrics for scatterplots}\label{app:evalmetric}
In Section~\ref{sec:intro}, we compared the quality of the scatterplots for Axis Tour and Skewness Sort by calculating the average distance of the top words from the origin. 
In this section, we first explain this metric and then, based on~(\ref{eq:axistour}), define a new metric derived from the average of the cosine similarities between adjacent axis embeddings. We then compare these metrics for the scatterplots in Figs.~\ref{fig:intro} and~\ref{fig:examples}. 
Similar to Appendix~\ref{app:fig-process}, we use the Axis Tour embeddings to explain these metrics.

For the two-dimensional projection $\mathbf{Q}_I$ of the Axis Tour embeddings, we define the average distance $d_I$ of the top words from the origin, using the index set $\text{Show}_I$ in (\ref{eq:show_set}), as follows:
\begin{align}
    d_I := \frac{1}{|\text{Show}_I|}\sum_{i\in\text{Show}_I}\|\mathbf{q}_i\|\label{eq:d-I}
\end{align}
A larger $d_I$ indicates that the scatterplot more accurately reflects the spatial distribution of the original embeddings, since the top words are positioned far from the origin.

We also define the average cosine similarity between adjacent axis embeddings, $c_I$, for the interval $I$ as follows:
\begin{align}
    c_I := \frac{1}{b-a}\sum_{\ell=a}^{b-1}\cos{(\mathbf{v}_\ell,\mathbf{v}_{\ell+1})}\label{eq:c-I}
\end{align}
As we saw in (\ref{eq:axistour}), since Axis Tour optimizes the order of the axes to maximize the sum of the cosine similarities, the value of $c_I$ reflects the semantic continuity of the axes in the scatterplot. It is important to note that $I$ represents a subinterval of $d$ indices, so we do not include the term of $\cos{(\mathbf{v}_a, \mathbf{v}_b)}$ in (\ref{eq:c-I}).

Both $d_I$ and $c_I$ can also be calculated in a similar manner for Skewness Sort. Table~\ref{tab:fig-results} shows the values of $d_I$ and $c_I$ for Figs.~\ref{fig:intro} and~\ref{fig:examples}. 
In these examples, Axis Tour shows higher values for both $d_I$ and $c_I$ compared to Skewness Sort, indicating better projection quality.

In addition to the results for several subintervals in Table~\ref{tab:fig-results}, we also see the semantic continuity of all axes. If $I = [d]$ in (\ref{eq:c-I}), we get $c_I = c_{[d]}$ as follows
\begin{align}
    c_{[d]} = \frac{1}{d}\left\{\cos{(\mathbf{v}_1,\mathbf{v}_{d})}+\sum_{\ell=1}^{d-1}\cos{(\mathbf{v}_\ell,\mathbf{v}_{\ell+1})}\right\}\label{eq:c-d}
\end{align}
where the term of $\cos{(\mathbf{v}_1, \mathbf{v}_{d})}$ is added according to (\ref{eq:axistour}). 
The value of $c_{[d]}$ is equal to the average of the histogram in Fig.~\ref{fig:cos-hist}, so the value for Axis Tour is $0.244$ and for Skewness Sort is $0.017$, thus confirming the semantic continuity across all axes.

\begin{table}[t]
\centering
\begin{tabular}{lrrrr}
\toprule
& \multicolumn{2}{c}{Axis Tour} &  \multicolumn{2}{c}{Skewness Sort} \\
\cmidrule(lr){2-3}
\cmidrule(lr){4-5}
Fig. &  $d_I$&  $c_I$  & $d_I$ &  $c_I$ \\
\midrule
\ref{fig:intro} & 0.76 &   0.27    & 0.61 &   0.18 \\
\ref{fig:axis27} &  0.69&        0.28    &  0.61&    0.10   \\
\ref{fig:axis105} &   0.80  &  0.30 &   0.62          & 0.15   \\
\ref{fig:axis241}&   0.86   &  0.38   &  0.76   &  0.23   \\
\bottomrule
\end{tabular}
\caption{
The values of $d_I$ and $c_I$ for Figs.~\ref{fig:intro} and~\ref{fig:examples}.
}
\label{tab:fig-results}
\end{table}

\section{Dimensionality reduction}\label{app:dim-reduction}
This section details and supplements Section~\ref{sec:dim-reduction}.

\subsection{Definition of $\mathbf{f}_r$}\label{app:def-f}
Here we first discuss the definition of $\mathbf{f}_r$ in terms of skewness and then explain the normalization.

\subsubsection{Skewness as weight}
In (\ref{eq:axis-embedding}), a vector $\mathbf{f}_r=(f_r^{(\ell)})_{\ell=1}^d\in\mathbb{R}_{\geq 0}^d$ for dimensionality reduction is defined by the skewness $\gamma_\ell \in \mathbb{R}_{\geq 0}$ of the $\ell$-th axis of $\mathbf{T}$. In particular, $f_r^{(\ell)}$, which corresponds to the weight of the $\ell$-th axis in the projection, is proportional to $\gamma_\ell^\alpha\,(\alpha\in \mathbb{R}_{\geq 0})$. 

This is based on the assumption that the axis becomes more meaningful as the skewness increases\footnote{While not specific to ICA-transformed embeddings, it is known from a study of sparse coding for language models that skewness correlates with interpretability~\cite{DBLP:journals/corr/abs-2309-08600}.}. Higher-order statistics such as skewness are known to be sensitive to outliers~\cite{DBLP:books/wi/HyvarinenKO01}, so we mitigate this effect by raising $\gamma_\ell$ to the power of $\alpha$. Since skewness is a third-order moment, we treat $\alpha = 1/3$ as the default value. $\alpha = 0$ is a uniform weight, while $\alpha = 1$ is the original $\gamma_\ell$. Next, we will explain why $\mathbf{f}_r$ is normalized.

\subsubsection{Reason for Normalizing $\mathbf{f}_r$}
$\mathbf{f}_r$ is normalized (i.e., $\|\mathbf{f}_r\|=1$) to equalize the scale of the one-dimensional projections for each subspace. Thus, when $p=d$, $\mathbf{T}\mathbf{F}=\mathbf{T}$ holds.

For $p=d$, $I_r=\{r\}$, and from (\ref{eq:axis-embedding}), $\mathbf{f}_r=(f_r^{(\ell)})_{\ell=1}^d$ satisfies:
\begin{align}
   f_r^{(\ell)}  = \delta_{\ell r} = \begin{cases} 
   1 & \ell=r \\
   0 & \ell\neq r
   \end{cases}.\label{eq:def-f2}
\end{align}
Here, $\delta_{\ell r}$ is the Dirac delta function. From (\ref{eq:def-f2}) the matrix $\mathbf{F}=[\mathbf{f}_1,\ldots,\mathbf{f}_d]\in\mathbb{R}^{d\times d}$ corresponds to the $d$-dimensional identity matrix $\mathbf{I}\in\mathbb{R}^{d\times d}$, so $\mathbf{T}\mathbf{F}=\mathbf{T}\mathbf{I}=\mathbf{T}$.

\subsection{Projection from subspace to one-dimensional space}
\begin{figure}
\centering
\includegraphics[width=\columnwidth]{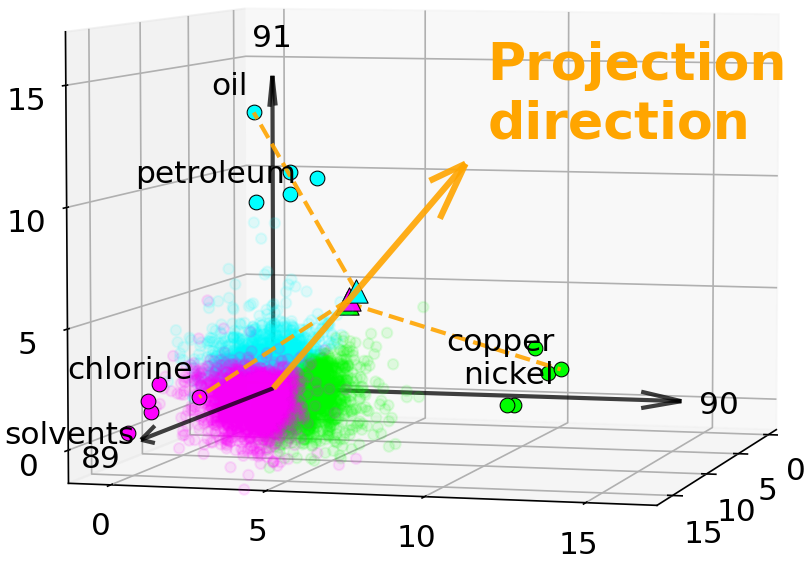}
\caption{
Projection of the subspace spanned by three consecutive axes in Fig.~\ref{fig:intro} into a one-dimensional space. Each word is assigned the color of the axis with the highest value. 
The projection direction is in the direction representing the subspace. For visualization, we randomly sampled 10,000 words, excluding the top five words on each axis.
}
\label{fig:3dproj}
\end{figure}
This section explains the projection from a subspace to a one-dimensional space using a specific example. Consider the subspace spanned by three consecutive axes (89, 90, 91) from Fig.~\ref{fig:intro}. Figure~\ref{fig:3dproj} shows the projection of this subspace using $\mathbf{f}_r$. The projection direction is in the direction representing the subspace, and the top words of each axis are projected close together.

\section{Distribution of cosine similarity}\label{app:dist-cos-sim}
Let us consider two random vectors $X=(X_1,\ldots,X_d), Y=(Y_1,\ldots,Y_d) \in \mathbb{R}^d$ with elements of mean zero $\mathbb{E}(X_\ell)=\mathbb{E}(Y_\ell)=0$ and finite variance $\mathbb{E}(X_\ell^2)=\sigma_X^2$, $\mathbb{E}(Y_\ell^2)=\sigma_Y^2$. We assume that the elements $X_1,\ldots,X_d$ and $Y_1,\ldots,Y_d$ are independent, and the sequence $X_1Y_1,\ldots,X_dY_d$ satisfies Lindeberg's condition \cite{billingsley1995probability}; for $Z_\ell = X_\ell Y_\ell/\sigma_X\sigma_Y$, $\forall \epsilon>0, \lim_{d\to\infty} d^{-1}\sum_{\ell=1}^d \mathbb{E}\Bigl[Z_\ell^2 1(|Z_\ell|>\epsilon \sqrt{d}) \Bigr]= 0$. Lindeberg's condition means no one $X_\ell Y_\ell$ dominates the inner product $\langle X, Y \rangle=\sum_{\ell=1}^d X_\ell Y_\ell$, and it is satisfied, for example, when all the elements follow the identical distribution.

Then, for sufficiently large $d$, the cosine similarity $\cos(X,Y)$ asymptotically follows $\mathcal{N}(0,1/d)$, the normal distribution with mean 0 and variance $1/d$. In other words, for sufficiently large $d$,
\begin{equation} \label{eq:clt-cosine-similarity}
    \sqrt{d} \cos(X,Y) \sim \mathcal{N}(0,1).
\end{equation}

This is easily shown as follows. First note that $\mathbb{E}(X_\ell Y_\ell)=\mathbb{E}(X_\ell)\mathbb{E}(Y_\ell)=0$, $\mathbb{E}(X_\ell^2Y_\ell^2)=\mathbb{E}(X_\ell^2)\mathbb{E}(Y_\ell^2)=\sigma_X^2\sigma_Y^2$. Thus the inner product, if scaled by dimension, $d^{-1/2} \langle X, Y \rangle = d^{-1/2}  \sum_{\ell=1}^d X_\ell Y_\ell$ has mean zero and variance $\sigma_X^2\sigma_Y^2$.
Furthermore, according to the Lindeberg-Feller Central Limit Theorem, the distribution of the inner product asymptotically converges to the normal distribution as $d$ grows large:
\begin{equation}  \label{eq:clt-inner-product}
  d^{-1/2} \langle X, Y \rangle \sim \mathcal{N}(0,\sigma_X^2\sigma_Y^2).  
\end{equation}
It also follows from the law of large numbers, $d^{-1} \| X \|^2 = d^{-1} \sum_{\ell=1}^d X_\ell^2$ converges in probability to $\mathbb{E}(X_\ell^2)=\sigma_X^2$. Similarly $d^{-1} \| Y \|^2 \to \sigma_Y^2$ in probability. Therefore,
\[
\sqrt{d} \cos(X,Y) =\frac{d^{-1/2} \langle X, Y \rangle } {\sqrt{d^{-1}\|X\|^2}\sqrt{d^{-1}\|Y\|^2}}
\]
converges to $d^{-1/2} \langle X, Y \rangle/\sigma_X\sigma_Y $ in probability, and thus (\ref{eq:clt-inner-product}) gives (\ref{eq:clt-cosine-similarity}).

\section{Details of dimensionality reduction experiments in Section~\ref{sec:experiments}}\label{app:experiments}
\begin{table*}[!tb]
\centering
\begin{adjustbox}{width=\linewidth}
\begin{tabular}{clrrrrrrrrrrrrr}
\toprule
& & \multicolumn{4}{c}{$p=5$} & \multicolumn{4}{c}{$p=20$} & \multicolumn{4}{c}{$p=100$} & $p=300$ \\
\cmidrule(lr){3-6}\cmidrule(lr){7-10}\cmidrule(lr){11-14}\cmidrule(lr){15-15}
& Tasks & PCA & Rand. & Skew. & Tour. & PCA & Rand. & Skew. & Tour. & PCA & Rand. & Skew. & Tour. & All \\
\midrule
\multirow{31}{*}{Analogy} & capital-common-countries & 0.00 & 0.00 & 0.00 & 0.00 & 0.37 & 0.00 & 0.06 & 0.11 & 0.95 & 0.56 & 0.85 & 0.87 & 0.95\\
 & capital-world & 0.00 & 0.00 & 0.00 & 0.00 & 0.26 & 0.02 & 0.05 & 0.11 & 0.90 & 0.73 & 0.80 & 0.82 & 0.95\\
 & city-in-state & 0.00 & 0.00 & 0.00 & 0.00 & 0.03 & 0.00 & 0.02 & 0.04 & 0.42 & 0.28 & 0.24 & 0.40 & 0.67\\
 & currency & 0.00 & 0.00 & 0.00 & 0.00 & 0.02 & 0.00 & 0.00 & 0.02 & 0.09 & 0.08 & 0.08 & 0.10 & 0.12\\
 & family & 0.01 & 0.00 & 0.00 & 0.00 & 0.40 & 0.02 & 0.08 & 0.22 & 0.78 & 0.68 & 0.80 & 0.75 & 0.88\\
 & gram1-adjective-to-adverb & 0.00 & 0.00 & 0.00 & 0.00 & 0.01 & 0.00 & 0.01 & 0.01 & 0.08 & 0.08 & 0.14 & 0.09 & 0.21\\
 & gram2-opposite & 0.00 & 0.00 & 0.00 & 0.00 & 0.00 & 0.00 & 0.00 & 0.00 & 0.14 & 0.12 & 0.18 & 0.14 & 0.26\\
 & gram3-comparative & 0.00 & 0.00 & 0.00 & 0.00 & 0.05 & 0.01 & 0.04 & 0.11 & 0.62 & 0.46 & 0.58 & 0.66 & 0.88\\
 & gram4-superlative & 0.00 & 0.00 & 0.00 & 0.00 & 0.01 & 0.00 & 0.00 & 0.01 & 0.31 & 0.54 & 0.23 & 0.31 & 0.69\\
 & gram5-present-participle & 0.00 & 0.00 & 0.00 & 0.00 & 0.03 & 0.00 & 0.03 & 0.07 & 0.44 & 0.30 & 0.59 & 0.58 & 0.69\\
 & gram6-nationality-adjective & 0.00 & 0.00 & 0.00 & 0.00 & 0.57 & 0.07 & 0.22 & 0.43 & 0.91 & 0.88 & 0.91 & 0.88 & 0.93\\
 & gram7-past-tense & 0.00 & 0.00 & 0.00 & 0.00 & 0.05 & 0.02 & 0.04 & 0.06 & 0.45 & 0.36 & 0.47 & 0.51 & 0.60\\
 & gram8-plural & 0.00 & 0.00 & 0.00 & 0.00 & 0.11 & 0.01 & 0.04 & 0.07 & 0.73 & 0.40 & 0.59 & 0.56 & 0.76\\
 & gram9-plural-verbs & 0.00 & 0.00 & 0.00 & 0.00 & 0.06 & 0.01 & 0.02 & 0.06 & 0.39 & 0.27 & 0.29 & 0.53 & 0.58\\
 & jj\_jjr & 0.00 & 0.00 & 0.00 & 0.00 & 0.01 & 0.00 & 0.01 & 0.03 & 0.35 & 0.23 & 0.29 & 0.43 & 0.66\\
 & jj\_jjs & 0.00 & 0.00 & 0.00 & 0.00 & 0.01 & 0.00 & 0.00 & 0.01 & 0.21 & 0.36 & 0.14 & 0.20 & 0.51\\
 & jjr\_jj & 0.00 & 0.00 & 0.00 & 0.00 & 0.01 & 0.00 & 0.03 & 0.02 & 0.33 & 0.27 & 0.32 & 0.33 & 0.54\\
 & jjr\_jjs & 0.00 & 0.00 & 0.00 & 0.00 & 0.01 & 0.00 & 0.01 & 0.01 & 0.24 & 0.37 & 0.15 & 0.20 & 0.55\\
 & jjs\_jj & 0.00 & 0.00 & 0.00 & 0.00 & 0.01 & 0.00 & 0.01 & 0.01 & 0.21 & 0.13 & 0.18 & 0.19 & 0.48\\
 & jjs\_jjr & 0.00 & 0.00 & 0.00 & 0.00 & 0.01 & 0.00 & 0.01 & 0.01 & 0.29 & 0.16 & 0.25 & 0.33 & 0.63\\
 & nn\_nnpos & 0.00 & 0.00 & 0.00 & 0.00 & 0.05 & 0.01 & 0.01 & 0.04 & 0.35 & 0.20 & 0.29 & 0.28 & 0.42\\
 & nn\_nns & 0.00 & 0.00 & 0.00 & 0.00 & 0.04 & 0.00 & 0.04 & 0.06 & 0.55 & 0.33 & 0.49 & 0.51 & 0.74\\
 & nnpos\_nn & 0.00 & 0.00 & 0.00 & 0.00 & 0.03 & 0.00 & 0.02 & 0.06 & 0.40 & 0.20 & 0.34 & 0.31 & 0.45\\
 & nns\_nn & 0.00 & 0.00 & 0.00 & 0.00 & 0.05 & 0.00 & 0.04 & 0.06 & 0.48 & 0.30 & 0.43 & 0.44 & 0.64\\
 & vb\_vbd & 0.00 & 0.00 & 0.00 & 0.00 & 0.11 & 0.03 & 0.09 & 0.07 & 0.45 & 0.36 & 0.40 & 0.54 & 0.58\\
 & vb\_vbz & 0.00 & 0.00 & 0.00 & 0.00 & 0.09 & 0.02 & 0.05 & 0.08 & 0.58 & 0.33 & 0.50 & 0.68 & 0.76\\
 & vbd\_vb & 0.00 & 0.00 & 0.00 & 0.00 & 0.09 & 0.04 & 0.12 & 0.08 & 0.46 & 0.36 & 0.56 & 0.56 & 0.69\\
 & vbd\_vbz & 0.00 & 0.00 & 0.00 & 0.00 & 0.08 & 0.01 & 0.04 & 0.05 & 0.48 & 0.26 & 0.38 & 0.54 & 0.63\\
 & vbz\_vb & 0.00 & 0.00 & 0.00 & 0.00 & 0.06 & 0.02 & 0.06 & 0.11 & 0.65 & 0.43 & 0.70 & 0.70 & 0.82\\
 & vbz\_vbd & 0.00 & 0.00 & 0.00 & 0.00 & 0.05 & 0.01 & 0.03 & 0.05 & 0.33 & 0.33 & 0.42 & 0.44 & 0.55\\
\cmidrule(lr){2-15}
 & Average & 0.00 & 0.00 & 0.00 & 0.00 & \textbf{0.09} & 0.01 & 0.04 & 0.07 & 0.45 & 0.34 & 0.42 & \textbf{0.46} & 0.63\\
\midrule
\multirow{9}{*}{Similarity} & MEN & 0.16 & 0.19 & 0.11 & 0.35 & 0.32 & 0.33 & 0.29 & 0.51 & 0.66 & 0.56 & 0.63 & 0.66 & 0.75\\
 & MTurk & 0.17 & 0.08 & 0.12 & 0.32 & 0.38 & 0.32 & 0.30 & 0.52 & 0.57 & 0.53 & 0.57 & 0.61 & 0.64\\
 & RG65 & 0.31 & 0.09 & 0.05 & 0.29 & 0.36 & 0.42 & 0.28 & 0.50 & 0.68 & 0.66 & 0.59 & 0.63 & 0.78\\
 & RW & 0.09 & 0.10 & 0.07 & 0.13 & 0.14 & 0.16 & 0.10 & 0.25 & 0.24 & 0.32 & 0.28 & 0.30 & 0.34\\
 & SimLex999 & 0.01 & 0.13 & 0.04 & 0.07 & 0.11 & 0.21 & 0.08 & 0.21 & 0.27 & 0.37 & 0.28 & 0.31 & 0.40\\
 & WS353 & 0.12 & 0.16 & 0.02 & 0.31 & 0.15 & 0.28 & 0.18 & 0.44 & 0.47 & 0.40 & 0.43 & 0.52 & 0.57\\
 & WS353R & 0.12 & 0.14 & 0.01 & 0.17 & 0.15 & 0.16 & 0.15 & 0.35 & 0.40 & 0.28 & 0.35 & 0.44 & 0.51\\
 & WS353S & 0.18 & 0.21 & 0.06 & 0.45 & 0.21 & 0.35 & 0.26 & 0.55 & 0.57 & 0.51 & 0.58 & 0.62 & 0.69\\
\cmidrule(lr){2-15}
 & Average & 0.15 & 0.14 & 0.06 & \textbf{0.26} & 0.23 & 0.28 & 0.20 & \textbf{0.42} & 0.48 & 0.45 & 0.46 & \textbf{0.51} & 0.59\\
\midrule
\multirow{7}{*}{Categorization} & AP & 0.33 & 0.22 & 0.22 & 0.27 & 0.36 & 0.26 & 0.28 & 0.40 & 0.51 & 0.45 & 0.54 & 0.56 & 0.66\\
 & BLESS & 0.31 & 0.28 & 0.27 & 0.36 & 0.42 & 0.36 & 0.35 & 0.51 & 0.73 & 0.68 & 0.69 & 0.76 & 0.79\\
 & Battig & 0.18 & 0.10 & 0.12 & 0.15 & 0.24 & 0.14 & 0.16 & 0.22 & 0.37 & 0.29 & 0.35 & 0.34 & 0.42\\
 & ESSLI\_1a & 0.50 & 0.41 & 0.45 & 0.64 & 0.61 & 0.57 & 0.48 & 0.77 & 0.73 & 0.59 & 0.73 & 0.75 & 0.70\\
 & ESSLI\_2b & 0.47 & 0.62 & 0.45 & 0.53 & 0.73 & 0.68 & 0.55 & 0.68 & 0.75 & 0.75 & 0.70 & 0.75 & 0.78\\
 & ESSLI\_2c & 0.44 & 0.44 & 0.42 & 0.44 & 0.53 & 0.60 & 0.49 & 0.56 & 0.58 & 0.53 & 0.60 & 0.60 & 0.58\\
\cmidrule(lr){2-15}
 & Average & 0.37 & 0.35 & 0.32 & \textbf{0.40} & 0.48 & 0.44 & 0.38 & \textbf{0.52} & 0.61 & 0.55 & 0.60 & \textbf{0.63} & 0.65\\
\bottomrule
\end{tabular}
\end{adjustbox}
\caption{
The performance of dimensionality reduction for $p$-dimensional embeddings. \emph{Rand.} stands for Random Order, \emph{Skew.} for Skewness Sort, and \emph{Tour.} for Axis Tour. The values in the table correspond to top 1 accuracy for analogy tasks, Spearman's rank correlation for word similarity tasks, and purity for categorization tasks. Note that at $p=300$, all embeddings give the same results.
}
\label{tab:downstream_tasks}
\end{table*}
\begin{figure*}[t]
    \centering
    \includegraphics[width=\linewidth]{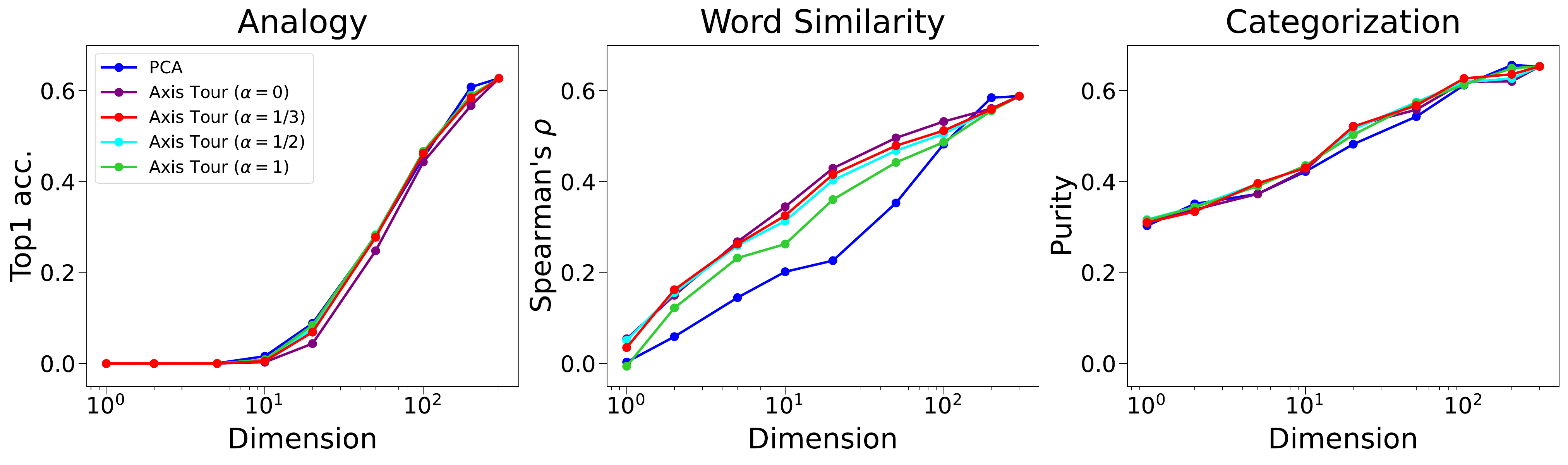}
    \caption{
The performance of dimensionality reduction for the PCA-transformed embeddings and the Axis Tour embeddings with $\alpha=0,1/3,1/2,1$. Each value represents the average of 30 analogy tasks, 8 word similarity tasks, or 6 categorization tasks. 
}
    \label{fig:comp-alpha}
\end{figure*}

\subsection{Detailed explanation of each task}

\paragraph{Analogy task.}
The embedding of the $\text{word}_i$ is denoted by $\mathbf{y}_i\in\mathbb{R}^{d}$. 
We used the Google Analogy Test Set~\cite{analogy-google}, which contains 14 types of analogy tasks, and the Microsoft Research Syntactic Analogies Dataset~\cite{analogy-msr}, which contains 16 types of analogy tasks. In the analogy tasks, the quality of the embeddings is evaluated by inferring $\text{word}_4$ to which $\text{word}_3$ corresponds if $\text{word}_1$ corresponds to $\text{word}_2$. We compute the vector $\mathbf{y}_{2} - \mathbf{y}_{1}+\mathbf{y}_{3}$ and see if the closest embedding is $\mathbf{y}_{4}$ (top1 accuracy).

\paragraph{Word similarity task.}
We used MEN~\cite{ws-MEN}, MTurk~\cite{ws-MTurk}, RG65~\cite{ws-RG65}, RW~\cite{ws-RW}, SimLex999~\cite{ws-SimLex999}, WS353~\cite{ws-WS353}, WS353R (WS353 Relatedness), and WS353S (WS353 Similarity). In the word similarity tasks, the quality of the embeddings is evaluated by measuring the cosine similarity of the word embeddings and comparing it to the human-rated similarity scores. As the evaluation metric, we used Spearman's rank correlation coefficient between the human ratings and the cosine similarity.

\paragraph{Categorization task.}
We used AP~\cite{cat-AP}, BLESS~\cite{cat-BLESS}, Battig~\cite{cat-Battig}, ESSLLI\_1a~\cite{cat-ESSLLI}, ESSLLI\_2b~\cite{cat-ESSLLI}, and ESSLLI\_2c~\cite{cat-ESSLLI}. In the categorization tasks, the quality of the embeddings is evaluated by clustering them in the setting where each word is assigned a class label. As the evaluation metric, we used Purity, which shows the proportion of the most frequent class in the clusters. As clustering methods, we used Hierarchical Clustering with five settings\footnote{By default, Word Embedding Benchmark uses the following affinity and linkage pairs for hierarchical clustering: (affinity, linkage) = (euclidean, ward), (euclidean, average), (euclidean, complete), (cosine, average), (cosine, complete).} and $K$-means\footnote{We used the same seed for all experiments.}, and then selected the one that gave the highest purity.

\subsection{Results}
Table~\ref{tab:downstream_tasks} shows detailed experimental results of PCA, Random Order, Skewness Sort, and Axis Tour at $p=5,20,100,300$ for each task.
As already seen, Fig.~\ref{fig:downstream} in Section~\ref{sec:experiments} shows the average of each task at $p=1,2,5,10,20,50,100,200,300$ for the embeddings.

The Axis Tour embeddings showed superior performance in the word similarity tasks and the categorization tasks for almost all dimensions compared to other methods. In the analogy tasks, the Axis Tour embeddings achieved performance comparable to PCA and better than Random Order and Skewness Sort in most dimensions.

\subsection{Setting of $p=d\,(=300)$}\label{app:setting_p_d}
Note that the experimental results are the same for all embeddings for $p=300$.
First, as we saw in Appendix~\ref{app:def-f}, when $p=d\,(=300$), the matrix $\mathbf{TF}$ (i.e., the projected $p$-dimensional embeddings) is equal to the matrix $\mathbf{T}$ (i.e., the $d$-dimensional Axis Tour embeddings). Then, by definition, Axis Tour, Random Order, and Skewness Sort are the embeddings obtained by reordering the axes of the ICA-transformed embeddings and flipping their signs as needed. Thus, these three can be seen as the embeddings obtained by applying an orthogonal matrix to the ICA-transformed embeddings. Since the ICA-transformed embeddings are derived from the PCA-transformed embeddings by applying an orthogonal matrix\footnote{Refer to the previous work for the relationship between PCA and ICA~\cite{DBLP:conf/emnlp/YamagiwaOS23}.}, cosine similarity and Euclidean distance remain unchanged for PCA, Random Order, Skewness Sort, and Axis Tour, leading to identical results in downstream tasks.

\section{Additional experiments}\label{app:additional-experiments}

\subsection{Qualitative observations for all axes of the Axis Tour embeddings}\label{app:all-examples}
\begin{table*}[!t]
\centering
\tiny
\begin{tabular}{@{\hspace{0.5em}}c@{\hspace{0.5em}}c@{\hspace{0.5em}}c@{\hspace{0.5em}}c@{\hspace{0.5em}}c@{\hspace{0.5em}}c@{\hspace{0.5em}}c@{\hspace{0.5em}}c@{\hspace{0.5em}}c@{\hspace{0.5em}}c@{\hspace{0.5em}}}
\toprule
         0 &                       1 &         2 &          3 &         4 &        5 &       6 &          7 &      8 &           9 \\
\midrule
     phaen &                  region &  mountain &      stage &     italy &     juan &      da &      state &  india &        vaas \\
 sandretto &                 goriška &     mount & vinokourov &   italian &    spain &     são & terengganu & indian &  jayasuriya \\
nakhchivan &               languedoc & mountains &     vuelta &        di &     luis &   paulo &      kedah &  singh &      wicket \\
 burghardt &                 regions &   everest &     stages & francesco & gonzález &    joão &     perlis &   shri & jayawardene \\
     regno & saguenay-lac-saint-jean &     peaks &  magicians &    pietro &   garcía & janeiro &   kelantan &  delhi &      kallis \\
\midrule
\midrule
          10 &          11 &            12 &       13 &         14 &          15 &        16 &      17 &            18 &               19 \\
\midrule
      andrew &       chief &       general &  bhavsar &  spearritt & contributed &        by & ***.com & micro-history &           family \\
      divoff &   executive & then-attorney &    beppe &     297.00 &       sebti &  sivuyile &   tburr &     1977-2010 &       lythraceae \\
vanwyngarden &         ceo &          gen. &     ayme & moçambique &        avet &  mulyanda &   fluto &     1951-1972 & chrysobalanaceae \\
      hruska & justiceship &       sindiso &    kahrd &      epoca &       aygin & boonradom & ***.com &           qc8 &     polyporaceae \\
    hampsten &    hienonen &        jiyane & ripstein &     huajun &       toosi &    tolkun & ***.com &     1977-2006 &   pyronemataceae \\
\midrule
\midrule
        20 &          21 &         22 &       23 &        24 &     25 &      26 &     27 &           28 &         29 \\
\midrule
     order &         the & hungarians &     serb &   russian &  czech & germany & france &       canada &  australia \\
neuroptera & macdougalls &     ethnic &  bosnian &    russia & prague &  german & french &     canadian & australian \\
 boletales &   powhatans &     asians &  croatia &    moscow & poland &  berlin &     le &      ontario & queensland \\
       svu &     andhras &     tatars & croatian &    sergei & polish &     von &  paris &       quebec &   brisbane \\
 poverelle &   sasanians &    berbers &  serbian & aleksandr & warsaw & cologne &     du & saskatchewan &      perth \\
\midrule
\midrule
          30 &        31 &       32 &       33 &         34 &      35 &        36 &        37 &            38 &    39 \\
\midrule
   wiltshire &      liga &    coach &    frank &       tony &    mike &   michael &    joseph &          jack & \&amp; \\
  shrewsbury & relegated & coaching &    capra &      ianno & johanns &  finnissy &       joe &       palance &   llp \\
lincolnshire &        fc &  scolari &  sinatra &  canzoneri & fetters &  cerveris &   macenka &     o'lantern &   amp \\
peterborough &      f.c. &    vogts &   wisner &    oursler &   petke & stuhlbarg &      papp & spring-heeled &     \& \\
     croydon &     serie &  capello & aigbogun & kornheiser &  lupica &   tomasky & nollekens &        lemmon &  firm \\
\midrule
\midrule
         40 &            41 &         42 &            43 &   44 &         45 &           46 &        47 &         48 &    49 \\
\midrule
    analyst &            an & three-week &   festivities & 27th &    striker &   goalkeeper & mickelson &        6-4 &   2-9 \\
 strategist &          rohp &  20-minute &   celebration & 22nd &  equalized &       keeper &     furyk &        7-5 &   8-8 \\
 securities & average-sized &   two-week &  celebrations & 26th & midfielder &       goalie &       els &    roddick &   1-9 \\
      hyoty &     waidhofen &  two-month & commemoration & 23rd &     header &   goaltender &     faldo & kafelnikov &  4-12 \\
udomsirikul &        taisce &  15-minute &    ceremonies & 35th &  equalised & jaaskelainen &     woods & dementieva & 17-65 \\
\midrule
\midrule
   50 &            51 &        52 &         53 &   54 &       55 &         56 &            57 &     58 &                59 \\
\midrule
24-14 &     touchdown &  rebounds &     inning & .301 & 9-for-13 &         on &        island &  greek &            george \\
24-17 &   quarterback &  hardaway &     hitter & .292 &  3-for-8 &   premised &       islands & greece &             takei \\
14-10 &    touchdowns &    pippen &    pitcher & .293 & 5-for-12 & picturized & fuerteventura & athens & stroumboulopoulos \\
20-10 &            qb & shaquille & outfielder & .289 & 4-for-11 &      ixnay &     conanicut & greeks &            w.bush \\
27-17 & interceptions &   mcdyess &    baseman & .288 & 12-of-19 &    gorging &    wangerooge &   zeus &           maharis \\
\midrule
\midrule
       60 &       61 &     62 &            63 &         64 &          65 &       66 &           67 &          68 &         69 \\
\midrule
president &  burundi & envoys &       between &   brouhaha &  challenges & numerous &      mistake &    horrific &  vitriolic \\
 teburoro &   uganda &  talks &  relationship &       over &    dilemmas &  various &        error &    terrible &     racist \\
     vice & tanzania &  envoy &       quarrel & editorship & confronting &   sundry &     mistakes &  horrendous & denouncing \\
 bagbandy &    kenya &  annan &       rivalry &    dispute &     dilemma &    other & misjudgments &    horrible &   strident \\
  issayas &   zambia & solana & relationships &   damocles &      vexing &  mishaps &    disregard & unspeakable &    insults \\
\midrule
\midrule
       70 &              71 &        72 &           73 &              74 &             75 &       76 &             77 &        78 &        79 \\
\midrule
  disgust &        tenacity &   informs &         life &           their &       arguably & livelier &     conclusive &    endear &   slowing \\
    anger &        humility &  destroys & expectancies &             our &      liveliest &  nastier &  corroboration & semblance &  slowdown \\
 feelings &       toughness &  sustains &     commuted &        forbears &      costliest &  rougher &             no &   wobbled & weakening \\
  sadness &        newfound & confronts &   transience & fellow-citizens &       quietest & prettier & substantiation &     shaky &  sluggish \\
revulsion & professionalism &      goes &   great-west &             its & best-preserved & deadlier &          scant & unscathed &   decline \\
\midrule
\midrule
   80 &   81 &    82 &      83 &        84 &      85 &       86 &             87 &       88 &        89 \\
\midrule
index & 1.48 & 20.45 & 30-year &         2 &   sauce &     beer & antidepressant & proteins &  chlorine \\
  asx & 1.62 & 15.55 &   yield &         1 &  cooked &    drink &          drugs &  protein &  solvents \\
  ase & 2.07 & 33.65 &    bond & teaspoons &  cheese &   drinks &           drug &    genes &   ammonia \\
 ftse & 2.05 & 13.45 & 10-year &         3 & roasted &   brewed &    medications &     gene &    liquid \\
 klse & 1.67 & 17.35 &  yields &  teaspoon &   bread & drinkers &         prozac &      rna & flammable \\
\midrule
\midrule
        90 &        91 &            92 &                   93 &          94 &          95 &       96 &           97 &            98 &         99 \\
\midrule
    copper &       oil &         power &                 line &        road & three-story &    hotel &       stores &      industry & subsidiary \\
    nickel & petroleum &   electricity &              railway &     highway &       brick &   hotels &        store &        export &    company \\
 manganese & oilfields &    substation &                lines &       route &      facade &   resort &      grocery & manufacturing &      maker \\
molybdenum &     crude & hydroelectric & arbatsko-pokrovskaya & north-south &   two-story &  resorts & supermarkets &  distributors &    alcatel \\
       ore &     pemex &     megawatts &               kolsås &    two-lane &    building & marriott &     retailer &    shoemaking &      corp. \\
\midrule
\midrule
        100 &      101 &        102 &         103 &       104 &        105 &         106 &        107 &         108 &        109 \\
\midrule
      hedge &      pay &       land &        laws &     court &   lawsuits &     charges &       camp &     corpses &     remain \\
       fund &     fees &   property & regulations &     judge &    lawsuit &     alleged &     prison &      corpse &   remained \\
      funds & payments &      lands &     enacted & appellate & litigation & prosecutors & buchenwald &     exhumed &     stayed \\
investments &  payment &     estate &         law &   appeals &      suits &    indicted &      camps & dismembered & stubbornly \\
  investing &     paid & bergisches &  provisions &   supreme &       suit &   convicted &    inmates &      bodies &      stays \\
\midrule
\midrule
        110 &         111 &       112 &       113 &              114 &           115 &         116 &          117 &    118 &       119 \\
\midrule
 hopelessly &  incredibly & ingenious &        to &        fostering &      tricking &        into &          has & 19,583 &     3,048 \\
 frustrated &   amazingly &   devious &    intend &      initiatives &        busily &     morphed &          had & 21,563 &     dolne \\
   woefully &   extremely &    clever &      able &      sustainable & concentrating &  transmuted &        beeen & 16,875 &   prateek \\
  hamstrung &        very & intricate & humiliate &       empowering &   classifying & degenerates & consistantly & 20,833 &     bugis \\
chronically & wonderfully &   cunning &       try & entrepreneurship &     disposing &     delving &   reinvented & 30,313 & gumbinnen \\
\bottomrule
\end{tabular}
\caption{The top five words for the $0$th axis to the $119$th axis when we apply Axis Tour to 300-dimensional GloVe.}
\label{tab:0-119}
\end{table*}

\begin{table*}[!t]
\centering
\tiny
\begin{tabular}{@{\hspace{0.3em}}c@{\hspace{0.3em}}c@{\hspace{0.3em}}c@{\hspace{0.3em}}c@{\hspace{0.3em}}c@{\hspace{0.3em}}c@{\hspace{0.3em}}c@{\hspace{0.3em}}c@{\hspace{0.3em}}c@{\hspace{0.3em}}c@{\hspace{0.3em}}}
\toprule
              120 &       121 &       122 &     123 &        124 &   125 &  126 &       127 &           128 &    129 \\
\midrule
           system &   desktop &     phone &     web & television &  2300 & 4:23 &         \_ &           new & 121.58 \\
          systems & macintosh &  cellular &   sites &         tv & 12:30 & 2:47 &    ondeck &          york & 114.78 \\
renin-angiotensin &        pc & cellphone &    site &    channel & 11:30 & 2:33 &  holimont &    orleanians & 121.22 \\
      centralized &  software & telephone &  online &  broadcast &  0330 & 8:38 & hawksnest & orleans-based & 119.76 \\
     computerised &    server &  wireless & myspace &        cbs &  0930 & 2:21 & belleayre &          n.y. & 106.12 \\
\midrule
\midrule
      130 &       131 &    132 & 133 & 134 &   135 &   136 &  137 &  138 & 139 \\
\midrule
seventies & 2001-2003 &   four &  89 & 445 & 1,149 & 209.6 & 95.3 & 28.4 & 5.5 \\
  sixties & 2003-2005 &   five &  83 & 244 & 4,461 & 218.8 & 96.2 & 23.1 & 4.8 \\
    2010s & 1999-2001 &  three &  86 & 285 & 1,737 & 218.3 & 89.8 & 23.9 & 5.7 \\
    1800s & 1998-2001 & eleven &  85 & 292 & 1,701 & 308.9 & 89.9 & 26.1 & 4.4 \\
    early & 1995-1997 &    six &  88 & 344 & 1,109 & 246.3 & 93.1 & 27.6 & 4.6 \\
\midrule
\midrule
    140 &        141 &            142 &      143 &           144 &          145 &      146 &          147 &     148 &           149 \\
\midrule
 23,000 &      union &          alike & theories &        church &          sex &  actress &  german-born &     old &  multibillion \\
110,000 &    workers &      academics &   theory &      anglican &       sexual &     wife & newspaperman & 35-year &    20-million \\
 39,000 & machinists & industrialists &  posited &     episcopal &   homosexual &   mother & entrepreneur & 14-year & multi-million \\
 43,000 &     unions &   sociologists &  notions & congregations & heterosexual & daughter &   politician & 50-year &   100-million \\
 48,000 &  teamsters &        pundits &   notion &  congregation &    unmarried &      née & russian-born & 22-year &  multimillion \\
\midrule
\midrule
    150 &       151 &               152 &         153 &        154 &     155 &       156 &          157 &         158 &          159 \\
\midrule
      ) &    rights &             group &       force & insurgents &    guns &    wheels &        sedan &  schumacher &    preakness \\
      ( &     human &            groups & gendarmerie &  militants &  rifles &     wheel &       sedans & barrichello &      belmont \\
  .0358 &   aprodeh & lashkar-e-jhangvi &  contingent &     gunmen &     gun &    rudder &           v8 &       massa &        filly \\
 3.7996 & zimrights &           forzani &  500-strong & guerrillas & caliber &    brakes & turbocharged &     ferrari &      baffert \\
unitals &    pillay &            harkat & contingents &     forces &  weapon & hydraulic &        camry &   raikkonen & thoroughbred \\
\midrule
\midrule
      160 &        161 &          162 &         163 &  164 &           165 &        166 & 167 &         168 &      169 \\
\midrule
     400m &     sports & welterweight &          la &  que & basidiomycota &      .2667 & lb3 &      drawno & behshahr \\
     100m &      sport &  heavyweight &       cerva & pero &       gorlice & estrategia & 5lb & krośniewice &    abyek \\
     200m &   softball & middleweight & cenerentola &  sus &       empleos &     creado & 8lb &     brochów &   ramian \\
freestyle & volleyball &          ibf &    ferrière &   en &       autovía &      hirta & lb7 &    lubaczów &   khamir \\
200-meter &     soccer &          wba &    louvière &  una &       pudiera &     a.m.-6 & 4lb & pobiedziska & javanrud \\
\midrule
\midrule
    170 &         171 &      172 &        173 &           174 &      175 &      176 &       177 &                178 &           179 \\
\midrule
   bank &       river &     lake &       city & international & thailand & minister & spokesman &              nafez &         adtac \\
central & ljubljanica &   mývatn &     ozamis &       airport &     thai &    prime & faizasyah &            sedwill &       coronae \\
 nivard &      tigris &    erhai & malaybalay &       tocumen &   nakhon &  minster &    cirtek & british-controlled &      maḩalleh \\
  banks &   tributary &   chilka &     phenix &        gökçen &  bangkok & minsiter &  tsanchev &           videoton & decarboxylase \\
   pboc &      rivers & waramaug &      hitec &        aiport &    thais & interior &   ladsous &              2-6-2 &      poderoso \\
\midrule
\midrule
     180 &      181 &            182 &        183 &          184 &                                     185 &          186 &   187 &           188 &                189 \\
\midrule
   8a-4p & qeshlaqi &    automoviles &     jreyes & ***-***-**** &      \_\_\_\_\_... &       moodin &    -- &             - &                  : \\
   romik &  dizaj-e &           vx-6 &     ilovar &        .0210 &                                 interbk & ***-***-**** & skway &           bkh &        www.***.com \\
  shefer &  kalayeh &   ***-***-**** &      hohtz & ***-***-**** & ooooo... &         wehz &  thoh &   wc2006-asia & http://www.***.org \\
 samayoa &    now-e & principalmente & ray-finned &  ***@***.com &                            ***-***-**** &       prahnk & kursh &        manutd & http://www.***.com \\
hermanas &  patryan &       yengejeh &   odalovic &        65stk &                            harrynytimes &        eesah &  tsih & pickup5thgraf & http://www.***.org \\
\midrule
\midrule
  190 &       191 &        192 &          193 &          194 &       195 &         196 &      197 &     198 &     199 \\
\midrule
    ; &      sarā &        rd2 &    ipnotinmx &      gibbosa &   1507.50 & ixmiquilpan &  prolegs &  masuku & artayev \\
 priu &        as &   clientes &     mordella & pratylenchus &   .000663 & ***@***.com &     grij & quartic & autuori \\
   sư &    chahār & weinzapfel &  significado &   bifrenaria &  analista &      .71078 &    macul &  havner &  tayyab \\
.8226 &    khvosh &  desempleo &  sandwicense &   fimbriatus & ramnarine &     jagdeep & rambusch &  shealy &  pareto \\
 sa/b & akbarābād &      gevar & ***-***-**** &     laticeps &       hàm &        naab &   yr.ago &   vidro & zermelo \\
\midrule
\midrule
                200 &               201 &       202 &        203 &         204 &     205 &   206 &      207 &         208 &       209 \\
\midrule
           swaffham &          phthalic &     72-77 &  magnifica & funderburke & cunxiao &  wang &    japan &       south &      park \\
       ***-***-**** &            chamba & gildernew &     compra &    hintikka & cunshen & zhang & japanese &      korean &   prenton \\
           ollivier &            nguyên &    deanne &        qc7 &     chofetz &   cunxu &    xu &    tokyo &       korea &   ji-sung \\
               zend &          kinghorn & ethelbert &   kowalska &       danys &  siyuan &    li &    akira &    dakotans & naturpark \\
cronulla-sutherland & matanuska-susitna &    kether & handanovic &   opticians & lanqing &   liu &  takashi & carolinians &   breffni \\
\midrule
\midrule
         210 &        211 &        212 &       213 &         214 &    215 &    216 &      217 &      218 &           219 \\
\midrule
    national & parliament &      party &   polling &  republican &      . &      i & silently & trousers &       whitish \\
      winema &       duma &     janata & electoral &        sen. &      , & 'cause &   stared &    pants &     yellowish \\
    jurnalul &    speaker &        bjp & balloting &         gop &    but &  gonna & screamed &  dresses &      brownish \\
ranthambhore &   150-seat & socialists &  election & republicans & though &   yeah &   yelled &    dress &       greyish \\
        chúa &   500-seat &        ndp &  tallying &      mccain &   even &   gosh &  sobbing &  wearing & reddish-brown \\
\midrule
\midrule
        220 &        221 &      222 &     223 &       224 &          225 &        226 &       227 &      228 &      229 \\
\midrule
ultraviolet & spacecraft & aircraft &  vessel & hurricane & temperatures & grasslands &    shrubs &  mammals &    virus \\
   infrared & astronauts &      jet &    ship &     storm &        humid &   habitats &     trees &    birds &     h5n1 \\
 telescopes &  astronaut &   planes &   ships &    storms &       chilly &      soils &     vines &  rabbits &    swine \\
      light &       nasa &   boeing & vessels &   typhoon &      weather &    marshes &   planted &  animals &      flu \\
wavelengths &      orbit &     f-15 &   boats &   cyclone & unseasonably &  sediments & seedlings & reptiles & outbreak \\
\midrule
\midrule
          230 &       231 &        232 &         233 &       234 &        235 &        236 &     237 &      238 &         239 \\
\midrule
    diagnosed &   medical &    undergo &      survey &    agency &  newspaper &        ign &   award &     film &   superhero \\
         lung &      care &   thorough &     surveys &   notimex &      daily & popmatters &  awards &    films &      marvel \\
 inflammation &  hospital & undergoing &  statistics & kathpress &    zeitung &   allmusic & awarded &    movie &    spin-off \\
complications & hospitals & evaluation &      gallup &       bss &   izvestia &   reviewer &   prize & starring & superheroes \\
    fractures & physician &  undergone & statistical &     telam & kommersant &   gamespot &    emmy & directed &  characters \\
\bottomrule
\end{tabular}
\caption{
The top five words for the $120$th axis to the $239$th axis when we apply Axis Tour to 300-dimensional GloVe.
}
\label{tab:120-239}
\end{table*}

\begin{table*}[!t]
\centering
\tiny
\begin{tabular}{@{\hspace{0.5em}}c@{\hspace{0.5em}}c@{\hspace{0.5em}}c@{\hspace{0.5em}}c@{\hspace{0.5em}}c@{\hspace{0.5em}}c@{\hspace{0.5em}}c@{\hspace{0.5em}}c@{\hspace{0.5em}}c@{\hspace{0.5em}}c@{\hspace{0.5em}}}
\toprule
        240 &        241 &        242 &         243 &        244 &         245 &       246 &     247 &        248 &             249 \\
\midrule
      album &      piano &  paintings &  manuscript &   language &        name & formula\_1 &     set &       mark &           brian \\
     albums &     violin &   painting & biographies &  languages &       names & formula\_2 &  aflame & high-water & trenchard-smith \\
       band &      cello &        art &       pages &     pashto &     surname & formula\_3 &   10-cd &     dindal &    doyle-murray \\
self-titled & percussion &  sculpture &        book & colloquial &      phrase & formula\_4 & setting &   wainberg &            cadd \\
         ep & orchestral & watercolor & handwritten &    dialect & misspelling & formula\_5 &    4-cd &     hoppus &         donlevy \\
\midrule
\midrule
     250 &       251 &         252 &      253 &         254 &         255 &           256 &       257 &       258 &     259 \\
\midrule
 ireland &    martin &       peter &     paul &        john &          `` &       bazzani &     david &     scott &    2-54 \\
 belfast &    scelzo &     guillam &  virilio & rhys-davies &          '' & looking-glass &     pittu & livengood & cretier \\
   irish & wansleben & shockheaded &  ricoeur &   canemaker &         xff &        mouret &    proval &   lobdell & veltman \\
  mowlam &    clunes &      maffay & langmack &      motson &           ` &         munro & margolick &       gud &    f.r. \\
northern &      damm &    rauhofer &    celan &    mcgahern & schizopolis &        mármol &   gorcyca &  speedman &  nahant \\
\midrule
\midrule
      260 &            261 &        262 &           263 &        264 &  265 &  266 &      267 &       268 &       269 \\
\midrule
     palm &         county &     calif. &        school & university & 1976 & 1853 & february &      king &  alwaleed \\
    beach & unincorporated & california &          high &  professor & 1973 & 1852 &  october & sigismund &    prince \\
     fla. &         dekalb &  inglewood &    elementary &   graduate & 1966 & 1847 & december &        iv &     saudi \\
ostrowski &        fayette &   pasadena & pine-richland &  doctorate & 1968 & 1856 &     june &   emperor &   kingdom \\
  broward &        pulaski &     pomona &         jr/sr &    faculty & 1971 & 1854 &    april &       iii & tupouto'a \\
\midrule
\midrule
     270 &       271 &       272 &       273 &         274 &          275 &      276 &    277 &      278 &      279 \\
\midrule
     ali &    israel &    daniel &   swedish &      thomas &      william &  charles &  james & wilkison &     nack \\
      al &   israeli &    briere &    sweden & kretschmann & mastrosimone &   dutoit &  frain & mudavadi & narrowly \\
mohammed &  israelis &   macivor & norwegian &     aquinas &     beaudine &   grodin &  roday &  henwood &  gyoergy \\
  sheikh & netanyahu & gildenlöw &    norway &   quasthoff &     fichtner &  sheeler & luceno & hogzilla &    khand \\
mohammad &      aviv & balavoine &   fredrik &      fingar &      saroyan & wuorinen &  remar &  mottola &    vanak \\
\midrule
\midrule
           280 &           281 &      282 &      283 &      284 &     285 &        286 &       287 &     288 &       289 \\
\midrule
          nhls &         janyk &  vranjes &   robert &      van &   chris &   marchena &     steve &  shaara &   richard \\
      ba'asyir &        tookie &  kilvert &   deniro & lieshout & volstad & staniforth & railsback &    jeff & stoltzman \\
organizaciones &     germanica &    bukan &    halmi &     dijk &  braide &     kaboul &     turre &   shain &    ayoade \\
      yordanov &      walthers & paratore & garrigus &    zandt & bordano &     pelzer &   forbert & friesen &  basehart \\
        millán & adquisiciones &    ponge &  redford &     tuyl & tidland &      43.18 &    kimock & sharlet &  sandomir \\
\midrule
\midrule
        290 &       291 &           292 &          293 &          294 &          295 &         296 &           297 &          298 &      299 \\
\midrule
      simon & malatesta & bilyaletdinov &      .000106 &      nazione &         olya &      gearon &       sirajul &         wonk & kerberos \\
 boccanegra &     .0217 &        pouget &          fhs &        ba872 & dodecahedral &     mikuláš &   overexcited &        kappa & mckelvey \\
napier-bell &   lockard &      cytidine & colecovision &       blouin &          cih & bracigliano &       nabhani &   fraternity &    wajir \\
      vouet & 330-pound &       zhulali &      ronghua & comunicacion &        alnus &      venero & then-reigning &     godsmack &      veg \\
     callow &      puji &           bc4 &      yunlong &       wachau &         9.29 &  mariangela &    karmichael & bibliophiles &     pdca \\
\bottomrule
\end{tabular}
\caption{The top five words for the $240$th axis to the $299$th axis when we apply Axis Tour to 300-dimensional GloVe.}
\label{tab:240-299}
\end{table*}
In Table~\ref{tab:examples}, we used 300-dimensional GloVe and showed the semantic continuity of the axes of the Axis Tour embeddings, with illustrative examples of the meaning of \emph{countries} (the $23$rd axis to the $31$st axis), \emph{law} (the $101$st axis to the $109$th axis), and \emph{art} (the $237$th axis to the $245$th axis).
This section presents the top five words of the normalized embeddings across all 300 axes in Tables~\ref{tab:0-119},~\ref{tab:120-239},and~\ref{tab:240-299}.

For example, in Table~\ref{tab:0-119}, the $45$th and $46$th axes are related to \emph{soccer}, the $47$th axis to \emph{golf}, the $48$th axis to \emph{tennis}, the $49$th and $50$th axes to \emph{scores}, the $51$st axis to \emph{American football}, the $52$nd axis to \emph{basketball}, and the $53$rd axis to \emph{baseball}. These axes illustrate the semantic continuity across \emph{sports}.

In Table~\ref{tab:120-239}, the $129$th to the $140$th axes are related to \emph{numbers}. 
It is also interesting to note that the top words of each axis have similar numerical scales, highlighting how well the axis of the ICA-transformed embeddings captures meaning. 
In addition, the meaning of each axis changes continuously, much like a game of word association: the $219$th axis relates to \emph{colors}, the 220th axis to \emph{light}, the 221st axis to \emph{space}, the 222nd axis to \emph{airplanes}, the 223rd axis to \emph{ships}, the 224th axis to \emph{storms}, the 225th axis to \emph{weather}, the 226th axis to \emph{biomes}, the 227th axis to \emph{plants}.

In Table~\ref{tab:240-299}, the 272nd to the 289th axes are related to \emph{personal names} from different linguistic regions as if this were a cluster of the meaning.

Note that due to space limitations, the top 1 and top 3 words on the $185$th axis are truncated because they are repetitive symbols, and that URLs, email addresses, and phone numbers are anonymized.

\subsection{Comparison of $\alpha$}\label{app:comp-alpha}
Figure~\ref{fig:comp-alpha} shows the average of each task at $p=1,2,5,10,20,50,100,200,300$ for the PCA-transformed embeddings and the Axis Tour embeddings with $\alpha=0,1/3,1/2,1$.

From Fig.~\ref{fig:comp-alpha}, we can see that the performance of the Axis Tour embeddings changes for each task, depending on $\alpha$. For example, when comparing across all $\alpha$, while $\alpha=0$ shows good performance on word similarity tasks and poor performance on analogy tasks, $\alpha=1$ shows the opposite.

These results suggest that the quality of low-dimensional embeddings by the Axis Tour embeddings depends on the vector $\mathbf{f}_r$ for projection. However, the overall changes for each $\alpha$ are not as large, and in all tasks the performance is better than or comparable to that of the PCA-transformed embeddings, indicating the ability to construct better or comparable low-dimensional embeddings compared to PCA.

\subsection{Comparisons of $k$}\label{app:comp-k}
While we have done experiments for the Axis Tour with $k=100$ using 300-dimensional Glove in Section~\ref{sec:experiments}, when computing the axis embedding in (\ref{eq:axis-embedding}), what is the appropriate value for $k$?

For example, if $k=1$ and the top $1$ word happens to have a meaning different from that of the axis, it is not desirable to define the axis embedding using only that word. 
Conversely, as $k$ increases, the number of words with meanings different from that of the axis in the top $k$ words also increases, hindering the ability of the axis embedding to represent the meaning of the axis. For example, in an extreme case where $k=n$, $\text{Top}_k^\ell$ becomes equal to $[n]$, and \emph{all} axis embeddings become the mean vector of $\hat{\mathbf{s}}_i$ over the vocabulary set. 
In this case, it impossible to find the order that maximizes the semantic continuity of the axes.

In this section, to address these questions, we compare the metric for the semantic continuity of the axes, as defined in (\ref{eq:axistour}), for $k=1,10,100,1000$, and then perform qualitative observation and dimension reduction experiments.

\subsubsection{Selection of $k$}
In preparation, we perform the Axis Tour for $k_1=1,10,100,1000$, resulting in the embedding matrices $\mathbf{T}_{1},\mathbf{T}_{10},\mathbf{T}_{100},\mathbf{T}_{1000}$. 
In this section, we \emph{redefine} the axis embedding for them with top $k_2=1,10,100,1000$, then evaluate the metric for the semantic continuity of the axes and thus compare the quality of $\mathbf{T}_{k_1}$.

\paragraph{Redefinition of axis embedding.}
Similar to $\hat{\mathbf{S}}$, we define the matrix $\hat{\mathbf{T}}_{k_1}\in\mathbb{R}^{n\times d}$ as the normalization of the embeddings $\mathbf{T}_{k_1}$ with row vectors $\hat{\mathbf{t}}_{k_1,i} = \mathbf{t}_{k_1,i} / \|\mathbf{t}_{k_1,i}\|$. 
Here, the $i$-th word embedding of $\mathbf{T}_{k_1}$ and $\hat{\mathbf{T}}_{k_1}$ are denoted by $\mathbf{t}_{k_1,i}, \hat{\mathbf{t}}_{k_1,i} \in \mathbb{R}^d$, respectively.
We compare the elements of the $\ell$-th axis of $\hat{\mathbf{T}}_{k_1}$ and denote the index set of words corresponding to the top $k_2$ elements as $\text{Top}_{k_2}^\ell$.

We redefine the $\ell$-th axis embedding $\mathbf{v}_{\ell}{(k_1,k_2)}$ for $k_1$ and $k_2$ as follows:
\begin{align}
    \mathbf{v}_{\ell}{(k_1,k_2)} := \frac{1}{k_2}\sum_{i\in \text{Top}_{k_2}^\ell}\hat{\mathbf{t}}_{k_1,i}\in\mathbb{R}^d.
\end{align}
If $k_2=k_1$, then $\mathbf{v}_{\ell}{(k_1,k_2)}$ coincides with the axis embedding used in the Axis Tour optimization. 
In contrast, if $k_2\neq k_1$ and the semantic continuity of the axes is still observed for $\mathbf{v}_{\ell}{(k_1,k_2)}$, then $\mathbf{T}_{k_1}$ can be considered as high-quality embeddings that are robust to changes for $k_2$. 

\paragraph{Semantic continuity of axes for $\mathbf{v}_{\ell}{(k_1,k_2)}$.}
Using $\mathbf{v}_{\ell}{(k_1,k_2)}$, we define the metric for the semantic continuity of the axes for each $k_1$ and $k_2$ as follows:
\begin{multline}
c(k_1,k_2):=\frac{1}{d}\cos(\mathbf{v}_{1}{(k_1,k_2)},\mathbf{v}_{d}{(k_1,k_2)})+\\
\frac{1}{d}\sum_{\ell=1}^{d-1}\cos(\mathbf{v}_{\ell}{(k_1,k_2)},\mathbf{v}_{\ell+1}{(k_1,k_2)}).\label{eq:c-k1-k2}
\end{multline}
Note that, unlike (\ref{eq:axistour}), the axes are already ordered by the Axis Tour using $k_1$.
Furthermore, in (\ref{eq:c-k1-k2}), the sum of $\cos(\mathbf{v}_{\ell}{(k_1,k_2)},\mathbf{v}_{\ell+1}{(k_1,k_2)})$ is divided by the dimension $d$, which can be interpreted as the average of the cosine similarities between adjacent axis embeddings $\mathbf{v}_{\ell}{(k_1,k_2)}$ and $\mathbf{v}_{\ell+1}{(k_1,k_2)}$.

\paragraph{Comparison method for $k$.}
We aim to compare the quality of $\mathbf{T}_{k_1}$. 
To facilitate this, we define the univariate functions of $k$ for $k_1=1,10,100,1000$:
\begin{align}
C_{1}(k)&:=c(1,k),\label{eq:c1-def}\\
C_{10}(k)&:=c(10,k),\label{eq:c10-def}\\
C_{100}(k)&:=c(100,k),\label{eq:c100-def}\\
C_{1000}(k)&:=c(1000,k).\label{eq:c1000-def}
\end{align}
For example, $C_1(k)$ in (\ref{eq:c1-def}) is a function that measures the semantic continuity for the axis embedding $\mathbf{v}_{\ell}{(1,k)}$ redefined by the top $k$ words of the $\ell$-th axis of $\mathbf{T}_1$. 
To evaluate the robustness of $\mathbf{T}_{1}$ to changes for $k$, we average the values of $C_1(k)$ for $k=1,10,100,1000$. A higher average value indicates better quality. 

The same explanation can be applied to $\mathbf{T}_{10}$, $\mathbf{T}_{100}$, $\mathbf{T}_{1000}$, which leads to the definition of a function $M(k)$ for calculating these averages:
\begin{multline}
M(k):=\\
\frac{C_{k}(1)+C_{k}(10)+C_{k}(100)+C_{k}(1000)}{4}
\end{multline}
We compare $M(1),M(10),M(100),M(1000)$ to evaluate the quality of $\mathbf{T}_{1},\mathbf{T}_{10},\mathbf{T}_{100},\mathbf{T}_{1000}$.

\paragraph{Results.}
\begin{figure}[t!]
    \centering
    \includegraphics[width=\columnwidth]{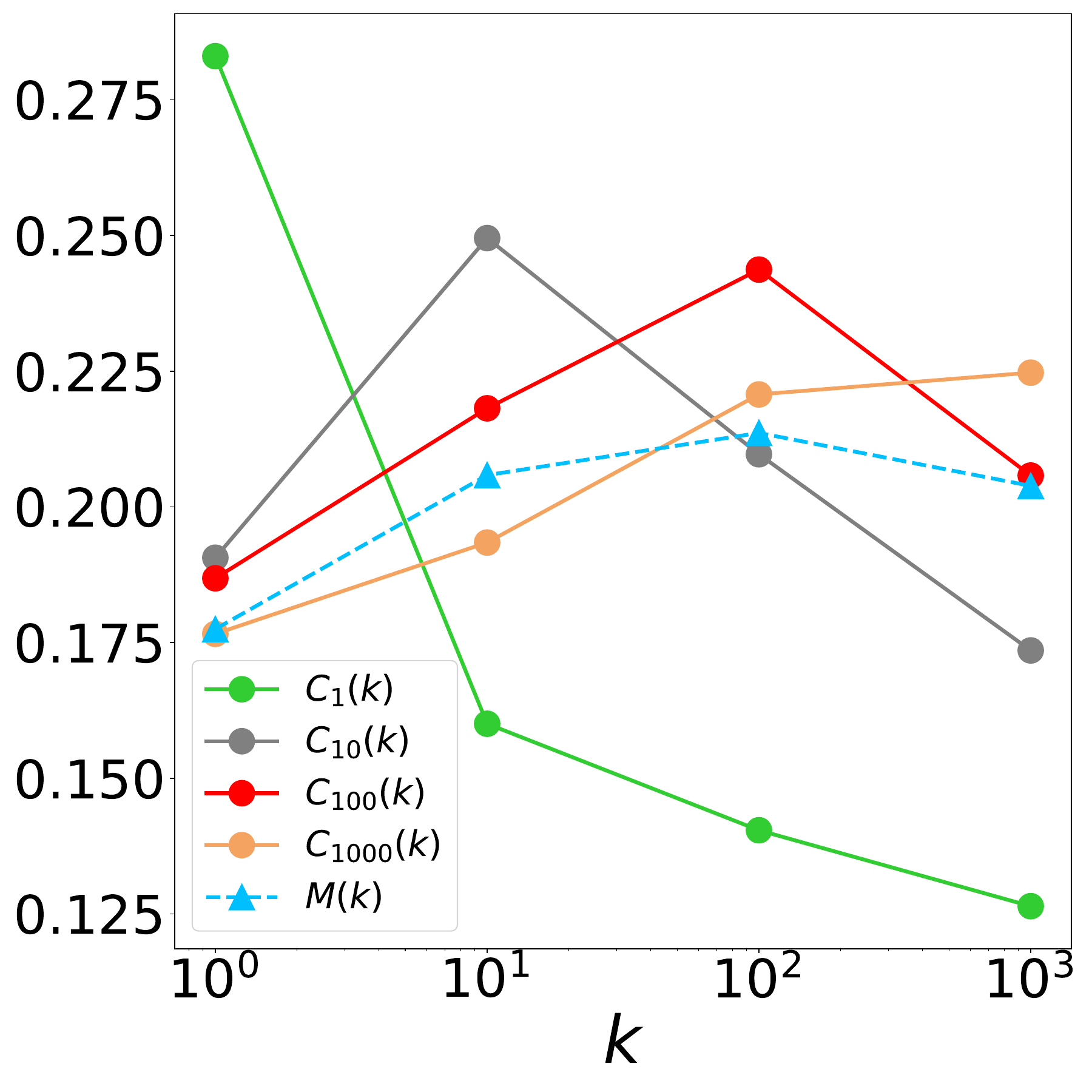}
    \caption{Plots of the functions $C_1(k)$, $C_{10}(k)$, $C_{100}(k)$, $C_{1000}(k)$ and $M(k)$ for $k = 1, 10, 100, 1000$.
    }
    \label{fig:topk-sumcossim}
\end{figure}

\begin{table*}[t!]
\tiny
\centering
\begin{subtable}{\textwidth}
\centering
\begin{tabular}{ccccccccc}
\toprule
   237 &      238 &      239 &     240 &    241 &       242 &      243 &    244 &        245 \\
\midrule
 india & thailand &    japan & germany & france &     italy &  ireland &  czech & hungarians \\
indian &     thai & japanese &  german & french &   italian &  belfast & prague &     ethnic \\
 singh &   nakhon &    tokyo &  berlin &     le &        di &    irish & poland &     asians \\
  shri &  bangkok &    akira &     von &  paris & francesco &   mowlam & polish &     tatars \\
 delhi &    thais &  takashi & cologne &     du &    pietro & northern & warsaw &    berbers \\
\midrule
\midrule
      112 &        113 &       114 &         115 &        116 &         117 &        118 &         119 &         120 \\
\midrule
       to &      order &     court &     charges &   lawsuits &  challenges &     remain &  hopelessly &  incredibly \\
   intend & neuroptera &     judge &     alleged &    lawsuit &    dilemmas &   remained &  frustrated &   amazingly \\
     able &  boletales & appellate & prosecutors & litigation & confronting &     stayed &    woefully &   extremely \\
humiliate &        svu &   appeals &    indicted &      suits &     dilemma & stubbornly &   hamstrung &        very \\
      try &  poverelle &   supreme &   convicted &       suit &      vexing &      stays & chronically & wonderfully \\
\midrule
\midrule
         220 &      221 &        222 &         223 &        224 &         225 &     226 &      227 &      228 \\
\midrule
       sedan &      van &  paintings &  manuscript &      piano &       album &   award &  actress &     film \\
      sedans & lieshout &   painting & biographies &     violin &      albums &  awards &     wife &    films \\
          v8 &     dijk &        art &       pages &      cello &        band & awarded &   mother &    movie \\
turbocharged &    zandt &  sculpture &        book & percussion & self-titled &   prize & daughter & starring \\
       camry &     tuyl & watercolor & handwritten & orchestral &          ep &    emmy &      née & directed \\
\bottomrule
\end{tabular}
\caption{$k=1$}
\label{tab:k1-examples}
\end{subtable}
\vspace{2em}

\begin{subtable}{\textwidth}
\centering
\begin{tabular}{ccccccccc}
\toprule
       102 &       103 &         104 &        105 &    106 &       107 &     108 &      109 &          110 \\
\midrule
       6-4 & mickelson &  schumacher &      stage & france &     italy &      da &     juan & welterweight \\
       7-5 &     furyk & barrichello & vinokourov & french &   italian &     são &    spain &  heavyweight \\
   roddick &       els &       massa &     vuelta &     le &        di &   paulo &     luis & middleweight \\
kafelnikov &     faldo &     ferrari &     stages &  paris & francesco &    joão & gonzález &          ibf \\
dementieva &     woods &   raikkonen &  magicians &     du &    pietro & janeiro &   garcía &          wba \\
\midrule
\midrule
          199 &          200 &         201 &       202 &        203 &         204 &        205 &         206 &       207 \\
\midrule
      between &          sex &        laws &     court &   lawsuits &     charges &       camp &     corpses &    rights \\
 relationship &       sexual & regulations &     judge &    lawsuit &     alleged &     prison &      corpse &     human \\
      quarrel &   homosexual &     enacted & appellate & litigation & prosecutors & buchenwald &     exhumed &   aprodeh \\
      rivalry & heterosexual &         law &   appeals &      suits &    indicted &      camps & dismembered & zimrights \\
relationships &    unmarried &  provisions &   supreme &       suit &   convicted &    inmates &      bodies &    pillay \\
\midrule
\midrule
         112 &     113 &        114 &         115 &        116 &        117 &         118 &     119 &       120 \\
\midrule
   preakness &   award &        ign &       album &      piano &  paintings &  manuscript &     web &     phone \\
     belmont &  awards & popmatters &      albums &     violin &   painting & biographies &   sites &  cellular \\
       filly & awarded &   allmusic &        band &      cello &        art &       pages &    site & cellphone \\
     baffert &   prize &   reviewer & self-titled & percussion &  sculpture &        book &  online & telephone \\
thoroughbred &    emmy &   gamespot &          ep & orchestral & watercolor & handwritten & myspace &  wireless \\
\bottomrule
\end{tabular}
\caption{$k=10$}
\label{tab:k10-examples}
\end{subtable}
\vspace{2em}

\begin{subtable}{\textwidth}
\centering
\begin{tabular}{ccccccccc}
\toprule
        104 &         105 &      106 &          107 &    108 &         109 &       110 &        111 &        112 \\
\midrule
        the &       river &     lake &       canada & france &          la &     italy &      stage &      piano \\
macdougalls & ljubljanica &   mývatn &     canadian & french &       cerva &   italian & vinokourov &     violin \\
  powhatans &      tigris &    erhai &      ontario &     le & cenerentola &        di &     vuelta &      cello \\
    andhras &   tributary &   chilka &       quebec &  paris &    ferrière & francesco &     stages & percussion \\
  sasanians &      rivers & waramaug & saskatchewan &     du &    louvière &    pietro &  magicians & orchestral \\
\midrule
\midrule
             189 &           190 &         191 &       192 &        193 &         194 &        195 &     196 &       197 \\
\midrule
       fostering &      tricking &        laws &     court &   lawsuits &     charges &       camp &     set & formula\_1 \\
     initiatives &        busily & regulations &     judge &    lawsuit &     alleged &     prison &  aflame & formula\_2 \\
     sustainable & concentrating &     enacted & appellate & litigation & prosecutors & buchenwald &   10-cd & formula\_3 \\
      empowering &   classifying &         law &   appeals &      suits &    indicted &      camps & setting & formula\_4 \\
entrepreneurship &     disposing &  provisions &   supreme &       suit &   convicted &    inmates &    4-cd & formula\_5 \\
\midrule
\midrule
   108 &         109 &       110 &        111 &        112 &         113 &         114 &      115 &         116 \\
\midrule
france &          la &     italy &      stage &      piano &       album &   superhero &     film &          `` \\
french &       cerva &   italian & vinokourov &     violin &      albums &      marvel &    films &          '' \\
    le & cenerentola &        di &     vuelta &      cello &        band &    spin-off &    movie &         xff \\
 paris &    ferrière & francesco &     stages & percussion & self-titled & superheroes & starring &           ` \\
    du &    louvière &    pietro &  magicians & orchestral &          ep &  characters & directed & schizopolis \\
\bottomrule
\end{tabular}
\caption{$k=1000$}
\label{tab:k1000-examples}
\end{subtable}
\vspace{2em}
\caption{
Semantic continuity of axes by Axis Tour for normalized ICA-transformed embeddings. First, we focus on each central axis of Table~\ref{tab:examples} where $k=100$. The axes are the 27th axis (\emph{france}, \emph{french}, \emph{le}, \emph{paris}, \emph{du}), the 105th axis (\emph{lawsuits}, \emph{lawsuit}, \emph{litigation}, \emph{suits}, \emph{suit}), and the 241st axis (\emph{piano}, \emph{violin}, \emph{cello}, \emph{percussion}, \emph{orchestral}). This table then shows the top five words of the axes near these selected axes for $k=1,10,1000$. Note that the axis indices change depending on the results of each Axis Tour. 
}
\label{tab:examples_k}
\end{table*}

\begin{figure*}[t!]
    \centering
    \includegraphics[width=\linewidth]{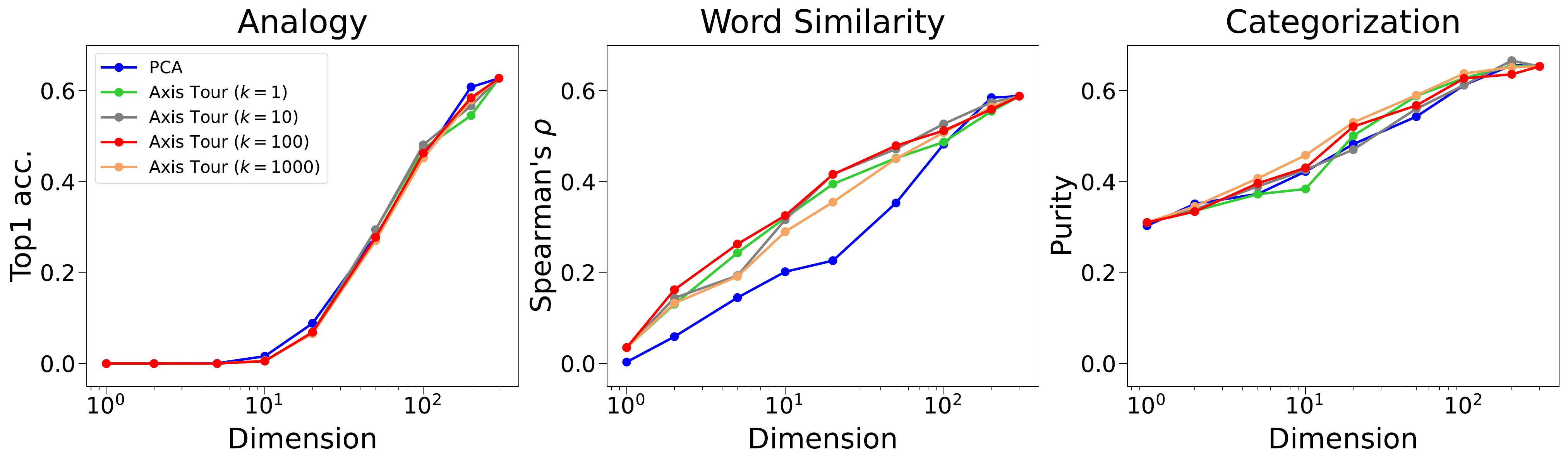}
    \caption{The performance of dimensionality reduction for the PCA-transformed embeddings and the Axis Tour embeddings with $k=1,10,100,1000$ and $\alpha=1/3$. Each value represents the average of 30 analogy tasks, 8 word similarity tasks, or 6 categorization tasks. }
    \label{fig:comp-k}
\end{figure*}

Figure~\ref{fig:topk-sumcossim} shows plots of the functions $C_1(k)$, $C_{10}(k)$, $C_{100}(k)$, $C_{1000}(k)$ and $M(k)$ for $k = 1, 10, 100, 1000$. 
$M(100)$ is the maximum value of $M(k)$. This result validates the use of $k=100$ as the default value for our experiment settings. 

Furthermore, $C_1(1)$ is the maximum value among $C_1(k)$, $C_{10}(k)$, $C_{100}(k)$, $C_{1000}(k)$ $(k = 1, 10, 100, 1000)$. 
In this setting, since we use the top 1 word of each axis for axis embedding, this Axis Tour is equivalent to the Word Tour of 300 words using cosine similarity distance. Thus, the axis embedding is identical to the word embedding, which avoids ambiguity and simplifies the task of finding semantically similar words. 
However, the lower value of $M(1)$ compared to $M(10), M(100), M(1000)$ illustrates the instability of representing the meaning of the axis by its top 1 word only.

\subsubsection{Qualitative observation}
In Table~\ref{tab:examples}, we observed examples of consecutive axes extracted from the Axis Tour embeddings with $k=100$. These include the 27th axis (the top five words are \emph{france}, \emph{french}, \emph{le}, \emph{paris}, and \emph{du}, so the \emph{France} axis), the 105th axis (the top five words are \emph{lawsuits}, \emph{lawsuit}, \emph{litigation}, \emph{suits}, and \emph{suit}, so the \emph{lawsuit} axis), and the 241st axis (the top five words are \emph{piano}, \emph{violin}, \emph{cello}, \emph{percussion}, and \emph{orchestral}, so the \emph{music instruments} axis). Table~\ref{tab:examples_k} shows the axes close to these axes for $k=1,10,1000$. Note that the axis indices change depending on the results of each Axis Tour. 

For each $k$, even the same axis shows significant differences in the nearby axes. For example, at $k=1$, in the bottom row, the meaning of the 227th axis is \emph{female}, but since the top 1 word is \emph{actress}, the axis is adjacent to the axes whose meanings are \emph{award} and \emph{movie}. This shows the disadvantage of the Axis Tour with $k=1$. At $k=10$, the top row shows that there are more axes related to \emph{personal names} than to \emph{countries} near the \emph{France} axis compared to $k=100$. At $k=1000$, it is interesting to see that the axes \emph{France} and \emph{musical instruments}, which are far apart at $k=100$, are close together.

Note that there is a selection bias in this comparison, as we use understandable examples for $k=100$ in Table~\ref{tab:examples} to compare with $k=1,10,1000$.

\begin{figure*}[t!]
    \centering
    \includegraphics[width=\linewidth]{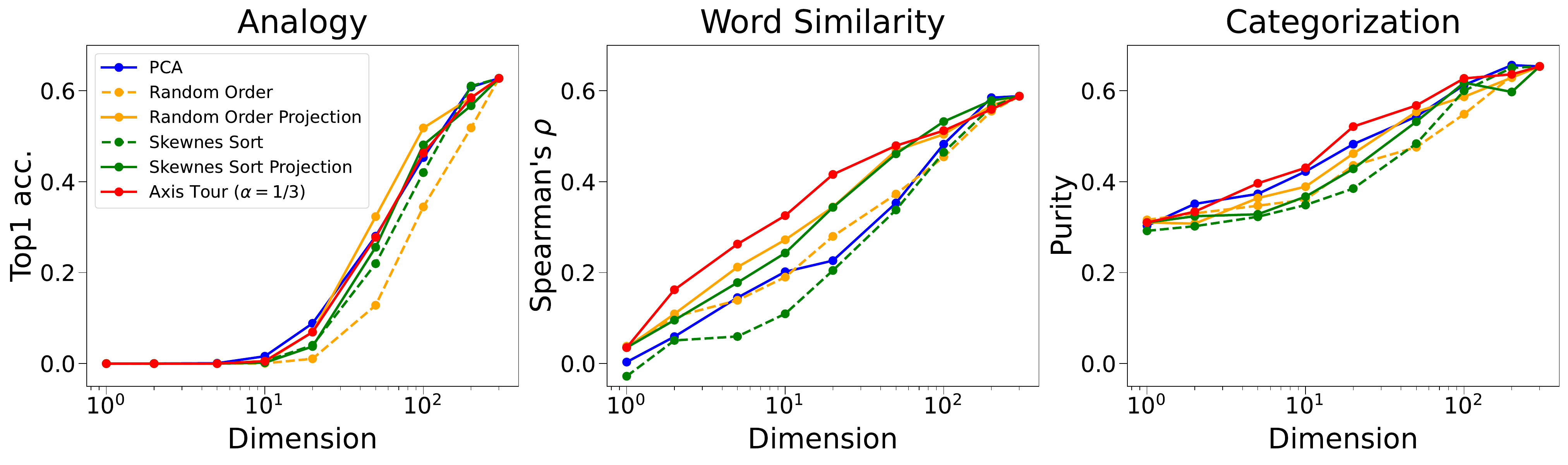}
    \caption{The performance of dimensionality reduction for the embeddings including Skewness Sort Projection and Random Order Projection with $\alpha=1/3$.
    Each value represents the average of 30 analogy tasks, 8 word similarity tasks, or 6 categorization tasks. 
}
    \label{fig:comp-baseline-projection}
\end{figure*}
\subsubsection{Dimensionality reduction}
Figure~\ref{fig:comp-k} shows the average of each task at $p=1,2,5,10,20,50,100,200,300$ for the PCA-transformed embeddings and the Axis Tour embeddings with $k=1,10,100,1000$ and $\alpha=1/3$.

From Fig.~\ref{fig:comp-k} we can see that the performance of the Axis Tour embeddings changes for each task, depending on $k$. For example, when comparing across all $k$, $k=1000$ shows good performance on categorization tasks and poor performance on word similarity tasks. In contrast, in lower dimensions, $k=1$ performs better on word similarity tasks than $k=1000$, but shows worse performance on categorization tasks. While $k=100$ does not always show the top performance in all three tasks, it consistently shows stable performance.

These results suggest that the quality of low-dimensional embeddings by the Axis Tour embeddings depends on $k$ for axis embedding.

\subsection{Dimensionality reduction by projection for Skewness Sort and Rand Order}~\label{app:dimreduction-by-projection}
The dimensionality reduction by projection in Section~\ref{sec:dim-reduction} can be applied not only to Axis Tour, but also to Skewness Sort and Random Order. Therefore, this section compares the dimensionality reduction method with those used for Skewness Sort and Random Order in Section~\ref{sec:downstream}, where the axes are selected sequentially from the first to last.

For the sake of explanation, let the embedding matrices for Skewness Sort and Random Order be denoted as $\mathbf{S}_\text{skew}$ and $\mathbf{S}_\text{rand}$, respectively. 
Let the skewness of the $\ell$-th axis for each be $\gamma_{\text{skew},\ell},\gamma_{\text{rand},\ell}\in\mathbb{R}$. By the definition of Skewness Sort, $\gamma_{\text{skew},\ell}\geq 0$. 
Similar to Section~\ref{sec:dim-reduction}, we consider reducing the dimensions from $d$ to $p(\leq d)$, divide $[d]$ into $p$ equal-length intervals, and use the index set for the $r$-th interval $I_{r} =\{a_r,\ldots,b_r\}\,(a_r,b_r\in [d], a_r\leq b_r)$. 

\subsubsection{Skewness Sort Projection}
Given $\gamma_{\text{skew},\ell}\geq 0$, similar to $\mathbf{f}_r=(f_r^{(\ell)})_{\ell=1}^d\in\mathbb{R}_{\geq 0}^d$ in (\ref{eq:def-f}), we define a unit vector $\mathbf{f}_{\text{skew},r}=(f_{\text{skew},r}^{(\ell)})_{\ell=1}^d\in\mathbb{R}_{\geq 0}^d$ for each $I_r$ as follows:
\begin{align}
   f_{\text{skew},r}^{(\ell)} = 
   \begin{cases} 
   \frac{\gamma_{\text{skew},\ell}^\alpha}{\sqrt{\sum_{m=a_r}^{b_r}\gamma_{\text{skew},m}^{2\alpha}}} & \text{for }\ell\in I_r \\
   0 & \text{otherwise},
   \end{cases}
   \label{eq:def-g}
\end{align}
where $\alpha \in \mathbb{R}_{\geq 0}$. 
We then define $\mathbf{F}_\text{skew} := [\mathbf{f}_{\text{skew},1}, \ldots, \mathbf{f}_{\text{skew},p}] \in \mathbb{R}^{d \times p}$ and obtain the $p$-dimensional embedding matrix $\mathbf{S}_\text{skew}\mathbf{F}_\text{skew}\in\mathbb{R}^{n\times p}$. We call these embeddings as \textbf{Skewness Sort Projection}.

\subsubsection{Random Order Projection}\label{app:random-order-projection}
For the Random Order embeddings, some axes may have $\gamma_{\text{rand},\ell}<0$. 
In this case, $\gamma_{\text{skew},\ell}^\alpha$ may not be defined in $\mathbb{R}$ for some $\alpha\in\mathbb{R}_{\geq 0}$. Consequently, we cannot directly use the definition of $\mathbf{f}_r$ as a vector for projection. Therefore, even for the Random Order embedding matrix, we flip the sign of each axis to ensure that the skewness is positive, thereby defining a matrix for projection\footnote{If the skewness of all axes in the Random Order matrix were initially set to be positive, this complex procedure could be avoided. However, our setup assumes that the sign and order of the axes are arbitrary after the ICA transformation, so the skewness of each axis of the matrix is not positive by default.}. To do this, we define a unit vector $\mathbf{f}_{\text{rand},r}=(f_{\text{rand},r}^{(\ell)})_{\ell=1}^d\in\mathbb{R}^d$ for each interval $I_r$ as follows:
\begin{align}
   f_{\text{rand},r}^{(\ell)} = 
   \begin{cases} 
   \frac{\text{sgn}(\gamma_{\text{rand},\ell})|\gamma_{\text{rand},\ell}|^\alpha}{\sqrt{\sum_{m=a_r}^{b_r}\gamma_{\text{rand},m}^{2\alpha}}} & \text{for }\ell\in I_r \\
   0 & \text{otherwise},
   \end{cases}
   \label{eq:def-h}
\end{align}
where $\alpha \in \mathbb{R}_{\geq 0}$ and $\text{sgn}(\cdot)$ represents the sign function. We then define $\mathbf{F}_\text{rand} := [\mathbf{f}_{\text{rand},1}, \ldots, \mathbf{f}_{\text{rand},p}] \in \mathbb{R}^{d \times p}$ and obtain the $p$-dimensional embedding matrix $\mathbf{S}_\text{rand}\mathbf{F}_\text{rand}\in\mathbb{R}^{n\times p}$. We call these embeddings as \textbf{Random Order Projection}.

\subsubsection{Results}
Similar to Axis Tour, we set the default value of $\alpha$ to $\alpha=1/3$ for both Skewness Sort Projection and Random Order Projection. Figure~\ref{fig:comp-baseline-projection} shows the average of each task at $p=1,2,5,10,20,50,100,200,300$ for the embeddings.

Dimensionality reduction for Axis Tour is still better than or nearly equivalent to Skewness Sort Projection and Random Order Projection for most dimensions in each task. Similar to the results in Section~\ref{sec:downstream}, these results also suggest the efficiency of dimensionality reduction for Axis Tour. 

Note that both Skewness Sort Projection and Random Order Projection show performance improvements over Skewness Sort and Random Order in most dimensions.
In particular, both show even better performance than PCA on word similarity tasks. This suggests that they are stronger baseline methods.

Interestingly, despite the lower performance of Random Order in analogy tasks, Random Order Projection performs slightly better than Axis Tour and PCA at $p=50, 100$. Considering that Axis Tour showed only equivalent performance to PCA for even different $\alpha$ in Fig.~\ref{fig:comp-alpha} or $k$ in Fig.~\ref{fig:comp-k}, this result suggests that by applying clustering or similar techniques to axis embeddings, we may obtain more effective low-dimensional vectors.

\subsubsection{Setting of $p=d\,(=300)$}
When $p=d$, similar to Skewness Sort and Random Order, the performance of Skewness Sort Projection and Random Order Projection for each task is equivalent to that of PCA.

For Skewness Sort Projection, from (\ref{eq:def-g}) and the discussion in Appendix~\ref{app:def-f}, it can be shown that when $p=d$, we have $\mathbf{F}_\text{skew} = \mathbf{I}$, and therefore $\mathbf{S}_\text{skew}\mathbf{F}_\text{skew} = \mathbf{S}_\text{skew}\mathbf{I} = \mathbf{S}_\text{skew}$.

In preparation for Random Order Projection, we denote $\mathbf{S}_{\text{rand},\geq 0}$ as the matrix obtained by flipping the sign of each axis in $\mathbf{S}_\text{rand}$ so that the skewness is positive. 
It then follows from the discussion in Appendix~\ref{app:random-order-projection} that, similar to the Skewness Sort Projection, $\mathbf{S}_\text{rand}\mathbf{F}_\text{rand} = \mathbf{S}_{\text{rand},\geq 0}\mathbf{I} = \mathbf{S}_{\text{rand},\geq 0}$ when $p=d$. 
Thus, we can see that the performance of $\mathbf{S}_{\text{rand},\geq 0}$ is equivalent to that of $\mathbf{S}_\text{rand}$ since $\mathbf{S}_{\text{rand},\geq 0}$ is derived from $\mathbf{S}_\text{rand}$ by applying the orthogonal matrix to flip the sign of each axis.

\begin{table*}[t!]
\tiny
\centering
\begin{subtable}{\textwidth}
\centering
\begin{tabular}{@{\hspace{1em}}c@{\hspace{1em}}c@{\hspace{1em}}c@{\hspace{1em}}c@{\hspace{1em}}c@{\hspace{1em}}c@{\hspace{1em}}c@{\hspace{1em}}c@{\hspace{1em}}c@{\hspace{1em}}c@{\hspace{1em}}}
\toprule
         50 &         51 &        52 &           53 &            54 &     55 &       56 &       57 &       58 &        59 \\
\midrule
frustration &    mindset &   aspects &     problems & deteriorating &   rise &     went &      was &     sees &     given \\
      anger & philosophy &   factors & difficulties &      evolving & rising &  slipped &      has &     gets &   receive \\
 resentment &  attitudes & variables &    hardships &        urgent &  spike & stumbled & requires & embraces &  received \\
indignation &   attitude &    facets & shortcomings &       rapidly &   soar &      ran &      had &  creates &      give \\
unhappiness &  worldview &  elements &     troubles &          dire &  rises &  drifted &       is & inspires & receiving \\
\midrule
\midrule
        100 &       101 &        102 &       103 &       104 &     105 &   106 &      107 &          108 &       109 \\
\midrule
   gameplay &  gambling &      filly & livestock &  planting &   birds &   dog &   babies &       father &    people \\
multiplayer &    casino &       colt &    cattle & seedlings & species &  dogs &  infants &       mother & americans \\
     capcom &   casinos & \#\#/\#\#-mile &      farm &    blooms &    fish & puppy &  newborn &     siblings &   britons \\
     gamers & blackjack &       mare &     dairy &     vines &    bird &   pet & newborns &          son &   patrons \\
      zelda &  gamblers &       gr.3 &      cows &     shrub & turtles &  cats &     baby & grandparents &   viewers \\
\midrule
\midrule
        150 &         151 &          152 &     153 &      154 &       155 &       156 &             157 &         158 &          159 \\
\midrule
     dh'ing &     pitched &      assists &     nba &  pointer &    header &   everton &  cambridgeshire &     milford &    allegheny \\
batting.\#\#\# &  \#/\#-inning &       steals &   rondo &  timeout &    footed &   arsenal &   hertfordshire &    westerly &   harrisburg \\
      ss/2b &     innings &       points & pistons &     foul & deflected & tottenham &     oxfordshire & marlborough & pennsylvania \\
      of/1b & \#\#/\#innings & .\#\#\#\_\#\#-\#-\#\# &    nets &   jumper &    angled &    fulham &   staffordshire & bridgewater &          pa. \\
      2b/of &      outing &     rebounds & nuggets & halftime &    keeper &   anfield & buckinghamshire &     amherst &         penn \\
\midrule
\midrule
          200 &       201 &      202 &     203 &         204 &       205 &          206 &         207 &        208 &           209 \\
\midrule
     proteins &    psn\#\#\# &    pills &    beer &  restaurant &     sauce &         diet &  instructor &     school &     sociology \\
      protein &   map\#\#\#\# &    drugs & whiskey &       grill &      soup &        diets &       teach &   teachers &  anthropology \\
       enzyme &   ap\#\#\#\#\# &   heroin &   lager &      bistro &     pasta &      dieting & instruction &    teacher & undergraduate \\
     molecule &  gw\#\#\#\#\#\# & morphine &   beers &      diners &    sauces &    nutrition &   beginners & elementary &  postgraduate \\
intracellular & ly\#\#\#\#\#\#\# &  ecstasy & brewery & restaurants & meatballs & carbohydrate &    training &     pupils &           phd \\
\midrule
\midrule
      250 &        251 &       252 &        253 &       254 &    255 &        256 &     257 &     258 &          259 \\
\midrule
  workers &        gdp &    budget &   payments &       tax & senate &  primaries &    bush &     mps &      ontario \\
\#.\#\#/hour & economists &      cuts &    payment &     taxes &   bill & democratic &   obama &  labour &           bc \\
 \#\#.\#\#/hr &  inflation &   deficit &        pay & surcharge & d-tex. &     voters & clinton & commons & saskatchewan \\
   9/hour &        cpi &   budgets &   paycheck &       gst &   rep. &   election &  cheney &    tory &      alberta \\
    wages &        ism & austerity & deductible &    levies &  r-md. & republican &   putin &  tories &       canada \\
\bottomrule
\end{tabular}
\caption{word2vec}
\label{tab:word2vec-examples}
\end{subtable}
\vspace{2em}

\begin{subtable}{\textwidth}
\centering
\begin{tabular}{@{\hspace{1em}}c@{\hspace{1em}}c@{\hspace{1em}}c@{\hspace{1em}}c@{\hspace{1em}}c@{\hspace{1em}}c@{\hspace{1em}}c@{\hspace{1em}}c@{\hspace{1em}}c@{\hspace{1em}}c@{\hspace{1em}}}
\toprule
      50 &         51 &        52 &       53 &       54 &         55 &       56 &      57 &       58 &        59 \\
\midrule
the\_1336 &      air\_7 &    rail\_0 &   of\_704 & emerge\_2 &     face\_2 &  make\_21 &   see\_9 &  truth\_0 &   takes\_1 \\
the\_1335 & aviation\_0 & service\_0 &  the\_900 &   to\_350 &   emerge\_0 &   look\_2 &  know\_3 &  truth\_5 &     a\_202 \\
   '\_306 &      air\_5 & express\_0 & the\_1609 &    a\_281 &     find\_1 & double\_5 &  knew\_0 &  truth\_2 &    maps\_0 \\
   s\_242 &    plane\_1 &   the\_191 &  the\_901 &  the\_723 & yourself\_0 &   sure\_2 & aware\_5 & share\_10 &   sweep\_0 \\
  it\_137 &      jet\_0 &   trans\_0 & the\_1159 &   turn\_3 &     to\_189 &  make\_26 &  know\_5 &  truth\_1 & chronic\_0 \\
\midrule
\midrule
        100 &       101 &     102 &      103 &     104 &     105 &       106 &     107 &         108 &         109 \\
\midrule
liverpool\_1 &  lineup\_1 &  some\_8 & little\_2 &  five\_0 &  four\_6 &       2\_9 &  both\_1 &  between\_12 & agreement\_0 \\
    annum\_1 &  group\_16 & some\_43 &    bit\_2 &  two\_14 &   six\_0 &    two\_47 & both\_10 &   between\_9 &       an\_82 \\
   shower\_0 &    band\_1 & some\_21 &    few\_8 &  two\_25 & four\_15 &       2\_1 &  both\_9 & relations\_1 &    struck\_1 \\
      ,\_755 &   group\_2 &  many\_5 &    bit\_1 &  few\_12 &  five\_1 & second\_11 &  two\_36 &  between\_10 &   reached\_1 \\
liverpool\_2 & number\_10 & some\_27 &    bit\_0 & three\_9 &  five\_6 &     two\_0 &  both\_6 &   bonding\_0 &     talks\_2 \\
\midrule
\midrule
   150 &    151 &    152 &    153 &    154 &   155 &  156 &      157 &       158 &     159 \\
\midrule
 ,\_115 &  ,\_432 & ,\_1022 & .\_1512 &  .\_114 & a\_159 &  r\_0 &  asher\_0 & michael\_1 &  emma\_0 \\
 ,\_701 &  ,\_901 &  ,\_181 &  .\_895 & .\_1525 &   r\_1 & d\_12 & hilton\_0 &   james\_0 & cindy\_0 \\
 ,\_172 &  ,\_876 &  ,\_180 & .\_1260 &  .\_110 &   h\_1 &  p\_8 &   colt\_0 &   peter\_1 &  mare\_0 \\
,\_1526 & ,\_1561 &  ,\_408 & .\_1339 & .\_1575 & .\_381 & m\_12 &  \#\#nie\_4 &    mike\_3 &  kate\_0 \\
,\_1527 & ,\_1367 & ,\_1024 & .\_1055 &  .\_492 & .\_600 &  v\_0 &  jesse\_0 &    mike\_1 &  beth\_0 \\
\midrule
\midrule
       200 &            201 &       202 &       203 &      204 &           205 &        206 &        207 &        208 &         209 \\
\midrule
software\_1 &     services\_3 & limited\_1 &   least\_0 & cannot\_2 &       also\_53 &  schools\_5 & bachelor\_0 &  workers\_4 &     users\_2 \\
software\_2 & capabilities\_0 &   the\_271 &    must\_2 &    a\_276 &         -\_613 & students\_4 &   degree\_0 &     jobs\_2 & investors\_7 \\
hardware\_0 &     services\_2 &    are\_26 &      a\_98 &   any\_10 &       cents\_2 &  schools\_1 &    the\_739 &  workers\_1 & investors\_1 \\
software\_0 &     services\_4 &    to\_141 & [SEP]\_190 &   in\_256 & government\_29 &  student\_2 &    major\_8 &  workers\_2 &     users\_0 \\
    free\_3 &     services\_7 &     a\_116 &    of\_116 &   to\_300 &         20\_14 &   school\_2 & subjects\_0 & workers\_10 &     users\_3 \\
\midrule
\midrule
    250 &     251 &       252 &     253 &    254 &         255 &        256 &        257 &      258 &         259 \\
\midrule
 time\_9 & year\_57 &    hour\_3 & year\_51 & day\_14 & thursday\_14 &   month\_10 & european\_1 & world\_27 &       \#\#8\_3 \\
based\_5 & year\_68 &   weeks\_3 & year\_37 &  day\_8 &   tuesday\_6 &    week\_10 & european\_6 & world\_31 &    their\_17 \\
based\_4 & year\_56 & minutes\_1 & year\_15 &  day\_7 &    monday\_8 &    year\_67 & european\_0 &  world\_7 &   nations\_0 \\
lined\_2 & year\_59 &   hours\_4 & year\_72 & day\_20 & wednesday\_3 &     week\_0 & european\_3 & world\_17 & countries\_0 \\
level\_0 & year\_69 &    hour\_5 & year\_13 &  day\_2 &  thursday\_0 & saturday\_0 & european\_5 &  world\_5 &   [SEP]\_249 \\
\bottomrule
\end{tabular}
\caption{BERT}
\label{tab:BERT-examples}
\end{subtable}
\vspace{2em}
\caption{Semantic continuity of axes by Axis Tour for normalized ICA-transformed embeddings. We apply Axis Tour to 300-dimensional word2vec and 768-dimensional BERT. For the 50th, 100th, 150th, 200th, and 250th axes, we extract ten consecutive axes from each of these axes and display the top five words for each of the extracted axes.}
\label{tab:word2vec-BERT-examples}
\end{table*}

\begin{figure*}[!t]
    \centering
    \begin{minipage}{0.33\linewidth}
        \centering
        \includegraphics[width=\linewidth]{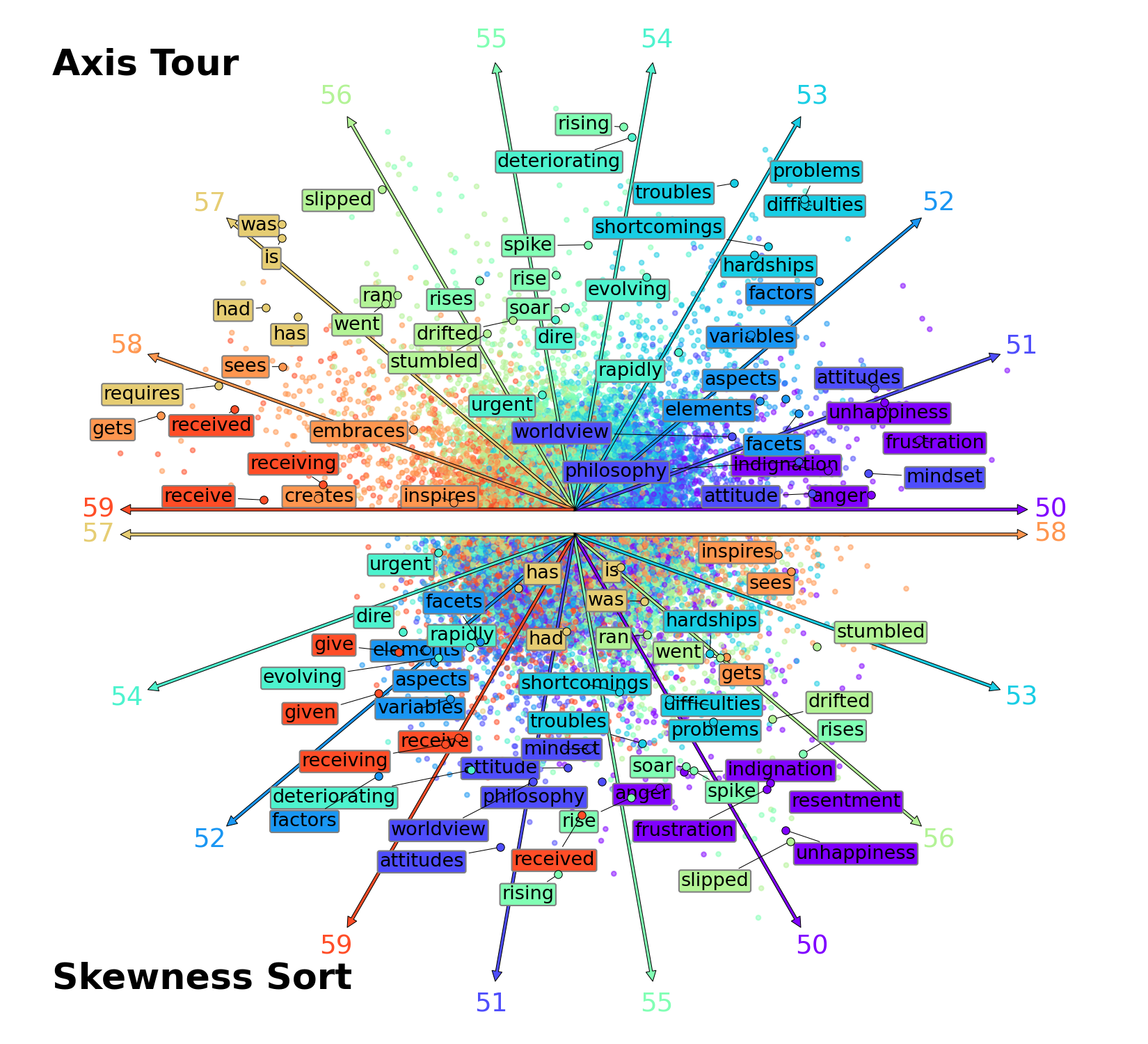}
        \subcaption{The 50th axis to 59th axis}
        \label{fig:word2vec-50}
    \end{minipage}\hfill
    \begin{minipage}{0.33\linewidth}
        \centering
        \includegraphics[width=\linewidth]{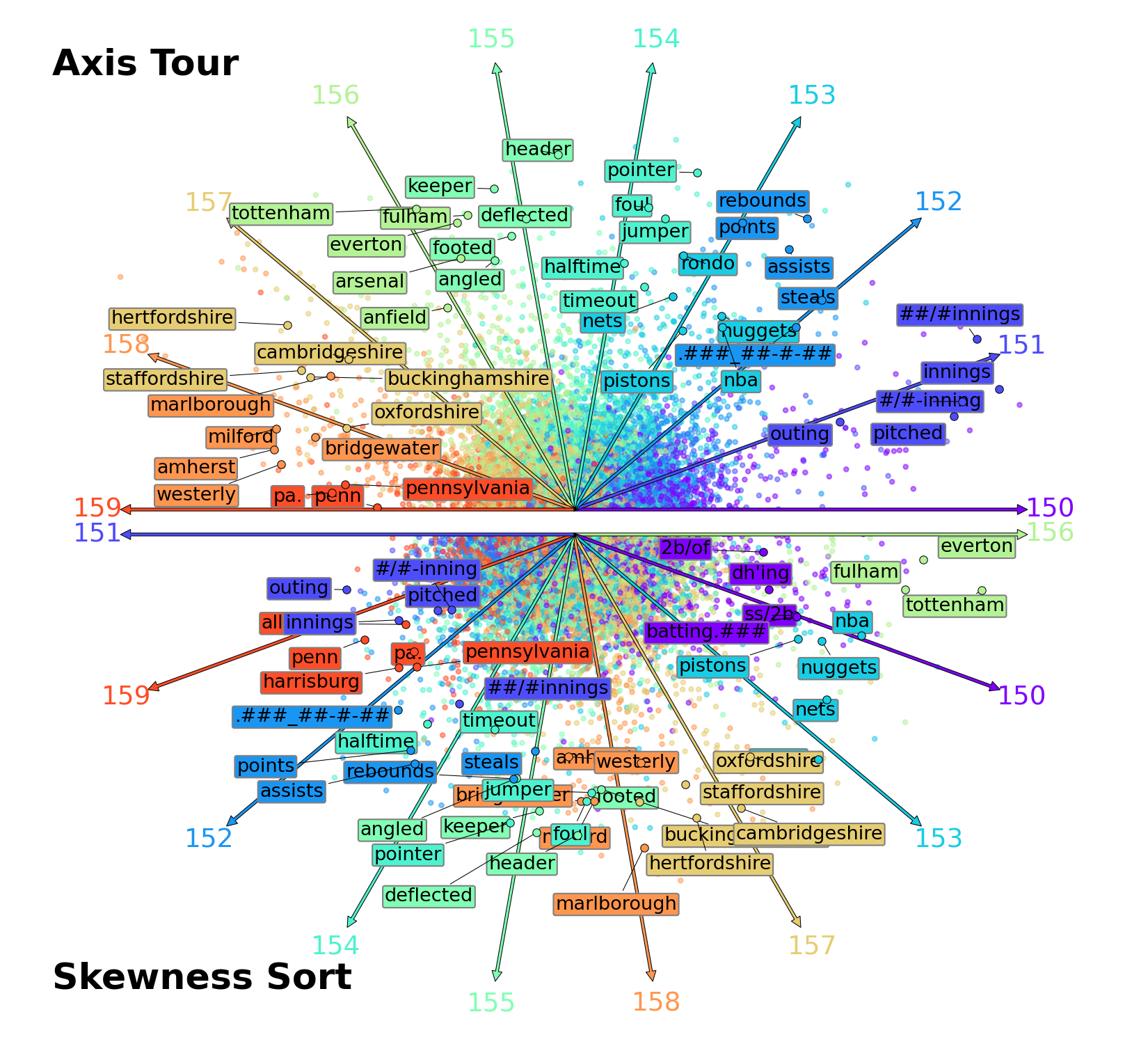}
        \subcaption{The 150th axis to 159th axis}
        \label{fig:word2vec-150}
    \end{minipage}
    \begin{minipage}{0.33\linewidth}
        \centering
        \includegraphics[width=\linewidth]{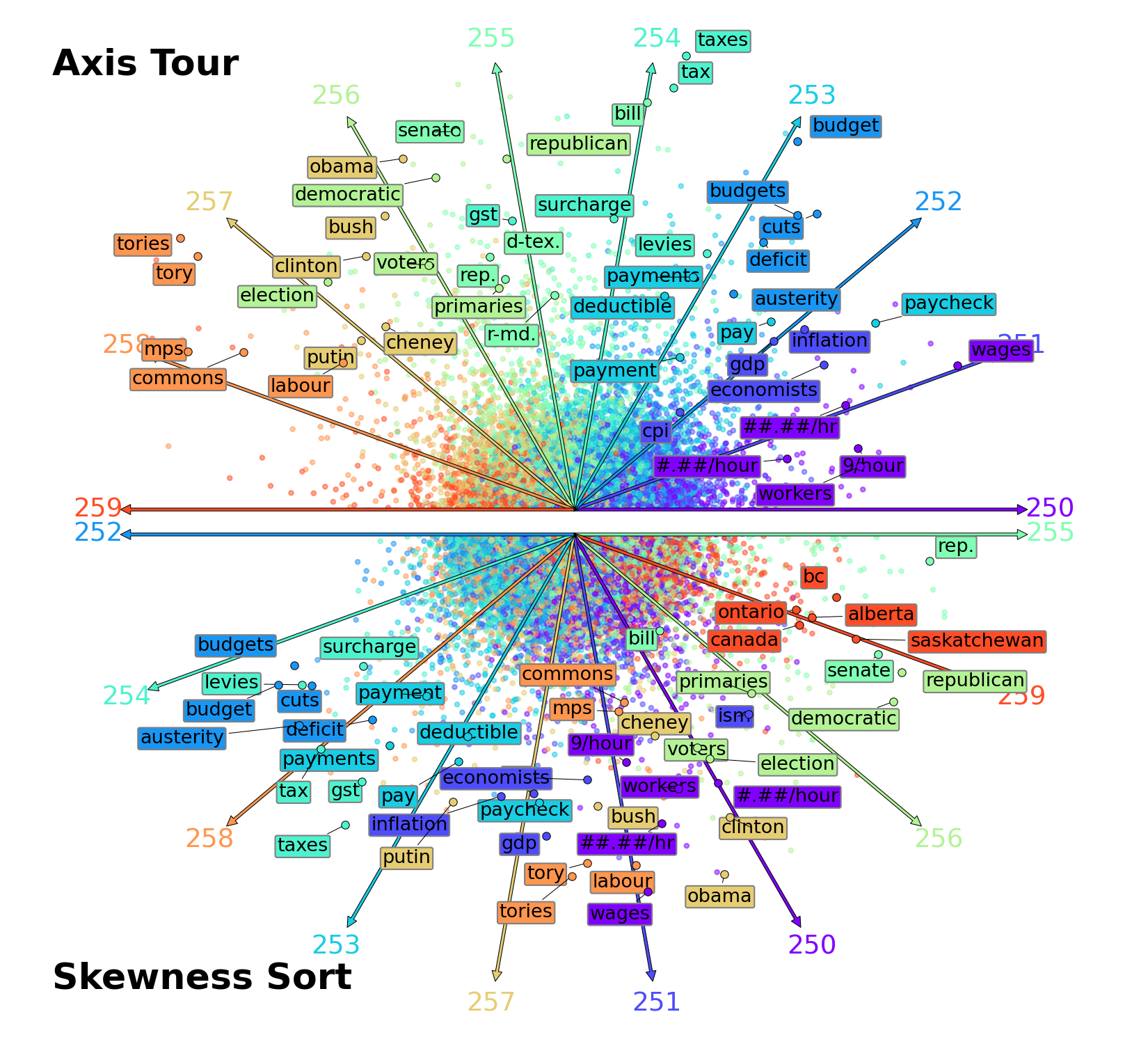}
        \subcaption{The 250th axis to 259th axis}
        \label{fig:word2vec-250}
    \end{minipage}
    \caption{
For word2vec, the scatterplots of the two-dimensional projections for the axes from the 50th to the 59th, from the 150th to the 159th,  and from the 250th to the 259th in Table~\ref{tab:word2vec-examples}.
}
    \label{fig:word2vec-plot}
\end{figure*}

\begin{figure*}[!t]
    \centering
    \begin{minipage}{0.33\linewidth}
        \centering
        \includegraphics[width=\linewidth]{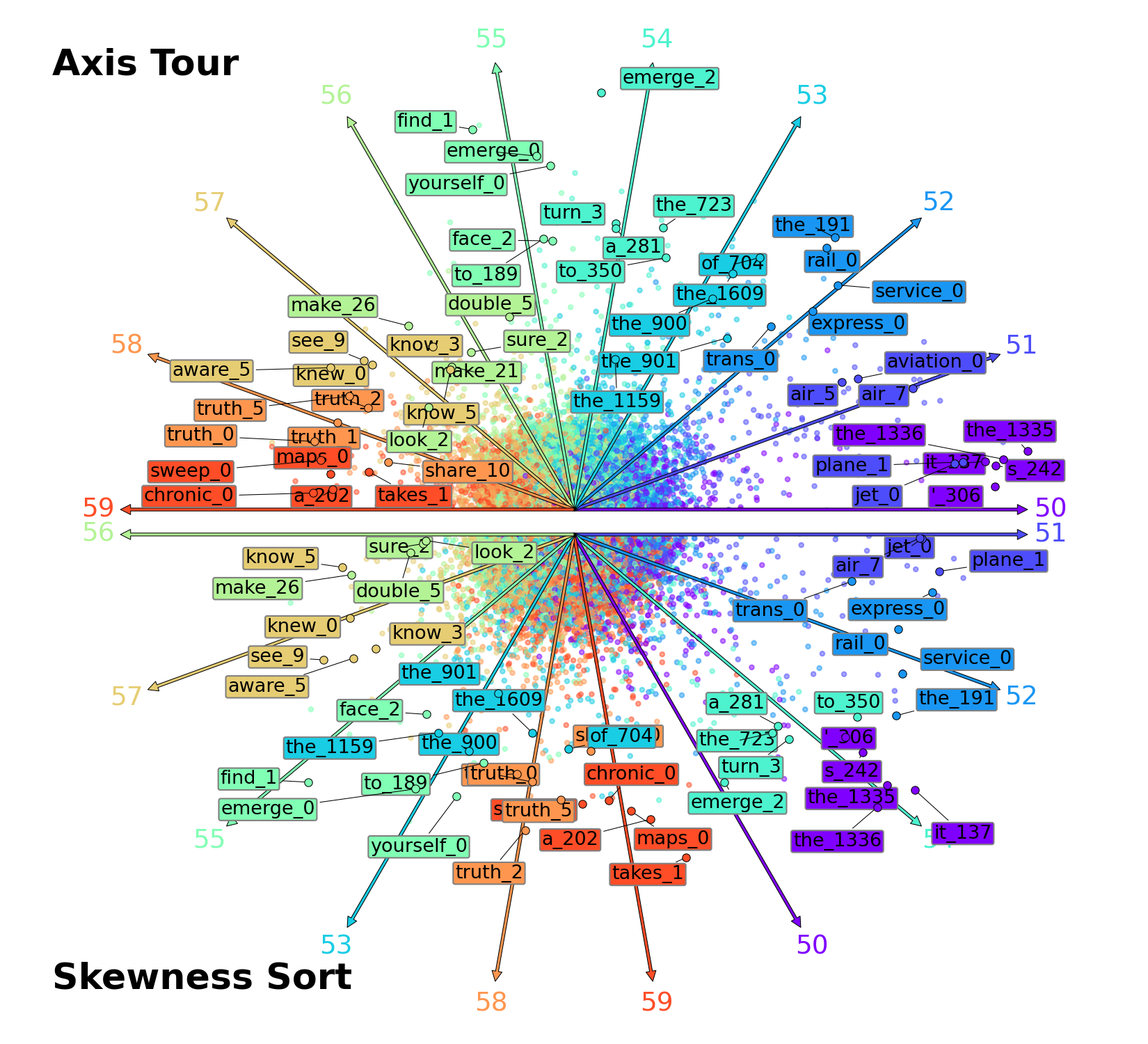}
        \subcaption{The 50th axis to 59th axis}
        \label{fig:BERT-50}
    \end{minipage}\hfill
    \begin{minipage}{0.33\linewidth}
        \centering
        \includegraphics[width=\linewidth]{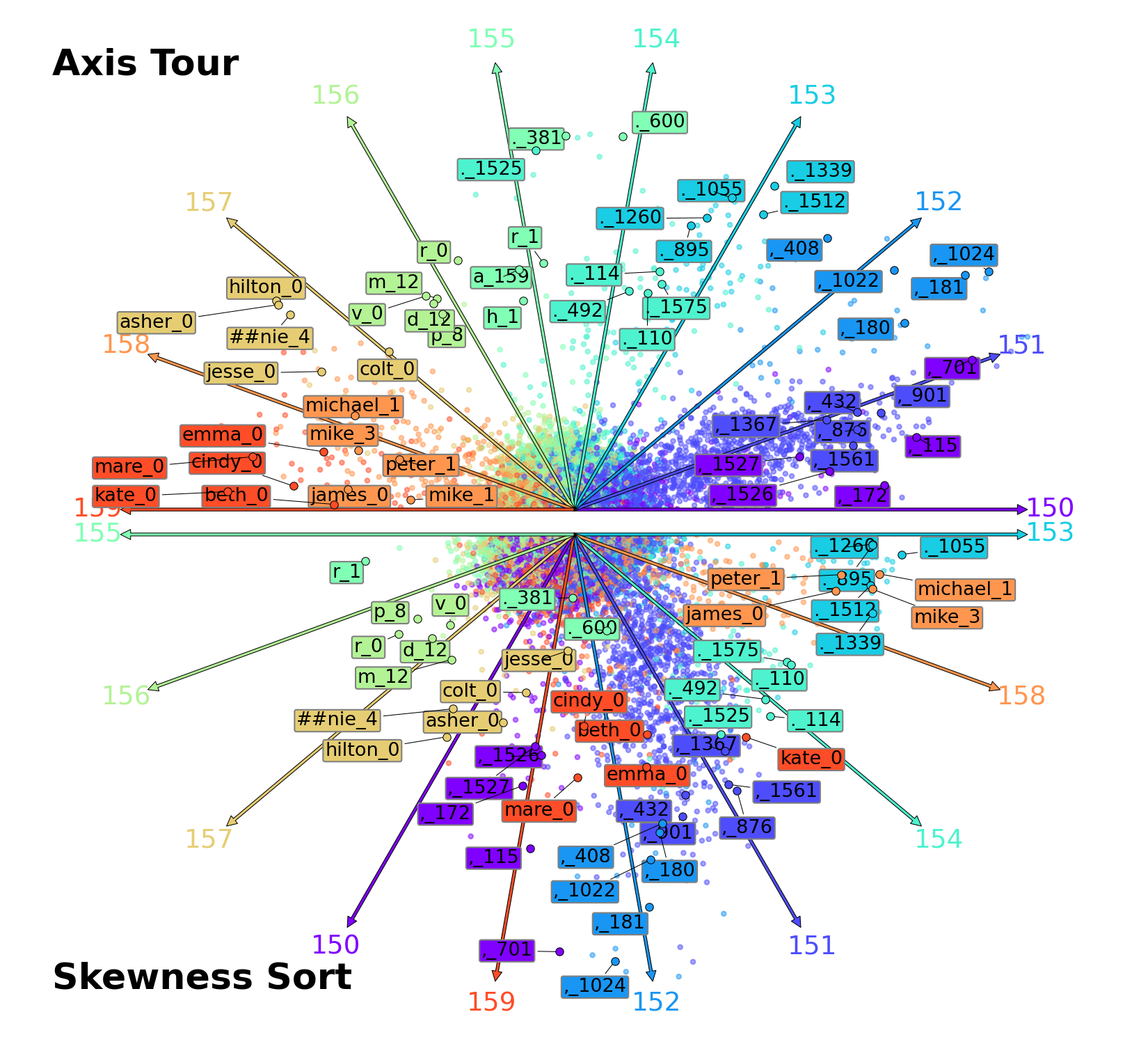}
        \subcaption{The 150th axis to 159th axis}
        \label{fig:BERT-150}
    \end{minipage}
    \begin{minipage}{0.33\linewidth}
        \centering
        \includegraphics[width=\linewidth]{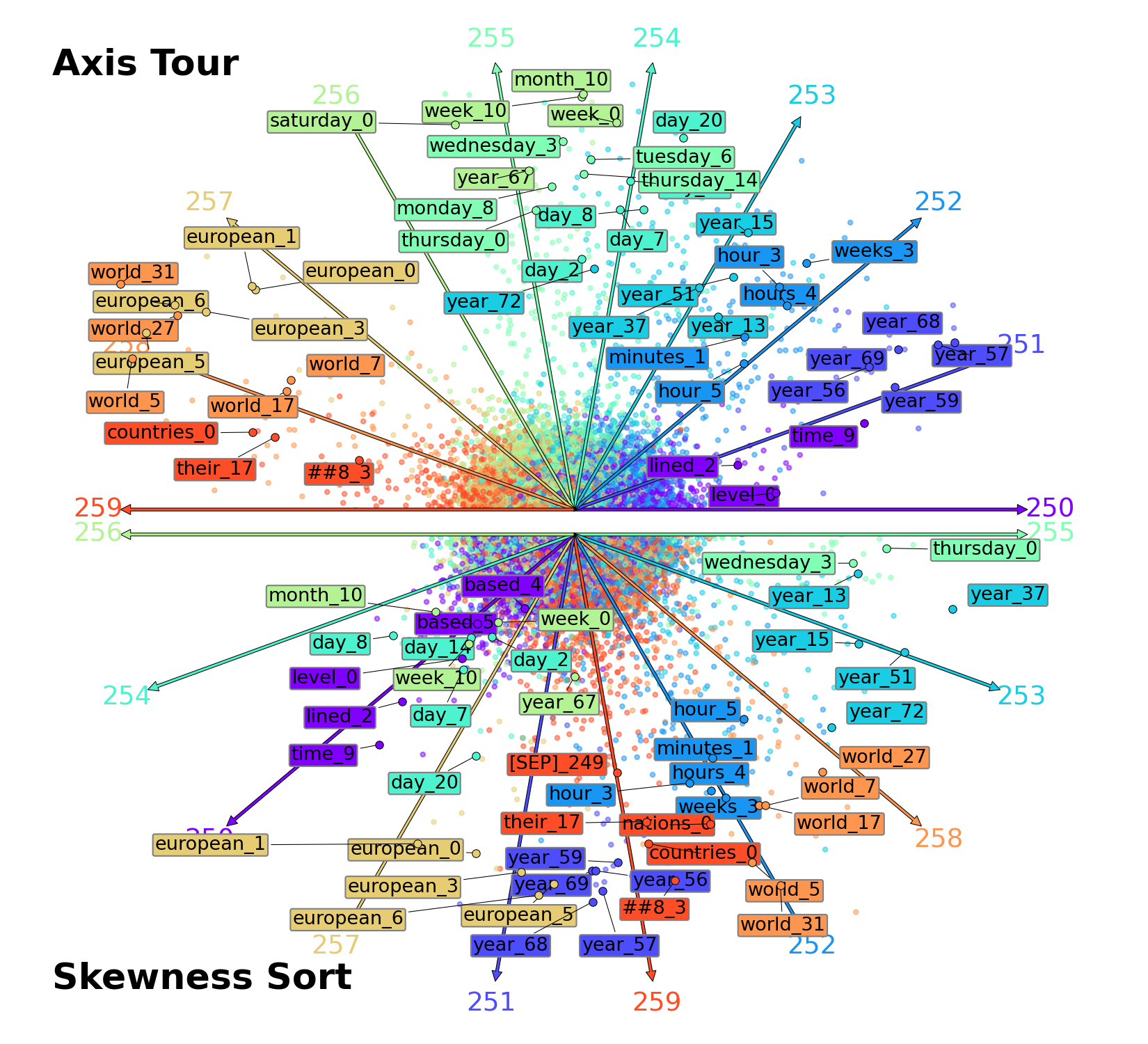}
        \subcaption{The 250th axis to 259th axis}
        \label{fig:BERT-250}
    \end{minipage}
    \caption{
For BERT, the scatterplots of the two-dimensional projections for the axes from the 50th to the 59th, from the 150th to the 159th, and from the 250th to the 259th in Table~\ref{tab:BERT-examples}.
}
    \label{fig:bert-plot}
\end{figure*}

\begin{table}[t]
\centering
\begin{tabular}{lrrrr}
\toprule
& \multicolumn{2}{c}{Axis Tour} &  \multicolumn{2}{c}{Skewness Sort} \\
\cmidrule(lr){2-3}
\cmidrule(lr){4-5}
Fig. &  $d_I$&  $c_I$  & $d_I$ &  $c_I$ \\
\midrule
\ref{fig:word2vec-50} & 0.70 & 0.30 & 0.55 & 0.00\\
\ref{fig:word2vec-150} & 0.79 & 0.34 & 0.68 & 0.13 \\
\ref{fig:word2vec-250} & 0.80 & 0.27 & 0.71 & 0.05 \\
\ref{fig:BERT-50} & 0.60 & 0.28 & 0.59 & 0.11\\
\ref{fig:BERT-150} & 0.71 &0.33  & 0.60 &0.06 \\
\ref{fig:BERT-250} & 0.64 & 0.31 & 0.53 & 0.07\\
\bottomrule
\end{tabular}
\caption{
The values of $d_I$ and $c_I$ for Figs.~\ref{fig:word2vec-plot} and~\ref{fig:bert-plot}.
}
\label{tab:fig-word2vec-bert-results}
\end{table}

\subsection{Other embeddings}\label{app:other-embeddings}
In Section~\ref{sec:experiments}, we used 300-dimensional GloVe\footnote{The embeddings can be downloaded here: \url{https://nlp.stanford.edu/data/glove.6B.zip}.} with $n=400{,}000$. 
In this section, we extend our experiments to other embeddings.  
We used 300-dimensional word2vec\footnote{\url{https://code.google.com/archive/p/word2vec/}}~\cite{DBLP:journals/corr/abs-1301-3781} as static embeddings and 768-dimensional BERT\footnote{\url{https://huggingface.co/bert-base-uncased}}~\cite{DBLP:conf/naacl/DevlinCLT19} from the Hugging Face transformers library~\cite{DBLP:conf/emnlp/WolfDSCDMCRLFDS20} as dynamic embeddings.

For word2vec, given the original vocabulary size of three million, we selected only the top $40{,}000$ words based on frequency after converting all words to lowercase. The word frequency information was obtained using wordfreq~\cite{robyn_speer_2022_7199437}.

For BERT, we first input sentences from the One Billion Word Benchmark~\cite{DBLP:conf/interspeech/ChelbaMSGBKR14} into BERT, and then used the first $40{,}000$ tokens, including special tokens like [CLS] and [SEP]. It is important to note that the embeddings are different even for identical tokens, so we differentiated tokens like \emph{cat} as \emph{cat}\_0, \emph{cat}\_1, and so on.

Similar to GloVe, for both word2vec and BERT, we set $k=100$ as the hyperparameter of the axis embedding for Axis Tour.

\subsubsection{Qualitative observation}
Tables~\ref{tab:word2vec-examples} and~\ref{tab:BERT-examples} show the examples of the axes of the Axis Tour embeddings for word2vec and BERT, respectively.
We can also see the semantic continuity of the axes of the Axis Tour embeddings for word2vec in Table~\ref{tab:word2vec-examples}, just as we observed the semantic continuity of those for GloVe in Tables~\ref{tab:0-119},~\ref{tab:120-239} and~\ref{tab:240-299}.
For BERT, the semantic continuity of the axes is also observed in Table~\ref{tab:BERT-examples}, although there are axes whose top tokens are the identical tokens. 
This is due to dynamic embeddings and differs from static embeddings as in GloVe and word2vec.

\subsubsection{Scatterplots of Table~\ref{tab:word2vec-BERT-examples}}
Figures~\ref{fig:word2vec-plot} and~\ref{fig:bert-plot} show the scatterplots of the two-dimensional projections for the axes of the Axis Tour embeddings for word2vec and BERT in Tables~\ref{tab:word2vec-examples} and~\ref{tab:BERT-examples}, respectively.
We used the procedure described in Appendix~\ref{app:fig-process}. 
Similar to Figs.~\ref{fig:intro} and~\ref{fig:examples}, for both word2vec and BERT, we can see that the top words of the axes are farther from the origin than those of the Skewness Sort, and the meanings of the adjacent axes change continuously. 

For Figs.~\ref{fig:word2vec-plot} and~\ref{fig:bert-plot}, Table~\ref{tab:fig-word2vec-bert-results} shows the values of $d_I$ and $c_I$ as the evaluation metrics defined in Appendix~\ref{app:evalmetric}. 
Similar to Table~\ref{tab:fig-results}, for both word2vec and BERT, Axis Tour shows higher values for both $d_I$ and $c_I$ than Skewness Sort, indicating better projection quality.

\subsubsection{Cosine similarity between adjacent axis embeddings}
\begin{figure}[!t]
\centering
\begin{subfigure}{\columnwidth}
    \includegraphics[width=\textwidth]{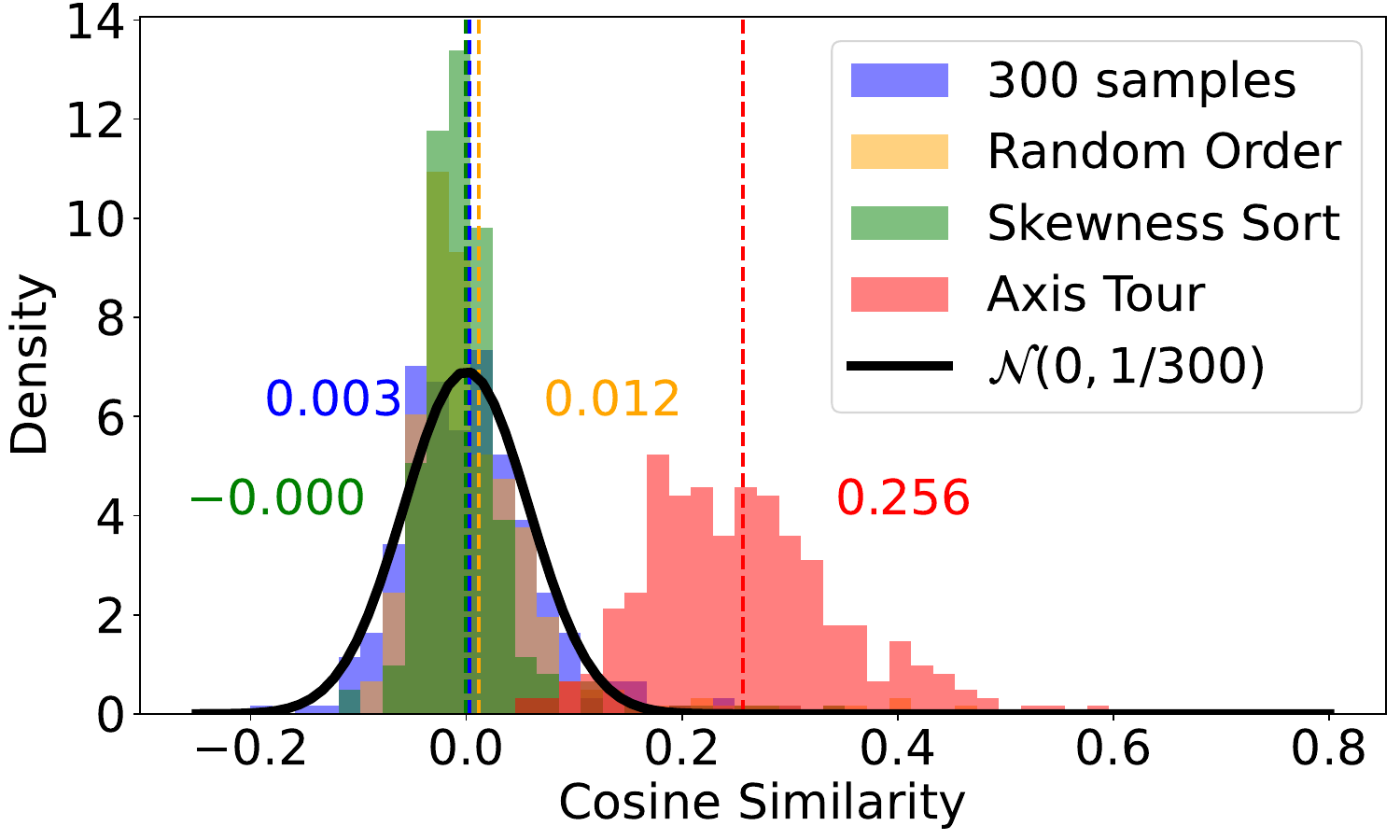}
    \subcaption{word2vec}
    \label{fig:word2vec-cossim-hist}
\end{subfigure}
\begin{subfigure}{\columnwidth}
    \includegraphics[width=\textwidth]{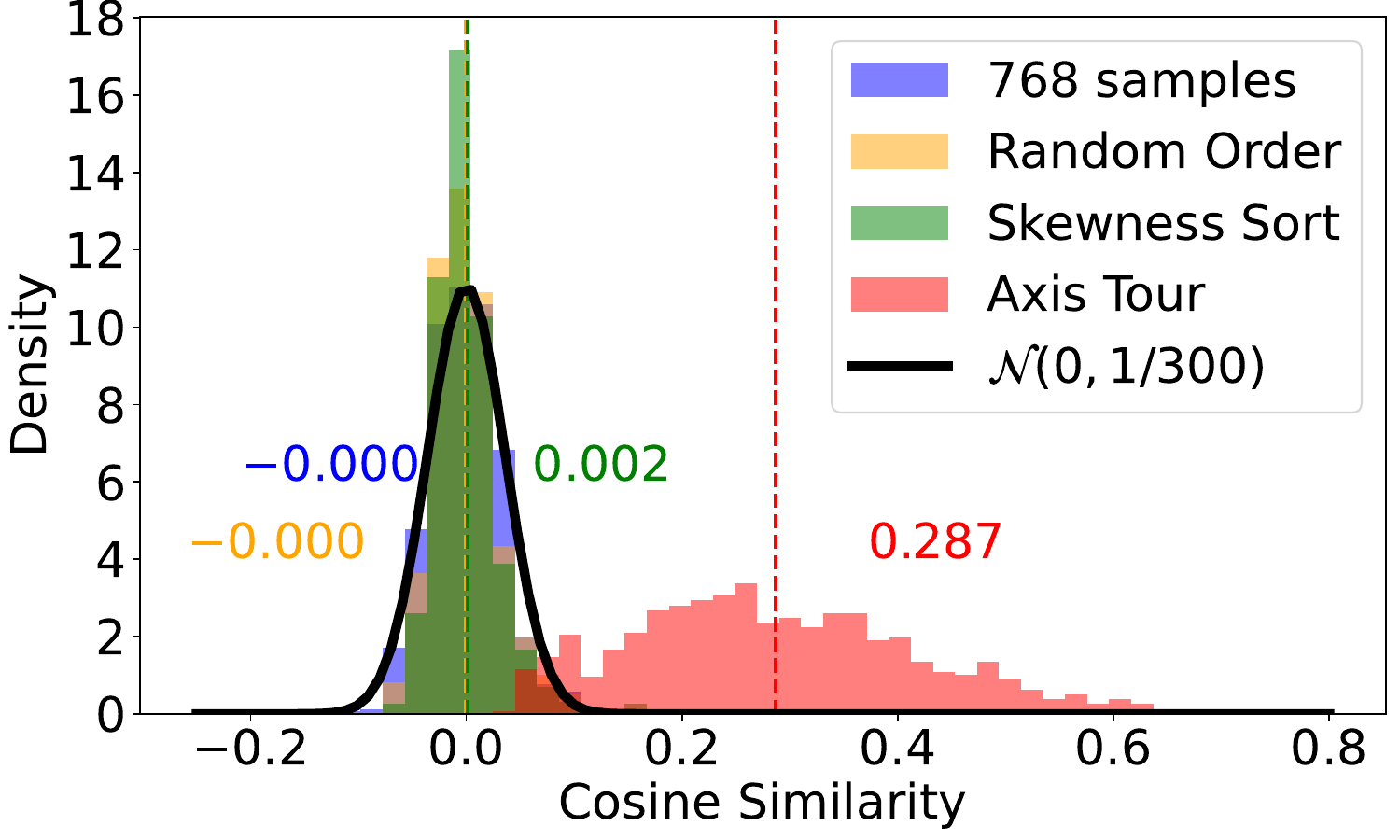}
    \subcaption{BERT}
    \label{fig:bert-cossim-hist}
\end{subfigure}
    \caption{
Histograms of $\cos{(\mathbf{v}_\ell,\mathbf{v}_{\ell+1})}$ for word2vec and BERT.
For BERT, we sampled 768 random words and drew the normal distribution $\mathcal{N}(0,1/768)$ instead of 300 words and the normal distribution $\mathcal{N}(0,1/300)$.
The rest of the procedure is the same as in Fig.~\ref{fig:cos-hist}.
}
\label{fig:word2vec-bert-cossim-hist}
\end{figure}

Figure~\ref{fig:word2vec-bert-cossim-hist} shows the histograms of $\cos(\mathbf{v}_\ell,\mathbf{v}_{\ell+1})$ of the Axis Tour embeddings and the baseline embeddings for both word2vec and BERT.
Similar to the results for GloVe in Fig.~\ref{fig:cos-hist}, the values of $\cos(\mathbf{v}_{\ell},\mathbf{v}_{\ell+1})$ are consistently higher in Axis Tour, while this trend is not observed in the other baselines. 

\begin{figure}[!t]
\centering
\begin{subfigure}{\columnwidth}
    \includegraphics[width=\textwidth]{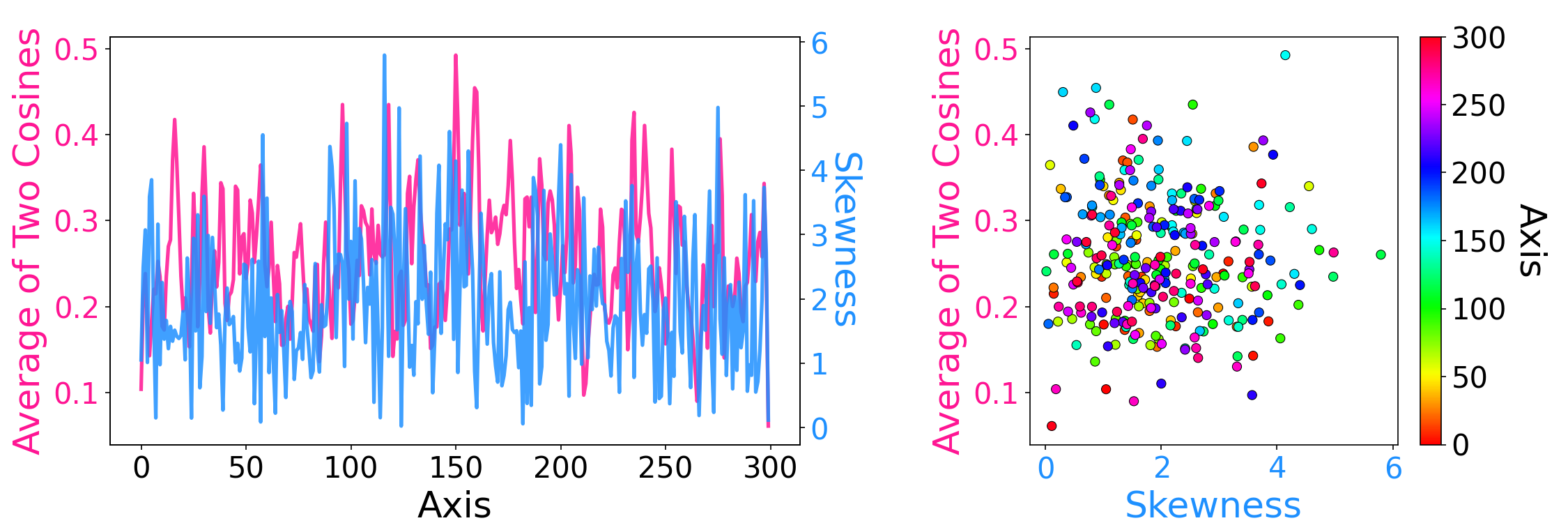}
    \subcaption{Axis Tour}
    \label{fig:word2vec-cos-skew-plot-axistour}
\end{subfigure}
\begin{subfigure}{\columnwidth}
    \includegraphics[width=\textwidth]{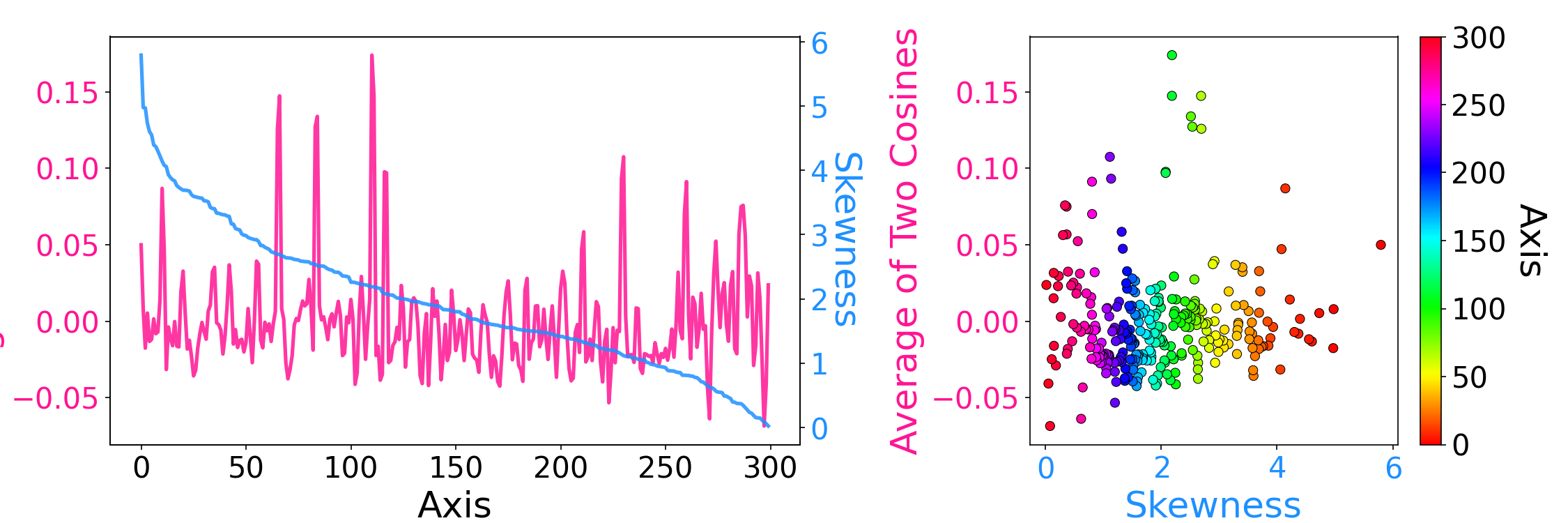}
    \subcaption{Skewness Sort}
    \label{fig:word2vec-cos-skew-plot-skewnessort}
\end{subfigure}
    \caption{
For word2vec, the relationship between the skewness and the average of consecutive two cosines for (a) Axis Tour and (b) Skewness Sort.
Spearman's rank correlation is $-0.04$ for Axis Tour, while it is $0.08$ for Skewness Sort.
}
\label{fig:word2vec-cos-skew-plot}
\end{figure}

\begin{figure}[!t]
\centering
\begin{subfigure}{\columnwidth}
    \includegraphics[width=\textwidth]{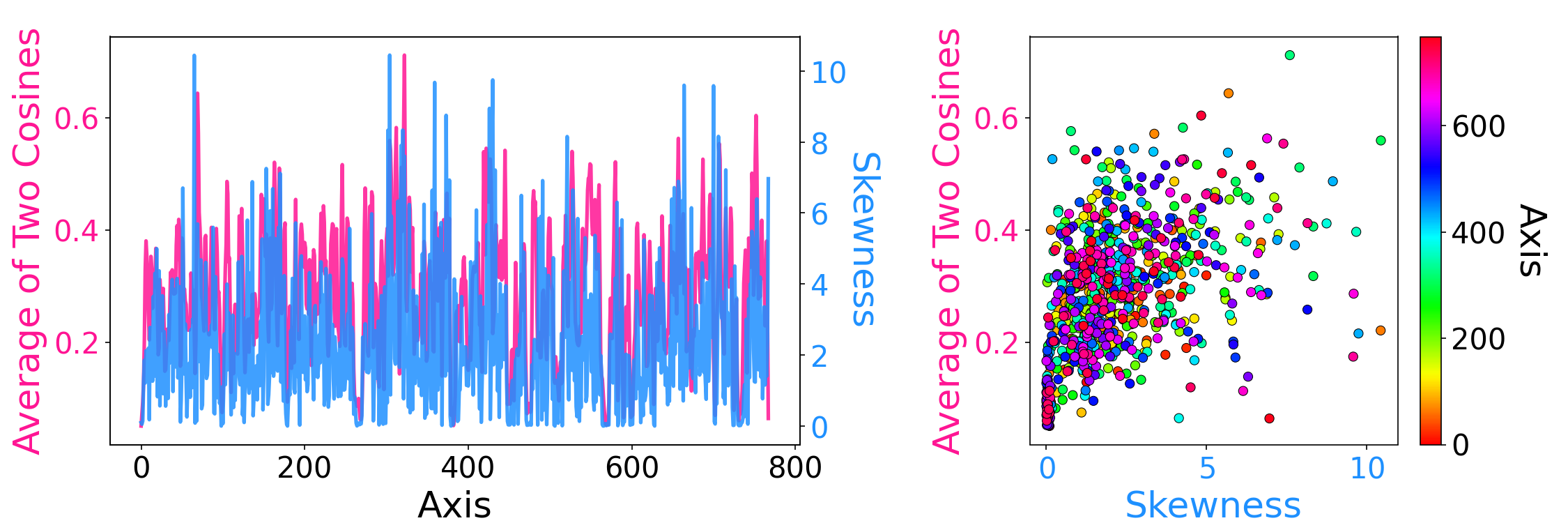}
    \subcaption{Axis Tour}
    \label{fig:bert-cos-skew-plot-axistour}
\end{subfigure}
\begin{subfigure}{\columnwidth}
    \includegraphics[width=\textwidth]{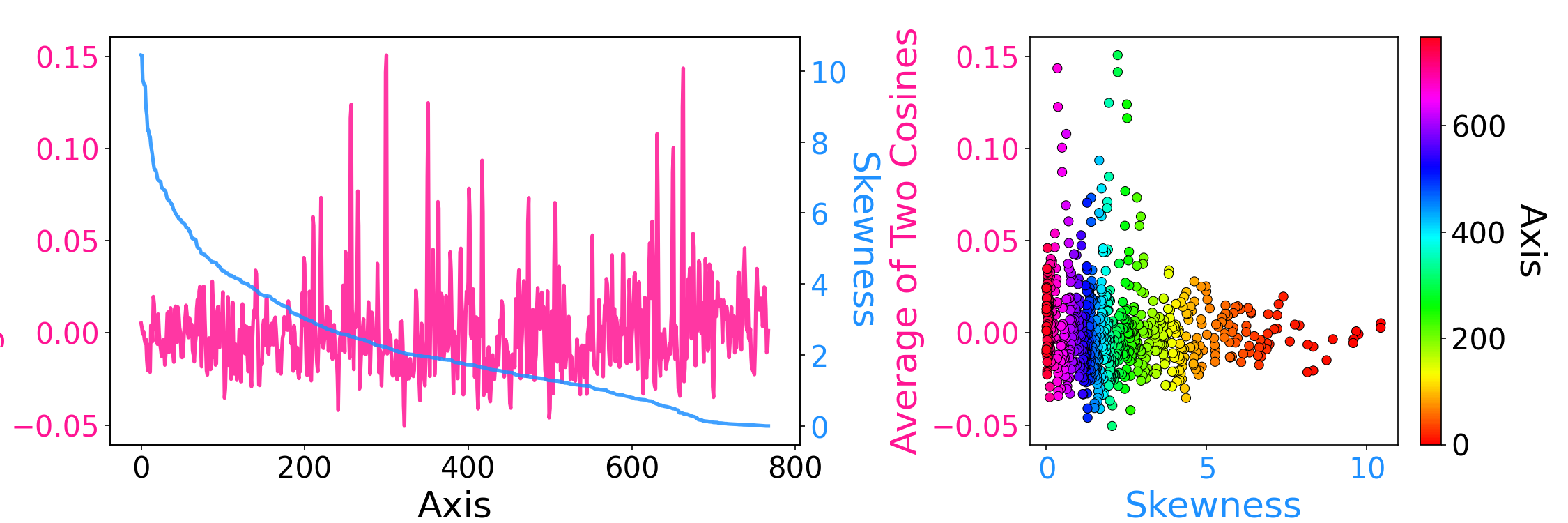}
    \subcaption{Skewness Sort}
    \label{fig:bert-cos-skew-plot-skewnessort}
\end{subfigure}
    \caption{
For BERT, the relationship between the skewness and the average of consecutive two cosines for (a) Axis Tour and (b) Skewness Sort.
Spearman's rank correlation is $0.50$ for Axis Tour, while it is $-0.10$ for Skewness Sort.
}
\label{fig:bert-cos-skew-plot}
\end{figure}

Figures~\ref{fig:word2vec-cos-skew-plot} and~\ref{fig:bert-cos-skew-plot} illustrate the relationship between the skewness and the average of consecutive two cosine similarities for Axis Tour and Skewness Sort.
Interestingly, while the Axis Tour embeddings for BERT show a strong correlation between the skewness and the average of two cosine similarities in Fig.~\ref{fig:bert-cos-skew-plot-axistour}, similar to the results for GloVe in Fig.~\ref{fig:cos-skew-plot-axistour}, the Axis Tour embeddings for word2vec do not show such a correlation in Fig.~\ref{fig:word2vec-cos-skew-plot-axistour}. 

As future work, it may be interesting to consider the detailed relationship between the skewness and the cosine similarity between adjacent axis embeddings, as well as the relationship between the skewness and the strength of meaning of an axis.

\subsubsection{Dimensionality reduction: analogy, word similarity, and categorization tasks}
In this section, similar to Section~\ref{sec:downstream}, we evaluate the performance of dimensionality reduction using the method described in Section~\ref{sec:dim-reduction}. 
It is important to note that since BERT embeddings are dynamic embeddings, the token embeddings obtained from sentences in the One Billion Word Benchmark are different from those in each task.
Therefore, this evaluation focuses only on word2vec embeddings.

\begin{figure*}[t]
    \centering
    \includegraphics[width=\linewidth]{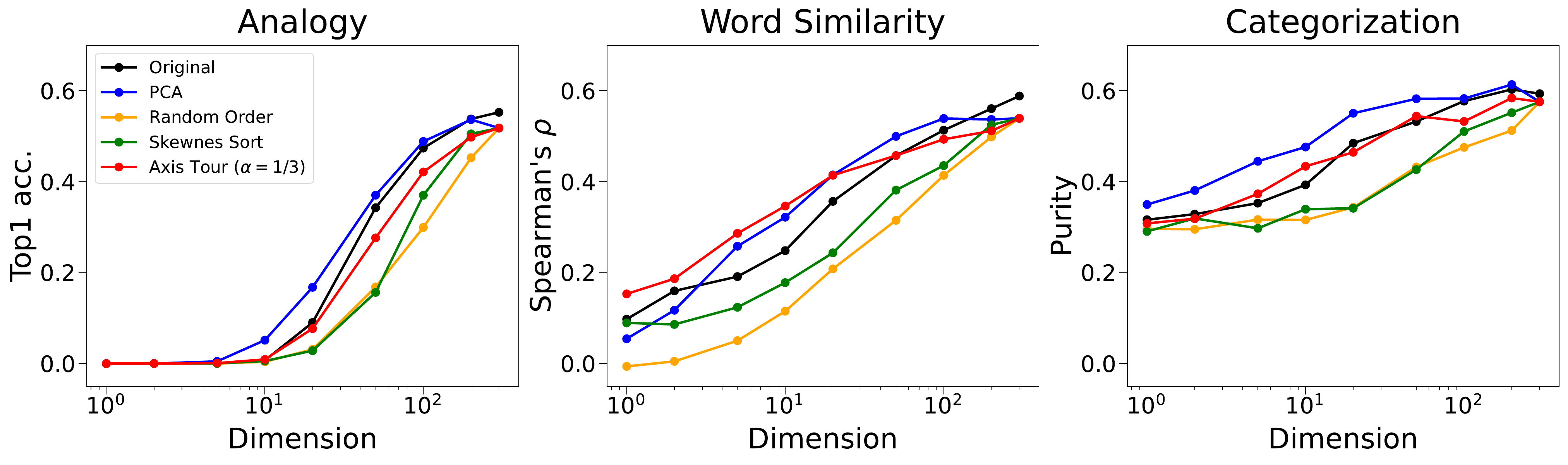}
    \caption{
The performance of dimensionality reduction for the embeddings derived from word2vec.
Each value represents the average of 30 analogy tasks, 8 word similarity tasks, or 6 categorization tasks. 
}
    \label{fig:word2vec-downstream}
\end{figure*}

Figure~\ref{fig:word2vec-downstream} shows the average of each task at $p=1,2,5,10,20,50,100,200,300$ for the embeddings derived from word2vec. Axis Tour outperformed both Random Order and Skewness Sort in each task. In analogy tasks, Axis Tour outperformed PCA in lower dimensions and was nearly equivalent in other dimensions. 

Although PCA remains a strong baseline for analogy and categorization tasks because it is an efficient dimensionality reduction method, these results suggest that there is room for performance improvement. 
However, this study focused primarily on maximizing the semantic continuity of the axes in ICA-transformed embeddings. 
As mentioned in Limitations, it remains an area for future work to construct more efficient low-dimensional embeddings based on the axes in ICA-transformed embeddings.

\section{Topographic ICA (TICA)}\label{app:tica}
Our proposed method, Axis Tour, is related to Topographic Independent Component Analysis (TICA)~\cite{DBLP:books/sp/Kohonen01,DBLP:journals/neco/HyvarinenHI01} in terms of ordering the axes of ICA-transformed embeddings, taking into account the sequential relationship between independent components. 
Therefore, this section first gives a brief overview of TICA. Then, we perform experiments on the TICA embeddings similar to those performed on the Axis Tour embeddings, and then show comparisons of Axis Tour and TICA.

\subsection{Overview of TICA}\label{app:tica-explanation}
Topographic ICA (TICA)~\cite{DBLP:books/sp/Kohonen01,DBLP:journals/neco/HyvarinenHI01} is a variant of linear ICA.
While classic ICA assumes independence of the source components\footnote{Regarding the indices of the axes, the notation for $\mathbf{f}_r=(f_r^{(\ell)})_{\ell=1}^d$, as seen in the Section~\ref{sec:dim-reduction}, would be written as $\mathbf{s}=(s^{(\ell)})_{\ell=1}^d$. However, for the sake of readability, this section uses the notation $\mathbf{s}=(s_\ell)_{\ell=1}^d$.} $\mathbf{s}=(s_\ell)_{\ell=1}^d \in \mathbb{R}^{d}$, TICA allows for positive higher-order correlations $\mathrm{cov}\left(s_\ell^2, s_m^2\right)$ and assumes that the variances of adjacent sources are correlated. In the probabilistic model of TICA, each variance $\sigma_\ell^2$ of source component $s_\ell$ is generated from the factors $\mathbf{u}=(u_\ell)_{\ell=1}^d \in \mathbb{R}^{d}$ as follows:
\begin{align}
    \sigma_\ell &= \phi\left(\sum_{m=1}^d h_{\ell m}u_m \right), \\
    s_\ell &= \sigma_\ell z_\ell,
\end{align}
where $\phi$ is some nonlinear function, $h_{\ell m}$ is a symmetric neighborhood relation, and $z_\ell$ are mutually independent random variables.
To estimate the decomposition matrix $\mathbf{W}=[\mathbf{w}_1,\ldots,\mathbf{w}_d]^\top \in \mathbb{R}^{d\times d}$, where $\mathbf{w}_\ell \in \mathbb{R}^d$, such that $\mathbf{s} = \mathbf{W}\mathbf{x}$ from the mixed signals $\mathbf{x} \in \mathbb{R}^{d}$, TICA maximize the log likelihood 
\begin{align}
    & \log L(\mathbf{W}) = \log p\left(\mathbf{s}, \mathbf{u} \mid \mathbf{W} \right) = \nonumber \\ & \mathbb{E}_{\mathbf{x}} \left[\log \int \prod_{\ell=1}^d \frac{p(z_\ell)p(u_\ell)}{\sigma_\ell} \left|\det \mathbf{W}\right| d \mathbf{u}\right].
\end{align}

By setting $\phi = \sqrt{\cdot}$ and applying some assumptions and approximations, the log likelihood is approximated as
\begin{align}
    \log \tilde L(\mathbf{W}) = \mathbb{E}_{\mathbf{x}}\left[\sum_{m=1}^d G \left( \sum_{\ell=1}^d h_{\ell m}s_\ell^2 \right) \right]\label{eq:tica-approximated-ll}
\end{align}
where $G$ is a function approximated by $G(y) = -\beta_{1/2} \sqrt{y} + \beta_0$ with constant $\beta_{1/2}=0.8, \beta_0=1.2$ in our experiment.
For further details on the derivation, see \citet{DBLP:journals/neco/HyvarinenHI01}.

The optimization of (\ref{eq:tica-approximated-ll}) can be done by gradient descent:
\begin{align}
        \Delta \mathbf{w}_\ell \propto \mathbb{E}_{\mathbf{x}}\left[\mathbf{x}s_\ell \sum_{m=1}^d h_{\ell m} \frac{dG}{dy}\left(\sum_{m'=1}^d h_{m m'} s_{m'}^2 \right) \right].
\end{align}
Note that we have to apply orthonormalization
\begin{align}
    \mathbf{W} \leftarrow (\mathbf{WW}^\top)^{-1/2}\mathbf{W}
\end{align}
to the decomposition matrix at each iteration.
In accordance with the notation in (\ref{eq:ica}), the extracted components $\mathbf{S}$ are expressed by the obtained $\mathbf{W}$ as
\begin{align}
    \mathbf{S} = \mathbf{X}\mathbf{W}^\top.
\end{align}
In summary, TICA, a kind of ICA, can extract topographic ordered features by using higher-order correlations.

\subsection{Experiments}
As in classic ICA, we transform the $300$-dimensional GloVe by TICA for our experiments.
Following the setup of the 1-D experiment in~\citet{DBLP:journals/neco/HyvarinenHI01}, a neighborhood relation was defined by convolving a rectangular-shaped filter $(\ldots, 0, 0, 0, \overbrace{1, 1, \ldots, 1, 1}^{\mathsf{width} ~ \text{times}}, 0, 0, 0, \ldots)$ twice with itself.
\citet{DBLP:journals/neco/HyvarinenHI01} defined the hyperparameter $\mathsf{width}=5$ for 20 components, so for 300 components, we consider two cases where the width is scaled logarithmically or linearly to $\mathsf{width}=9$ and $\mathsf{width}=75$.
These are denoted as TICA9 and TICA75, respectively\footnote{For the TICA9-transformed and TICA75-transformed embeddings, similar to the Axis Tour embeddings, we flip the sign of each axis as needed so that the skewness is positive.}.
We performed $10{,}000$ iterations of gradient descent in both settings.

\subsubsection{Qualitative observation}
\begin{table*}[t!]
\tiny
\centering
\begin{subtable}{\textwidth}
\centering
\begin{tabular}{@{\hspace{0.5em}}c@{\hspace{0.5em}}c@{\hspace{0.5em}}c@{\hspace{0.5em}}c@{\hspace{0.5em}}c@{\hspace{0.5em}}c@{\hspace{0.5em}}c@{\hspace{0.5em}}c@{\hspace{0.5em}}c@{\hspace{0.5em}}c@{\hspace{0.5em}}}
\toprule
       50 &        51 &          52 &       53 &       54 &       55 &        56 &       57 &      58 &       59 \\
\midrule
  coretta &     david &        john & betjeman &    prine &     zuko &     peter &     luís &      de &     paul \\
    janny &    paymer & rhys-davies &     john &    paulk & fabricio & mcwhinney &     joao &   jorge &     juan \\
   duncum &   pannick &       naber & blystone &   kerrys &  farriss &     katis & monteiro & enrique &    pablo \\
  strouse & costabile &    banville & bucchino & spinello &  partida &  lindroos &  pereira &    josé &  mendoza \\
livengood &   zdrilic &       koepp & newcombe &   pessoa &   rionda & fatialofa &     joão &    juan & scharner \\
\midrule
\midrule
       100 &    101 &                102 &        103 &             104 &         105 &           106 &         107 &          108 &       109 \\
\midrule
  sompolno &    lb4 &                  : & nehzatabad &       anobiidae &       tburr &   poeciliidae &       onoba &       wasbir & xerocomus \\
  drużbice & 3.7995 & http://www.***.org &   poshtkuh &          ethmia &     ***.com &     dipodidae &    sinezona & gurubacharya &   firered \\
wartkowice &   4-97 & http://www.***.com &    eqbal-e &    grallariidae & ***@***.com &      pipridae &     anatoma &       tolkun & quilodran \\
dzierzgowo &  13/11 & http://www.***.com &  zahray-ye & sclerosomatidae & ***@***.com &     anthidium & cantharidus &        aomar & nishadham \\
 zanjanrud &      , &        www.***.com & khatunabad &  leiothrichidae &     ***.com & dendrobatidae &     syrnola &           by &     kirka \\
\midrule
\midrule
   150 &           151 &     152 &      153 &      154 &      155 &       156 &      157 &               158 &            159 \\
\midrule
 rilla &        singen &   bijar &    hanus &       gò &    mamat &  boyolali &   dammit & hockeyallsvenskan &       v-league \\
  vard & referees\_mike &    lali & goldring &    ubayy & myogenic &      gono &     swac &             94-86 & lausanne-sport \\
eguchi &        jennys &   prete &    leura & kazagham & ragnvald &     hines & ritholtz &              gøta &           nasl \\
kyeong &     regazzoni & markazi &   larmer & munnetra &    yamba & strayhorn &   yorick &             86-82 &            whl \\
achiou &           x13 &  garone &  similis &   terefe &    swabs &  sabzevar &   elmyra &       second-tier &      divisione \\
\midrule
\midrule
     200 &     201 &        202 &       203 &           204 &         205 &       206 &        207 &        208 &          209 \\
\midrule
two-week & 35-year & epigraphic & 2001-2003 &         snags & sponsorship & spectator &    frantic &    appeals & translations \\
six-week & 50-year &     fogies & 1998-2001 &    objections &        logo &       bhb & hysterical &  frivolous &  translation \\
  a-half & 60-year &        old & 1996-1998 & disagreements &     pitfall &      moot & liquidator & economized &       pashto \\
    five & 25-year &    blighty & 1999-2002 &    procedural &     tagline &     unfit &   quixotic &     appeal &         word \\
     six & 40-year &    gutnish & 1995-1997 &    weaknesses &    drawcard &       sjc &   counting & 360-degree &       arabic \\
\midrule
\midrule
  250 &   251 &    252 &      253 &      254 &    255 &       256 &      257 &     258 &        259 \\
\midrule
 ltte &   gam &  czech &  russian &  anatoly & korean &    indian &     serb & romania &       minh \\
  ira &  aceh & poland &   moscow &  ossetia &  south & mukherjee &  croatia & moldova &     nsanje \\
eelam &  milf & krakow &   vasily & ossetian &   jang &    bengal &   zagreb &    inyo &    chennai \\
tamil & rebel & warsaw &   sergei &     oleg &  korea & jharkhand &  serbian & iceland & srinivasan \\
 fein &  mnlf & polish & vladimir & interfax &    kim &     bihar & croatian &  norway &   chikwawa \\
\bottomrule
\end{tabular}
\caption{TICA9}
\label{tab:TICAw9-examples}
\end{subtable}
\vspace{2em}

\begin{subtable}{\textwidth}
\centering
\begin{tabular}{@{\hspace{0.5em}}c@{\hspace{0.5em}}c@{\hspace{0.5em}}c@{\hspace{0.5em}}c@{\hspace{0.5em}}c@{\hspace{0.5em}}c@{\hspace{0.5em}}c@{\hspace{0.5em}}c@{\hspace{0.5em}}c@{\hspace{0.5em}}c@{\hspace{0.5em}}}
\toprule
        50 &       51 &          52 &        53 &        54 &       55 &        56 &          57 &         58 &             59 \\
\midrule
gartenberg &    faden & trescothick &      .652 & billcliff &     dani &  lentulus &      s.s.d. & dakhlallah &     matthewson \\
      nart &   tilles &      bopara & ramnaresh &     plzen &  siggins & sivasspor &      torrin &   ollivier &         feiner \\
 radziwill & kolderie &      harmse &     razaq &   hakohen &    feith &     najdi &      sastra &  clemetson &       hatchell \\
stachowski &   strada &     briones &   taufeeq &  ashraful & daugaard &      epps & vendémiaire &     herrin & carrickmacross \\
     recio &  durrant &        ruvo &    sarwan &  xiangfan &   chedid &   sohlman &     wenping &    sampley &        lynskey \\
\midrule
\midrule
         100 &       101 &       102 &       103 &         104 &          105 &       106 &         107 &    108 &         109 \\
\midrule
     vnccpbn & lomartire &      kihn & jędrzejów & cooperacion & ***-***-**** &   prusice &  extranjero &    :53 &        sa/b \\
     ustbimp &     rsst1 &    apenas &   pleszew &        cvik &   ohernandez & rmartinez &  finalmente &    qe5 &     ***.com \\
     gromada &     36.41 &     zgray &   sokołów &    hrvatske &         .367 &  kmorales & ***@***.com & jgreig & ***@***.com \\
        a\_21 & darreh-ye & divergens & przysucha &        0255 &        (928) & alifereti &    cantidad & fedbud & ***@***.com \\
***-***-**** & immatures &      nohv &    raciąż &     entidad &       cjones &  separado &     ***.net &   rxe5 & ***@***.com \\
\midrule
\midrule
        150 &              151 &    152 &          153 &     154 &                 155 &        156 &          157 &           158 &    159 \\
\midrule
   reteamed &     leaf-sheaths &  risen &     auguring &       i & backward-compatible & languishes & thursday.the & expeditiously & surged \\
  unbeknown &       upsmanship & surged &     two-song &     n’t &      satisfactorily &        5.4 &     one-disc &      resolved & soared \\
unbeknownst &           udalls & obrija &      18-yard &     ’ve &           outwardly &        5.8 &        50-50 &      sometime &  leapt \\
 co-operate & great-grandchild & soared &   unproduced &  happen &          co-existed &        4.9 &      32-page &      commence & leaped \\
 re-connect &         nineteen &  07:20 & epithalamium & presume &           satisifed &        4.1 &        hefty &            be &  leaps \\
\midrule
\midrule
        200 &           201 &         202 &         203 &      204 &      205 &           206 &             207 &  208 &          209 \\
\midrule
     expert &      ceremony &       exams &   vandalism & adjacent &  rocking &       hazards & recommendations & 27th &       cd-rom \\
    experts &   celebrating & mathematics &      campus &   blocks &      bed & investigators &        advisory & 13th &         mods \\
feasibility &           eve &      ribbon &  antisocial & building &  sitting &        detect &           301st & 10th &    evolution \\
   forensic &        parade &        math &    graffiti &  erected &   padded &     exposures &     advancement & 15th &      famines \\
      study & commemorating &      school & streetscape &    built & pounding &   remediation &             psc & 23rd & experimental \\
\midrule
\midrule
       250 &          251 &         252 &              253 &      254 &          255 &           256 &     257 &     258 &       259 \\
\midrule
   titanic &     saratoga &    smolensk &           ms-dos &       2g &          pol &   supermarine & ***.com &     ump &  under-17 \\
      clot &         1772 &  hasselblad &            pelts &     enos &         muse &         xk120 &  aretha &     iwf &      huon \\
    vessel &         1793 &   bonifacio & production-based & blandest & plasterboard & ground-attack &  banana &   icing &     pixie \\
    tanker & thoroughbred & cornplanter &       conquering &    pagan &        spray &  turbocharged &   canes &    mapo & southwood \\
stotesbury &         1794 &        1314 &             1405 &    istar &          bmw &       madelyn &  amazon & vedanta &    trikke \\
\bottomrule
\end{tabular}
\caption{TICA75}
\label{tab:TICAw75-examples}
\end{subtable}
\vspace{2em}
\caption{
Semantic continuity of axes for normalized TICA9-transformed and TICA75-transformed embeddings. For the 50th, 100th, 150th, 200th, and 250th axes, we extract ten consecutive axes from each of these axes and display the top five words for each of the extracted axes.
}
\label{tab:TICAw9-TICAw75-examples}
\end{table*}

\begin{figure*}[!t]
    \centering
    \begin{minipage}{0.33\linewidth}
        \centering
        \includegraphics[width=\linewidth]{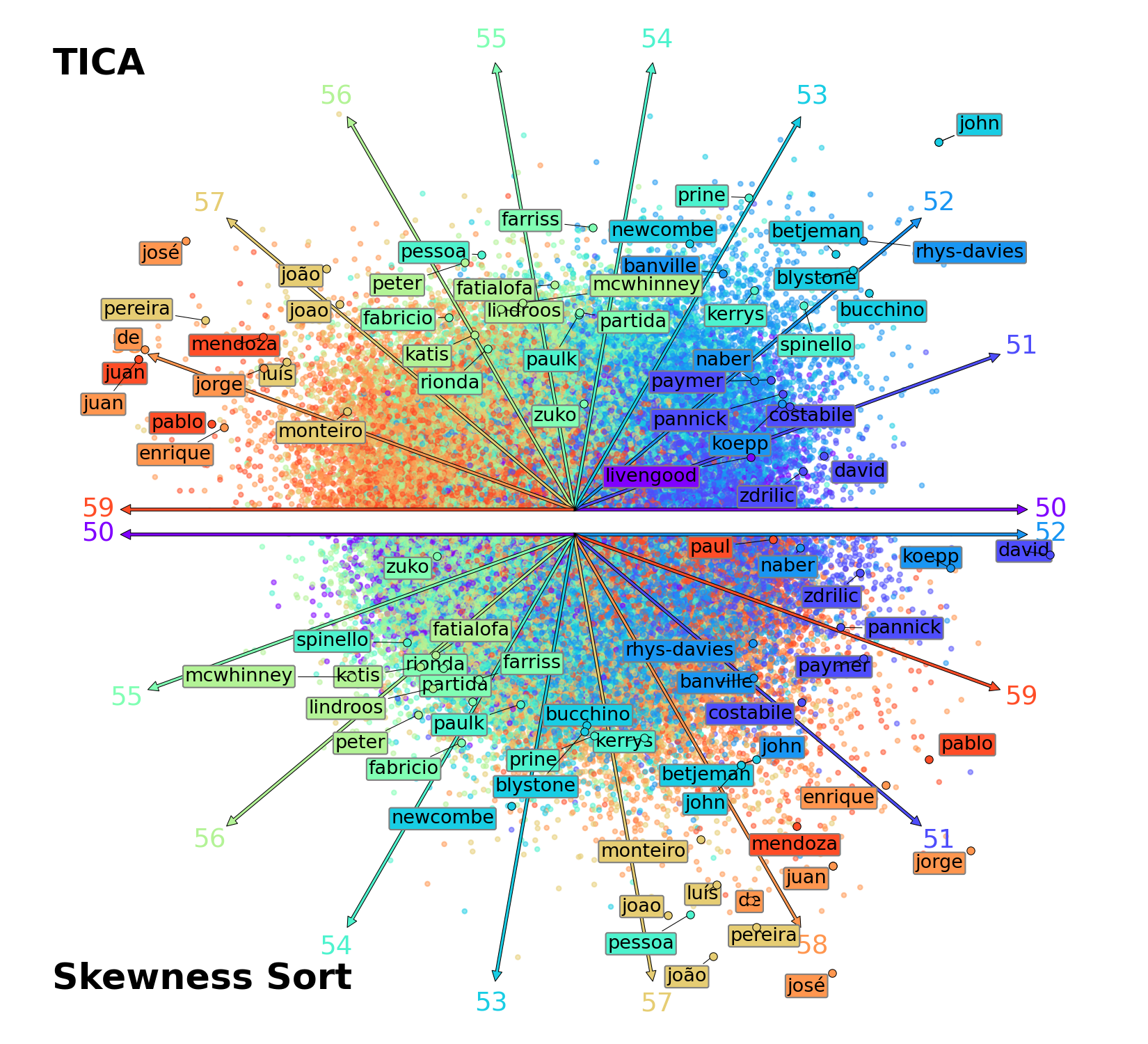}
        \subcaption{The 50th axis to the 59th axis}
        \label{fig:TICAw9-50}
    \end{minipage}\hfill
    \begin{minipage}{0.33\linewidth}
        \centering
        \includegraphics[width=\linewidth]{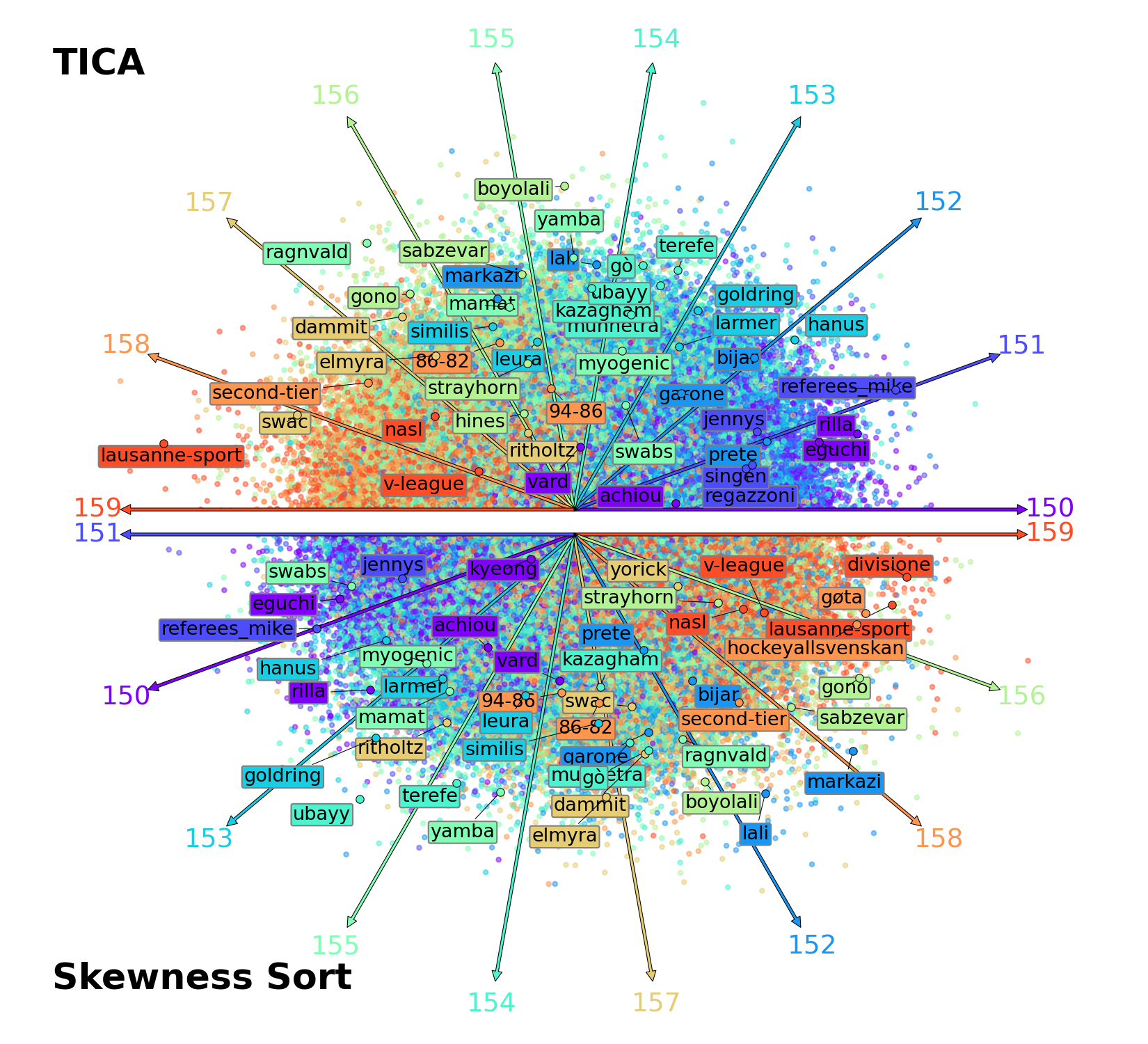}
        \subcaption{The 150th axis to the 159th axis}
        \label{fig:TICAw9-150}
    \end{minipage}
    \begin{minipage}{0.33\linewidth}
        \centering
        \includegraphics[width=\linewidth]{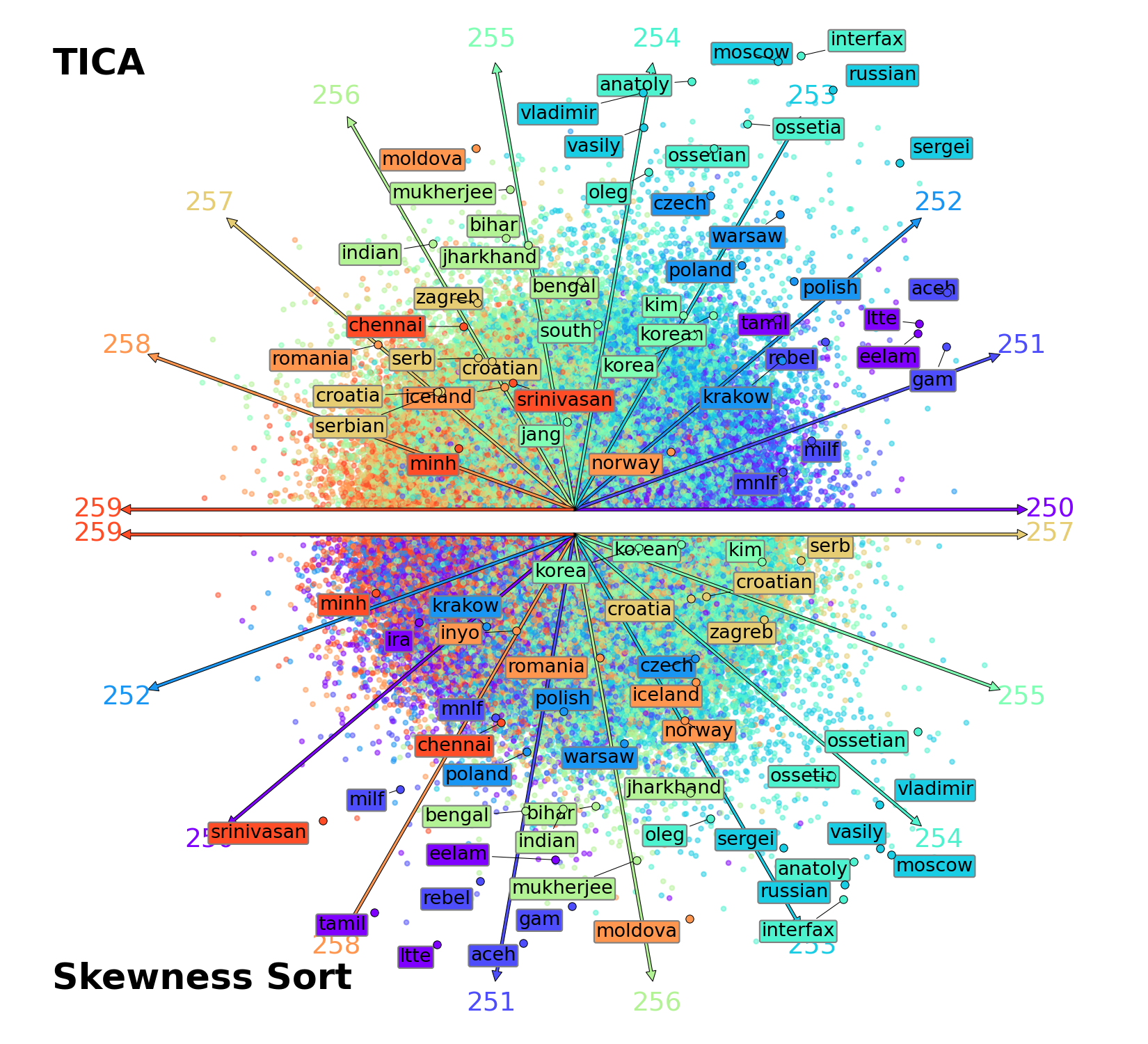}
        \subcaption{The 250th axis to the 259th axis}
        \label{fig:TICAw9-250}
    \end{minipage}
    \caption{For TICA9, the scatterplots of the two-dimensional projections for the axes from the 50th to the 59th, from the 150th to the 159th, and from the 250th to the 259th in Table~\ref{tab:TICAw9-examples}.}
    \label{fig:TICAw9-plot}
\end{figure*}

\begin{figure*}[!t]
    \centering
    \begin{minipage}{0.33\linewidth}
        \centering
        \includegraphics[width=\linewidth]{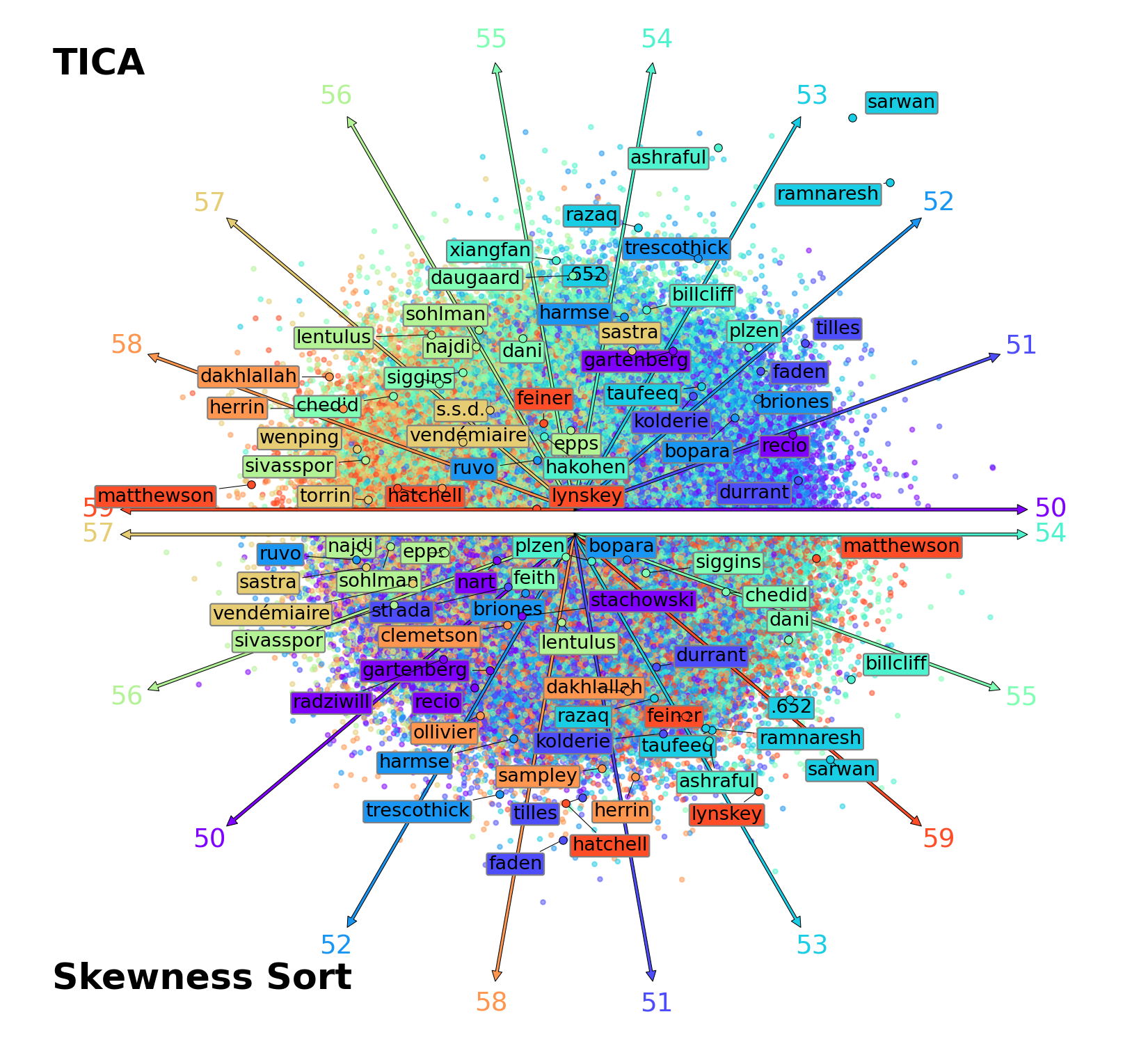}
        \subcaption{The 50th axis to the 59th axis}
        \label{fig:TICAw75-50}
    \end{minipage}\hfill
    \begin{minipage}{0.33\linewidth}
        \centering
        \includegraphics[width=\linewidth]{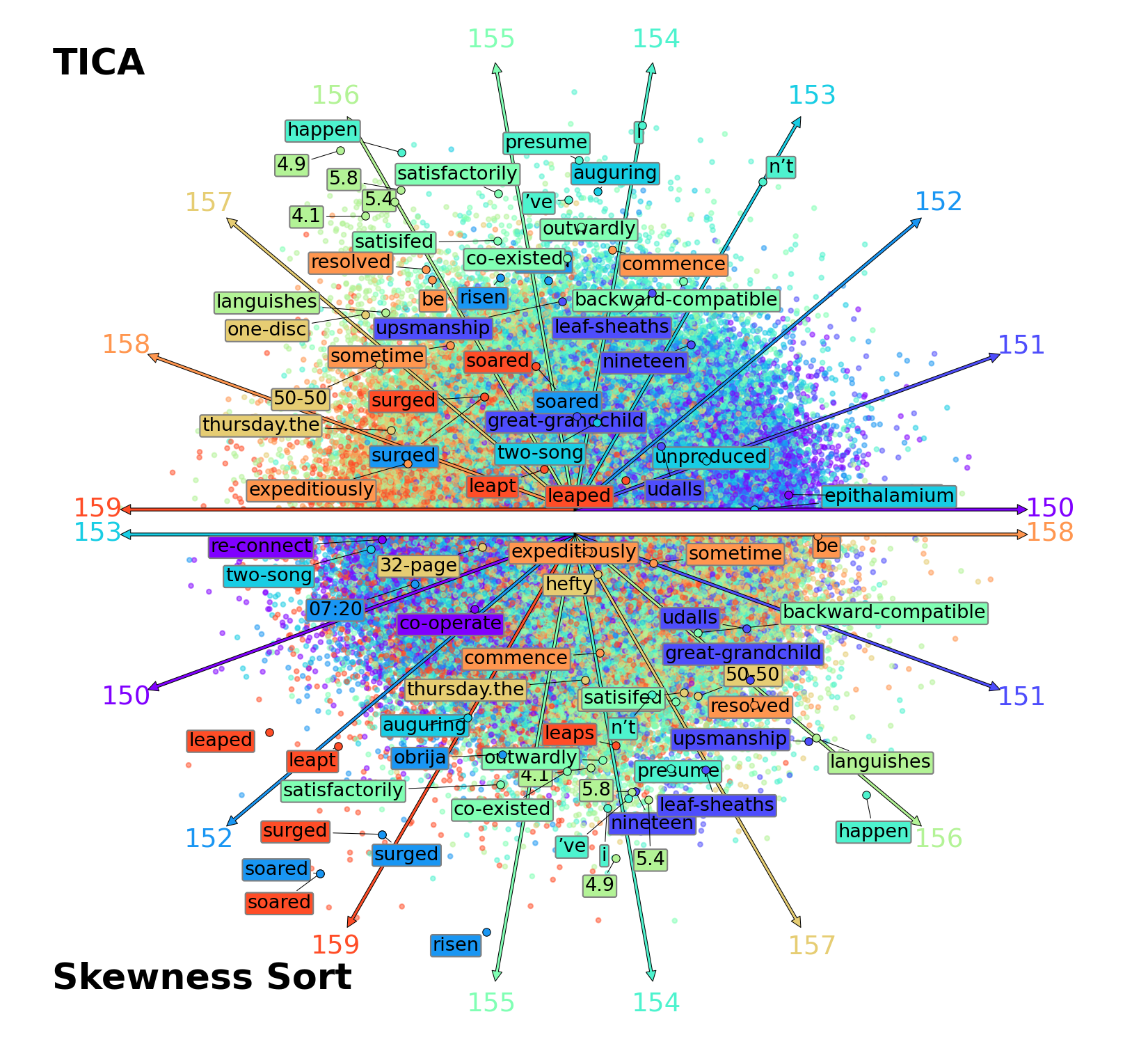}
        \subcaption{The 150th axis to the 159th axis}
        \label{fig:TICAw75-150}
    \end{minipage}
    \begin{minipage}{0.33\linewidth}
        \centering
        \includegraphics[width=\linewidth]{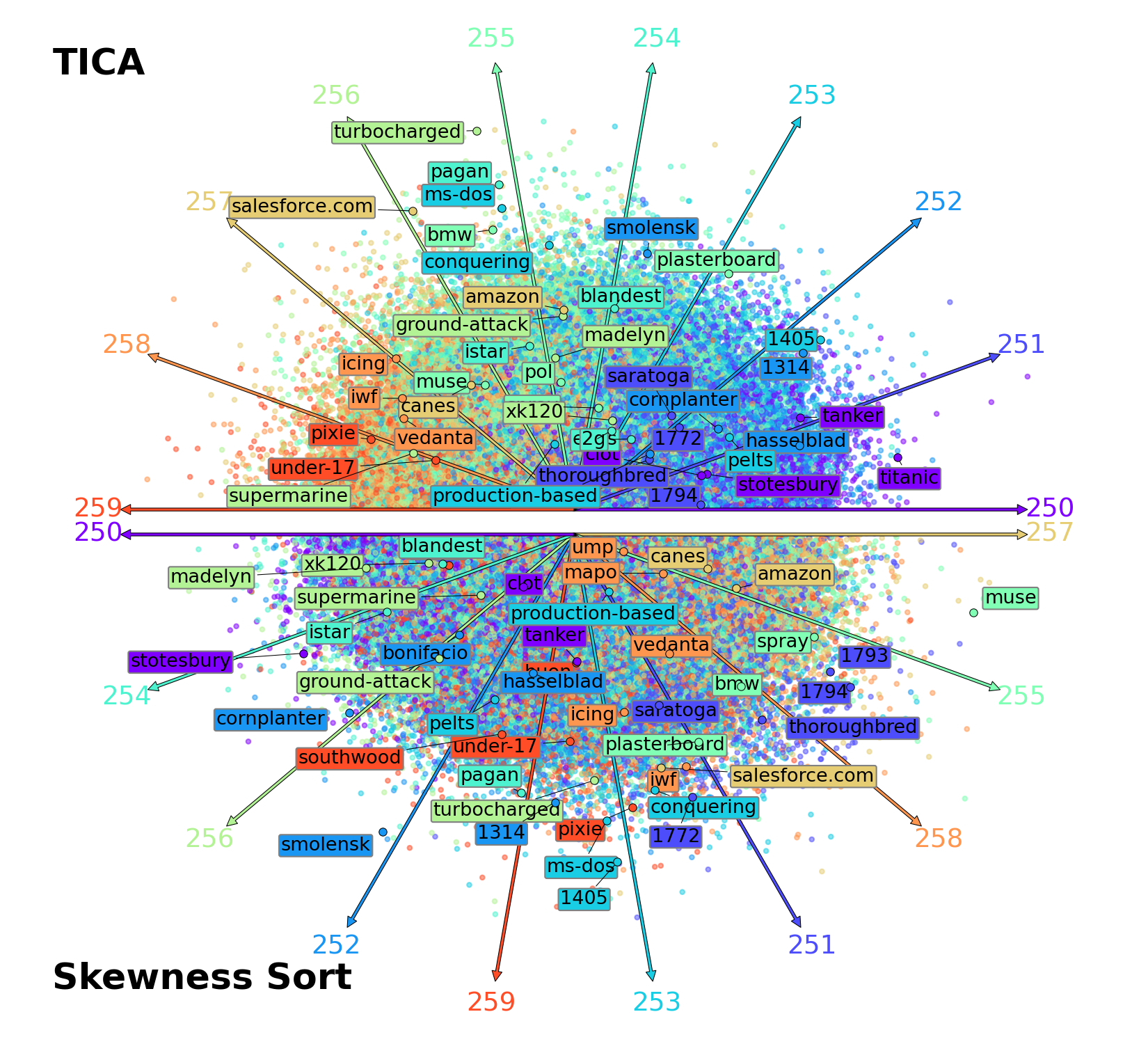}
        \subcaption{The 250th axis to the 259th axis}
        \label{fig:TICAw75-250}
    \end{minipage}
    \caption{For TICA75, the scatterplots of the two-dimensional projections for the axes from the 50th to the 59th, from the 150th to the 159th, and from the 250th to the 259th in Table~\ref{tab:TICAw75-examples}.}
    \label{fig:TICAw75-plot}
\end{figure*}

\begin{figure*}[!t]
    \centering
    \begin{minipage}{0.33\linewidth}
        \centering
        \includegraphics[width=\linewidth]{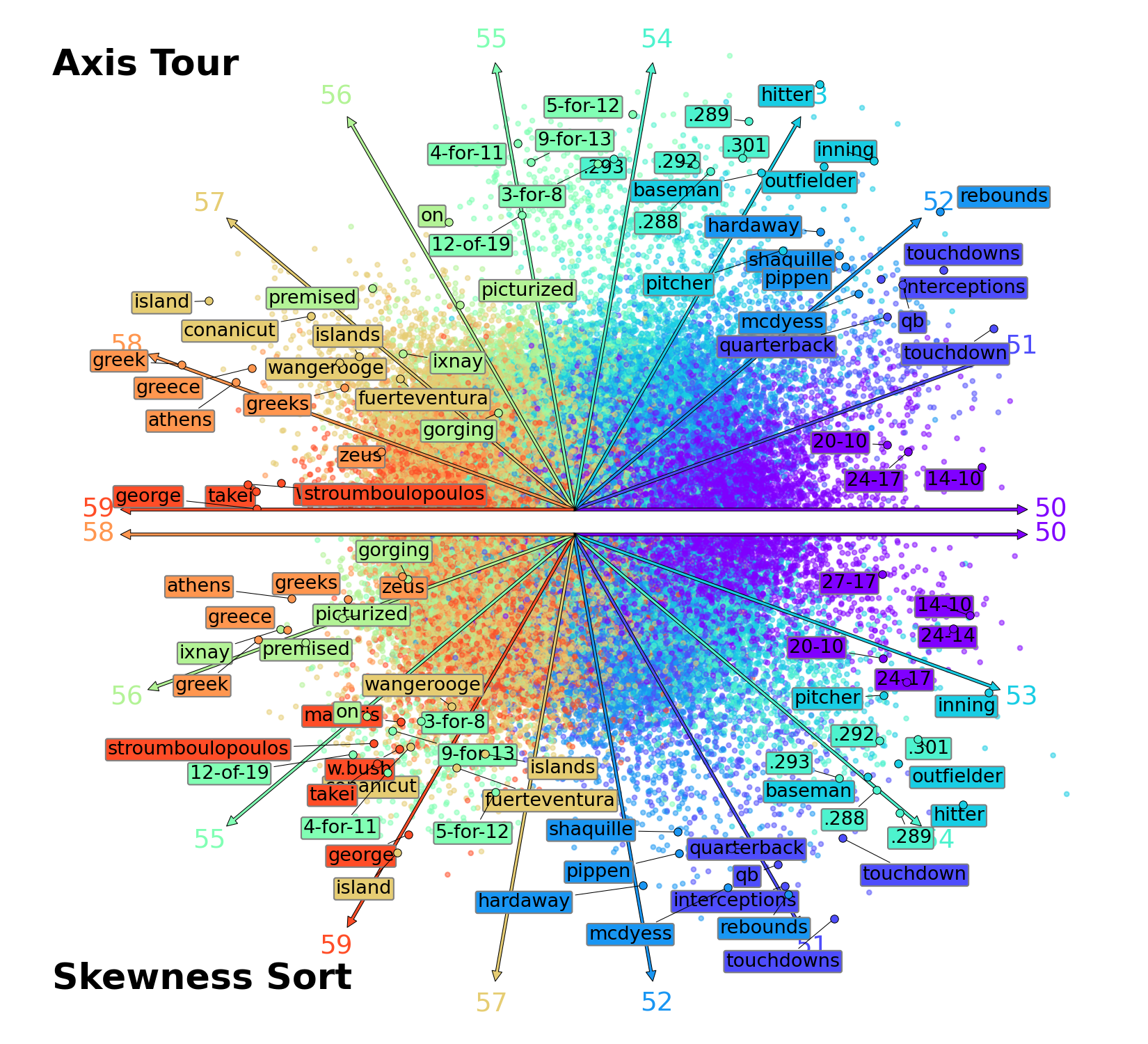}
        \subcaption{The 50th axis to the 59th axis}
        \label{fig:glove-50}
    \end{minipage}\hfill
    \begin{minipage}{0.33\linewidth}
        \centering
        \includegraphics[width=\linewidth]{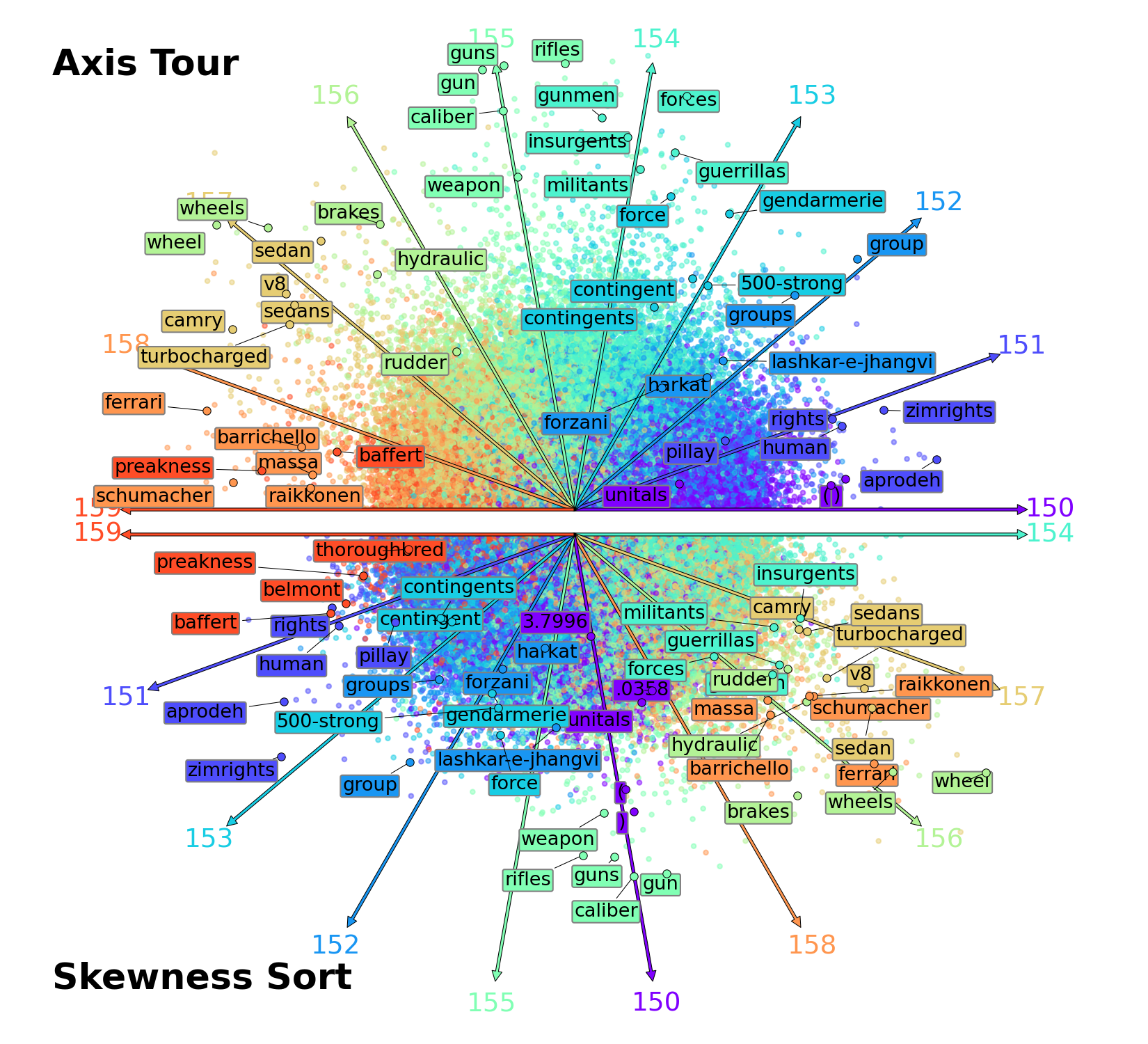}
        \subcaption{The 150th axis to the 159th axis}
        \label{fig:glove-150}
    \end{minipage}
    \begin{minipage}{0.33\linewidth}
        \centering
        \includegraphics[width=\linewidth]{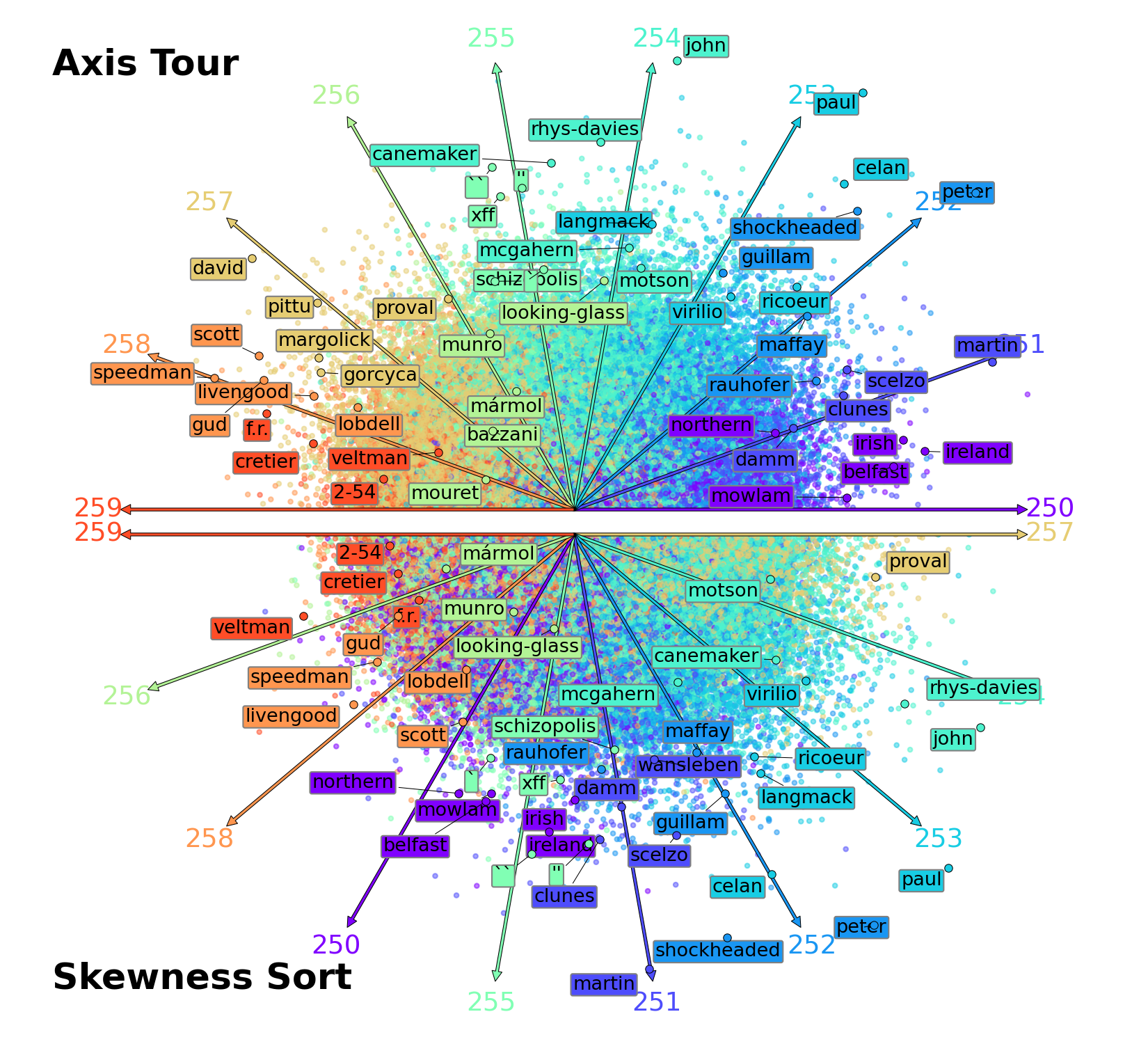}
        \subcaption{The 250th axis to the 259th axis}
        \label{fig:glove-250}
    \end{minipage}
    \caption{For Axis Tour, the scatterplots of the two-dimensional projections for the axes from the 50th to the 59th in Table~\ref{tab:0-119}, from the 150th to the 159th in Table~\ref{tab:120-239}, and from the 250th to the 259th  in Table~\ref{tab:240-299}.}
    \label{fig:glove-plot}
\end{figure*}

Table~\ref{tab:TICAw9-TICAw75-examples} shows the examples of the axes of the TICA9-transformed embeddings and the TICA75-transformed embeddings. 
Similar to Axis Tour, we observe the semantic continuity of the axes in some subintervals. 
For example, the 250th to the 259th axes in Table~\ref{tab:TICAw9-examples} show the semantic continuity with respect to \emph{countries}. 
Moreover, for these examples in Table~\ref{tab:TICAw9-TICAw75-examples}, TICA9 seems to show the clearer semantic continuity of the axes compared to TICA75. 
Note that URLs, email addresses and phone numbers are anonymized in Table~\ref{tab:TICAw9-TICAw75-examples}.

\subsubsection{Scatterplots of Table~\ref{tab:TICAw9-TICAw75-examples}}
Figures~\ref{fig:TICAw9-plot} and~\ref{fig:TICAw75-plot} show the scatterplots of the two-dimensional projections for the the axes of the TICA9-transformed embeddings and the TICA75-transformed embeddings in Tables~\ref{tab:TICAw9-examples} and~\ref{tab:TICAw75-examples}.
In these examples, for both TICA9 and TICA75, the interpretability of the embedding space from the scatterplots decreases as the top words of the axes show significant movement or approach the origin. 
This property is particularly noticeable in TICA75. For example, in Fig.~\ref{fig:TICAw75-150}, the top words of the 159th axis are far from the axis: \emph{surged} and \emph{soared} are concentrated near the 156th axis, and \emph{leapt} and \emph{leaped} are concentrated near the origin.

\begin{table}[t]
\centering
\begin{tabular}{llrrr}
\toprule
Fig. &  embeddings & $d_I$ & Skew. $d_I$ & diff\\
\midrule
\ref{fig:TICAw9-50} & \multirow{3}{*}{TICA9} & 0.61 & 0.59 & 0.02\\
\ref{fig:TICAw9-150} &  & 0.33 & 0.35 & -0.02\\
\ref{fig:TICAw9-250} &  & 0.57 & 0.54 & 0.03\\
\ref{fig:TICAw75-50} & \multirow{3}{*}{TICA75} & 0.35  &  0.31 & 0.04\\
\ref{fig:TICAw75-150} &  & 0.44 &  0.44 & 0.00\\
\ref{fig:TICAw75-250} &  & 0.31 &  0.34 & -0.03\\
\midrule
\ref{fig:glove-50} & \multirow{3}{*}{Axis Tour} & 0.78 & 0.73 & 0.05\\
\ref{fig:glove-150} &  & 0.72 & 0.58 & 0.14\\
\ref{fig:glove-250} &  & 0.57 & 0.52 & 0.05\\
\midrule
\midrule
\multirow{3}{*}{Avg.} & TICA9 & 0.59 & 0.54 & 0.05\\
& TICA75 & 0.41 & 0.41 & 0.00\\
& Axis Tour & 0.68 & 0.59 & 0.09\\
\bottomrule
\end{tabular}
\caption{
The values of $d_I$ for Figs.~\ref{fig:TICAw9-plot},~\ref{fig:TICAw75-plot} and~\ref{fig:glove-plot}, and the average values of $d_I$ over 300 subintervals with $|I|=10$. \emph{Skew.} stands for Skewness Sort.
}
\label{tab:tica-axistour-d-I}
\end{table}

\begin{table}[t]
\centering
\begin{tabular}{llrrr}
\toprule
Fig. &  embeddings & $c_I$ & Skew. $c_I$ & diff\\
\midrule
\ref{fig:TICAw9-50} & \multirow{3}{*}{TICA9} & 0.35 & 0.21 & 0.14 \\
\ref{fig:TICAw9-150} & & 0.09 & 0.03 & 0.06\\
\ref{fig:TICAw9-250} & & 0.25 & 0.16 & 0.09\\
\ref{fig:TICAw75-50} & \multirow{3}{*}{TICA75} & 0.07 & 0.06 & 0.01 \\
\ref{fig:TICAw75-150} & &  0.06 & 0.11 & -0.05\\
\ref{fig:TICAw75-250} & &  0.09 & 0.06 & 0.03\\
\midrule
\ref{fig:glove-50} & \multirow{3}{*}{Axis Tour} & 0.27 & 0.11 & 0.16\\
\ref{fig:glove-150} & & 0.27 & 0.07 & 0.20\\
\ref{fig:glove-250} & & 0.12 & 0.04 & 0.08\\
\midrule
\midrule
\multirow{3}{*}{Avg.} & TICA9 & 0.23 & 0.14 & 0.09\\
& TICA75 & 0.13 & 0.13 & 0.00\\
& Axis Tour & 0.24 & 0.09 & 0.15\\
\bottomrule
\end{tabular}
\caption{
The values of $c_I$ for Figs.~\ref{fig:TICAw9-plot},~\ref{fig:TICAw75-plot} and~\ref{fig:glove-plot}, and the average values of $c_I$ over 300 subintervals with $|I|=10$. \emph{Skew.} stands for Skewness Sort.
}
\label{tab:tica-axistour-c-I}
\end{table}

\begin{figure}[t!]
    \centering
    \includegraphics[width=\columnwidth]{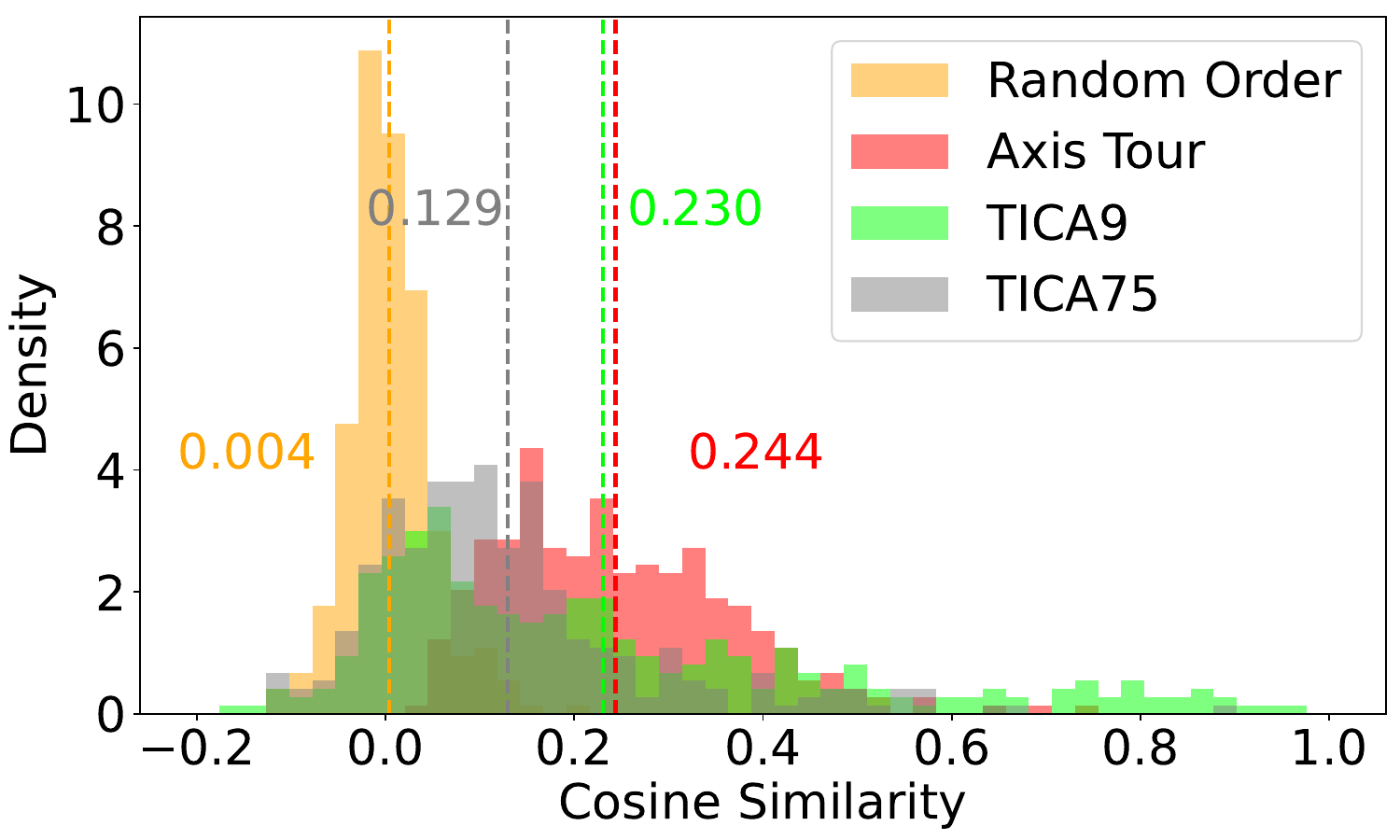}
    \caption{
Histograms of cosine similarities between adjacent axis embeddings for the TICA-transformed embeddings, the Axis Tour embeddings and the Random Order embeddings. 
}
\label{fig:TICA-axistour-cossims}
\end{figure}

\begin{figure}[t!]
    \centering
    \includegraphics[width=\columnwidth]{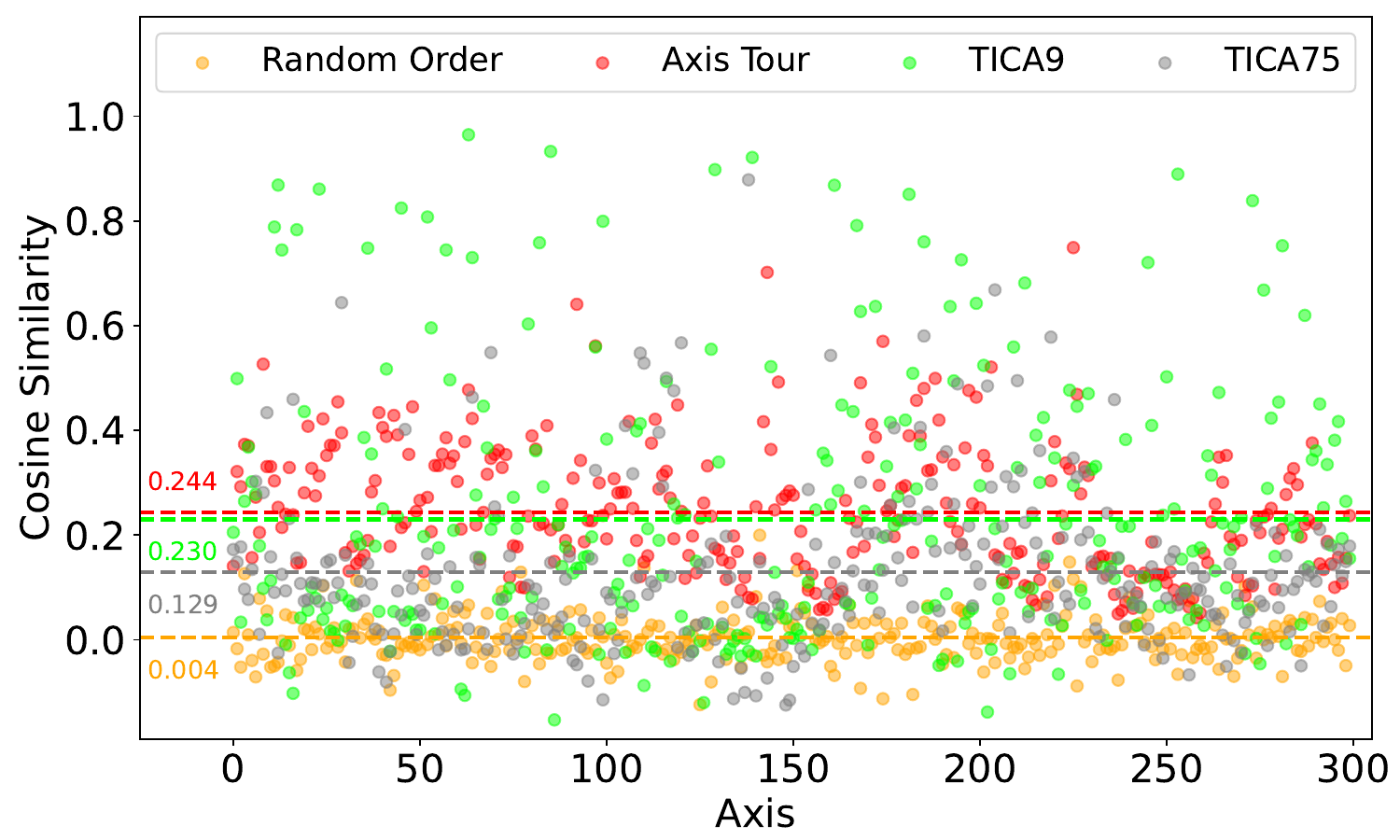}
    \caption{
Scatterplots of $\cos{(\mathbf{v}_\ell,\mathbf{v}_{\ell+1})}$ for Random Order, Axis Tour, TICA9 and TICA75.
}
\label{fig:tica-axsitour-cos-scatter}
\end{figure}

\begin{figure*}[t!]
    \centering
    \includegraphics[width=\linewidth]{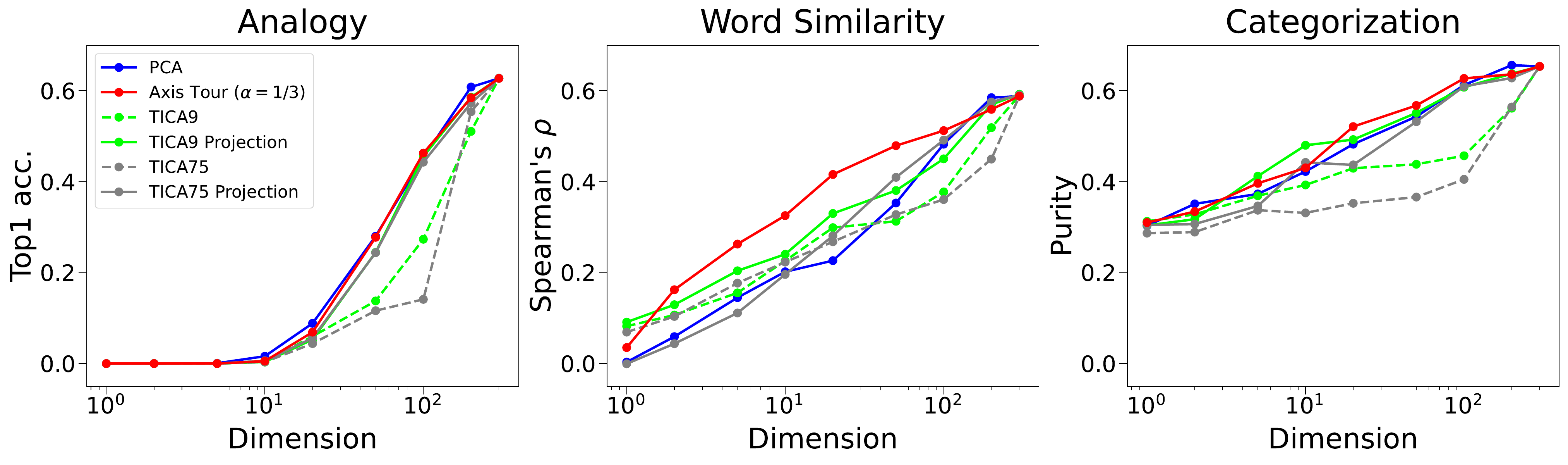}
    \caption{
The performance of dimensionality reduction for the PCA-transformed embeddings, the Axis Tour embeddings with $\alpha=1/3$, the TICA-transformed embeddings, and the TICA Projection embeddings with $\alpha=1/3$. 
Each value represents the average of 30 analogy tasks, 8 word similarity tasks, or 6 categorization tasks. 
}
    \label{fig:tica-downstream}
\end{figure*}

\begin{figure}[t!]
    \centering
    \includegraphics[width=\columnwidth]{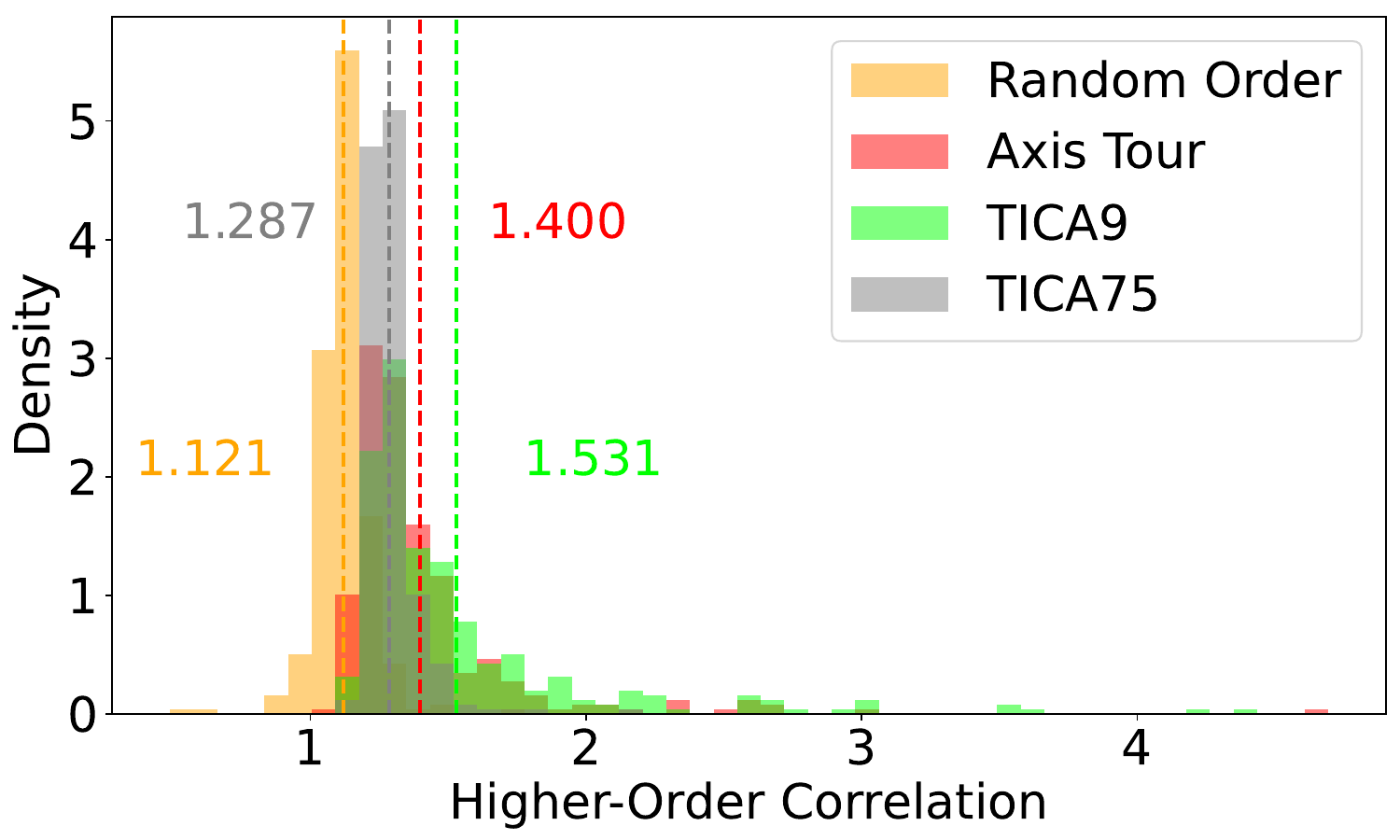}
    \caption{
Histograms of higher-order correlations between adjacent axes for the TICA-transformed embeddings, the Axis Tour embeddings, and the Random Order embeddings. 
}
\label{fig:TICA-axistour-enercorr}
\end{figure}

\subsection{Comparisons of Axis Tour and TICA}\label{app:axistour-vs-tica} 

\subsubsection{Scatterplots}
This section compares the scatterplots of the TICA9-transformed embeddings in Fig.~\ref{fig:TICAw9-plot} and the TICA75-transformed embeddings in Fig.~\ref{fig:TICAw75-plot} with those of Axis Tour. 
Since TICA9, TICA75, and Axis Tour yield different embeddings, we also compare them based on Skewness Sort.

First, similar to Figs.~\ref{fig:TICAw9-plot} and~\ref{fig:TICAw75-plot}, we use 300-dimensional GloVe and show the scatterplots for comparison\footnote{
Note that Figs.~\ref{fig:intro} and ~\ref{fig:examples} show the scatterplots for the illustrative examples, so it is unfair to compare them directly with those of TICA.
} in Fig.~\ref{fig:glove-plot}, which shows the two-dimensional projections for the axes of the Axis Tour embeddings.

Then Table~\ref{tab:tica-axistour-d-I} shows the values of $d_I$ defined in Appendix~\ref{app:evalmetric} for Figs.~\ref{fig:TICAw9-plot},~\ref{fig:TICAw75-plot}, and~\ref{fig:glove-plot}, and the average values of $d_I$ over 300 subintervals with $|I|=10$. 
In these examples, Axis Tour shows larger $d_I$ values and larger differences with Skewness Sort compared to TICA9 and TICA75.
These results are also observed for the average values. 
In addition, we can see the cases where $d_I$ is smaller than that of Skewness Sort in both TICA9 and TICA75, such as Figs.~\ref{fig:TICAw9-150} and~\ref{fig:TICAw75-250}.

Table~\ref{tab:tica-axistour-c-I} shows the values of $c_I$ defined\footnote{
For the TICA9- and TICA75-transformed embeddings, as with Axis Tour embeddings, $c_I$ can be computed by the cosine similarities of adjacent axis embeddings.
} in Appendix~\ref{app:evalmetric} for Figs.~\ref{fig:TICAw9-plot},~\ref{fig:TICAw75-plot}, and~\ref{fig:glove-plot}, and the average values of $c_I$ over 300 subintervals with $|I|=10$. 
In these examples, TICA9 shows the $c_I$ values comparable to Axis Tour.
In fact, their average values are almost the same. 
However, TICA9 shows larger $c_I$ values even when Skewness Sort is performed on $I$. This suggests that for the TICA-transformed embeddings, the axes with similar meanings tend to cluster locally rather than being ordered so that the meanings are continuous. This hypothesis could explain why, in the scatterplots such as Figs.~\ref{fig:TICAw9-plot} and~\ref{fig:TICAw75-plot}, the top words of the axes show significant movement or approach the origin. 

In TICA75, the $c_I$ values are lower, and the difference in the average value compared to Skewness Sort is 0. This indicates that the axes with less similar meanings are grouped together in $I$, resulting in no difference in $c_I$ between TICA and Skewness Sort.

\subsubsection{Cosine similarity and higher-order correlation}
Figure~\ref{fig:TICA-axistour-cossims} shows the histograms of $\cos(\mathbf{v}_\ell,\mathbf{v}_{\ell+1})$ of the TICA-transformed embeddings, the Axis Tour embeddings and the Rnadom Order embeddings.
Similar to Axis Tour, The distribution for TICA shifts towards a more positive mean than that of Random Order. 
Note that while TICA9 has an average close to that of Axis Tour, it has a significantly higher variance of cosine similarity.

To highlight this, Fig.~\ref{fig:tica-axsitour-cos-scatter} shows scatterplots of cosine similarities between adjacent axis embeddings. 
The variance is $0.05$ for Random Order, $0.12$ for Axis Tour, $0.25$ for TICA9, and 0.15 for TICA75. 
The variance for TICA9 is significantly larger than that for Axis Tour, which indicates that the semantic continuity of the axes changes more drastically in the TICA9-transformed embeddings. 

Figure~\ref{fig:TICA-axistour-enercorr} shows the histograms of higher-order correlations between
adjacent axes of the TICA-transformed embeddings, the Axis Tour embeddings and the Rnadom Order embeddings. 
We can see that the average higher-order correlation for TICA is higher than that for Axis Tour, reflecting the learning settings of TICA.

\subsubsection{Dimensionality reduction: analogy, word similarity, and categorization tasks}
This section performs dimensionality reduction for TICA, selecting the axes sequentially starting from the first, similar to Random Order and Skewness Sort.

As we saw in Appendix~\ref{app:dimreduction-by-projection}, we defined Random Order Projection and Skewness Sort Projection by the dimensionality reduction process similar to that of Axis Tour. A similar approach can be considered for TICA, and we call this method TICA projection.

\begin{table*}[t]
\small
\centering
\begin{tabular}{ll}
\toprule
Role & Prompt \\
\midrule
system & You are an excellent NLP annotator. Your response should be in JSON format with the key `choice'. \\
\midrule
\multirow{3}{*}{user} & Which of the following words are related to the words [\textit{top1 word},$\ldots$, \textit{top10 word}]. Answer A or B. \\
& A. [\textit{Axis Tour top1 word},$\ldots$, \textit{Axis Tour top10 word}] \\
& B. [\textit{Skewness Sort top1 word},$\ldots$, \textit{Skewness Sort top10 word}] \\
\bottomrule
\end{tabular}
\caption{
Prompts for the GPT models.  
Since the Axis Tour embeddings and the Skewness Sort embeddings differ solely in the order of the axes, they share $d\,(=300)$ common axes.  
The top 10 words for each common axis are denoted as [\textit{top1 word},$\ldots$, \textit{top10 word}].  
We focus on the next axis of Axis Tour and the next axis of Skewness Sort, based on the common axis.
The top 10 words for the Axis Tour axis are denoted as [\textit{Axis Tour top1 word},$\ldots$, \textit{Axis Tour top10 word}], while the top 10 words for the Skewness Sort axis are denoted as [\textit{Skewness Sort top1 word},$\ldots$, \textit{Skewness Sort top10 word}].  
We then use the top words for the prompt.
}
\label{tab:GPT-prompts}
\end{table*}

\begin{table}[t]
\centering
\begin{tabular}{ll}
\toprule
Model & Version\\
\midrule
GPT-3.5 Turbo & gpt-3.5-turbo-0125\\
GPT-4 Turbo & gpt-4-turbo-2024-04-09\\
GPT-4o & gpt-4o-2024-05-13 \\
GPT-4o mini & gpt-4o-mini-2024-07-18 \\
\bottomrule
\end{tabular}
\caption{Version of each GPT model.}
\label{tab:GPT-models}
\end{table}

Figure~\ref{fig:tica-downstream} shows the average of each task at $p=1,2,5,10,20,50,100,200,300$ for the Axis Tour embeddings, and TICA-transformed embeddings. 
Axis Tour outperformed both TICA and TICA Projection for most dimensions in each task. 
Note that similar to the results for Random Order and Skewness Sort in Appendix~\ref{app:dimreduction-by-projection}, TICA Projection shows performance improvements over TICA in most dimensions. 
These results demonstrate the utility of projection-based dimensionality reduction even in TICA, an algorithm that relaxes the assumption of statistical indepence for ICA.

\section{Details of the quantitative evaluation of semantic continuity by GPT models}\label{app:gpt}
This section provides details on the GPT models, accessed via the OpenAI API, and the prompts used in Section~\ref{sec:GPT}.

The versions of the GPT models used in the experiments are listed in Table~\ref{tab:GPT-models}. 
The prompts used for the experiments are shown in Table~\ref{tab:GPT-prompts}. 
In Fig.~\ref{fig:GPT}, the responses obtained from each model using these prompts are aggregated and shown. 
Note that GPT-4 Turbo failed to provide responses relevant to either Axis Tour or Skewness Sort for two queries.

\end{document}